\documentclass[preprint,12pt]{elsarticle}

\usepackage{setspace}
\usepackage{amssymb}
\usepackage{hyperref}
\usepackage{graphicx}
\usepackage{caption}
\usepackage{subfigure}
\usepackage{booktabs}
\usepackage{float}
\usepackage{placeins}
\usepackage{adjustbox}
\usepackage{amsmath}
\usepackage{bm}
\usepackage{rotating}
\newtheorem{theorem}{Theorem}
\newtheorem{lemma}{Lemma}
\newtheorem{remark}{Remark}
\newtheorem{corollary}{Corollary}
\usepackage{comment}
\usepackage{algorithm}%
\usepackage{algorithmicx}%
\usepackage{algpseudocode}
\usepackage{threeparttable}
\usepackage{xcolor}
\usepackage{multirow}

\begin{document}
\begin{frontmatter}


\title{Randomized Neural Network with Adaptive Forward Regularization for Online Task-free Class Incremental Learning}
\author[1,2]{Junda Wang} \ead{jundawang@sjtu.edu.cn}
\author[3]{Minghui Hu} \ead{e200008@e.ntu.edu.sg}
\author[1,2]{Ning Li} \ead{ning_li@sjtu.edu.cn}
\author[4]{Abdulaziz Al-Ali} \ead{a.alali@qu.edu.qa}
\author[4]{Ponnuthurai Nagaratnam Suganthan\corref{cor1}} \ead{p.n.suganthan@qu.edu.qa}

\cortext[cor1]{Corresponding author}
\affiliation[1]{organization={School of Electronic Information and Electrical Engineering},
             addressline={Shanghai Jiao Tong University},
             city={Shanghai},postcode={200240},
             country={China}}
\affiliation[2]{organization={Key Laboratory of System Control and Information Processing}, addressline={Ministry of Education of China},  city={Shanghai}, postcode={200240},country={China}}
\affiliation[3]{organization={School of Electrical and Electronic Engineering}, addressline={Nanyang Technological University},  city={Singapore}, postcode={639798},country={Singapore}}
\affiliation[4]{organization={Computer Science and Engineering Department, College of Engineering}, addressline={Qatar University},  city={Doha}, postcode={ P.O. Box 2713},country={Qatar}}
\fntext[]{This work was supported by the National Natural Science Foundation of China under Grant 62273230 and 62203302, and the State Scholarship Fund of China Scholarship Council under Grant 202206230182. This paper was submitted to an Elsevier journal in Feb. 2025.}


\begin{abstract}
Class incremental learning (CIL) requires an agent to learn distinct tasks consecutively with knowledge retention against forgetting. Problems impeding the practical applications of CIL methods are twofold: (1) non-i.i.d batch streams and no boundary prompts to update, known as the harsher online task-free CIL (OTCIL) scenario; (2) CIL methods suffer from memory loss in learning long task streams, as shown in Fig. \ref{fig 0} (a). To achieve efficient decision-making and decrease cumulative regrets during \textcolor{black}{the} OTCIL process, a randomized neural network (Randomized NN) with forward regularization (-F) is proposed to resist forgetting and enhance learning performance. This general framework integrates unsupervised knowledge into recursive convex optimization, has no learning dissipation, and can outperform the canonical ridge style (-R) in OTCIL. Based on this framework, we derive the algorithm of the ensemble deep random vector functional link network (edRVFL) with adjustable forward regularization (-$k$F), where $k$ mediates the \textcolor{black}{intensity of the intervention}. edRVFL-$k$F generates one-pass closed-form incremental updates and variable learning rates, effectively avoiding past replay and catastrophic forgetting while achieving superior performance. Moreover, to curb unstable penalties caused by non-i.i.d and mitigate intractable tuning of -$k$F in OTCIL, we improve it to the plug-and-play edRVFL-$k$F-Bayes, enabling all hard $k$s in multiple sub-learners to be self-adaptively determined based on Bayesian learning. Experiments were conducted on 2 image datasets including 6 metrics, dynamic performance, \textcolor{black}{ablation tests, and compatibility,} which distinctly validates the efficacy of our OTCIL frameworks with -$k$F-Bayes and -$k$F styles.
\end{abstract}


\begin{highlights}
\item To better acclimate OTCIL scenarios, \textcolor{black}{forward knowledge is exploited to reduce regret and deliver efficient decision-making for ensemble Randomized NN learning in long task streams. This framework realizes one-pass incremental updates with less loss and superiority over ridge.}
 
\item Based on the framework, edRVFL-$k$F algorithm with adjustable forward regularization is derived, effectively avoiding previous replay and catastrophic forgetting.

\item To overcome the intractable tuning and \textcolor{black}{distribution drifting} of -$k$F, we further propose edRVFL-$k$F-Bayes with $k$s synchronously self-adapted based on Bayesian learning in non-i.i.d OTCIL streams.

\item \textcolor{black}{Extensive experiments were conducted on image datasets and the results were analyzed from multiple views (including 6 metrics, dynamic behaviors, and ablation tests), revealing the outstanding performance of edRVFL-$k$F-Bayes and robustness even with a large PTM.}

\end{highlights}


\end{frontmatter}



\section{Introduction}\label{sec1}

\begin{figure}[!t]
\centering
\subfigure[]{\includegraphics[width=0.53\columnwidth,trim=7mm 2mm 17mm 5mm, clip]{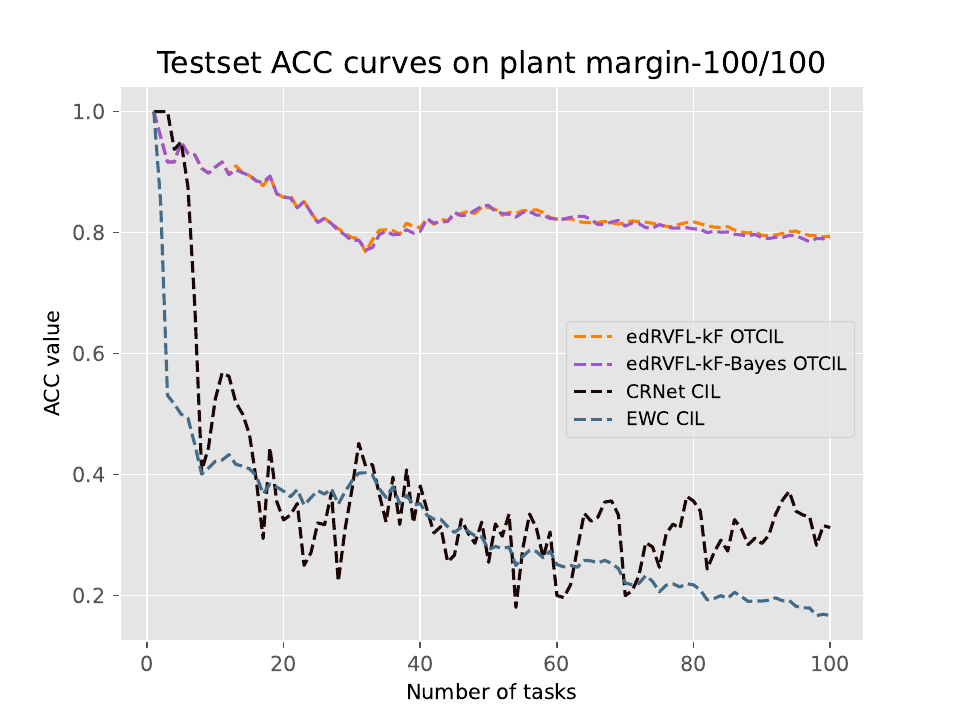}}
\hfil
\subfigure[]{\includegraphics[width=0.28\columnwidth, trim=8mm 6mm 6mm 6mm, clip]{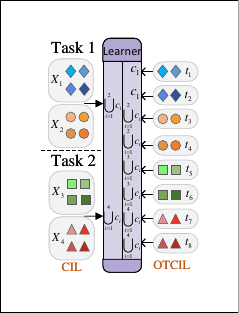}}
\hfil
\caption{(a) The task accuracy of CIL methods on UCI dataset plant margin-100/100. Each task contained one class and replay was forbidden. EWC \cite{6} and CRNet \cite{29} learned in CIL while our edRVFL-$k$F and edRVFL-$k$F-Bayes learned in harsher OTCIL without task i.i.d and boundary. Classic methods show serious knowledge forgetting and intractable adjustments in long task streams, while ours achieve immediate response, high efficacy, and less loss. (b) The diagram of CIL and OTCIL. Dashed line denotes task boundary.}
\label{fig 0}
\end{figure}

Deep neural networks (DNN) have excelled at various complex tasks such as computer vision and natural language processing \cite{bib1,bib11}. \textcolor{black}{A key premise is that offline data is independently and identically distributed (i.i.d), ensuring the weights gradually converge to a consistent optimum for empirical risk minimization (ERM) in the feasible zone. Unfortunately, issues such as catastrophic forgetting and performance degradation occur if DNN works directly in the continuous learning (CL) context, defined as learning on a sequence of distinct tasks, including never-seen classes \cite{bib2,bib22}.} In contrast, human brains can grasp many skills consecutively \cite{4}. The discrepancy motivates researchers to emphasize CL techniques for learners, paving the road for \textcolor{black}{industrial lifelong model} deployment and artificial general intelligence (AGI) \cite{6}.


\textcolor{black}{The challenging class incremental learning (CIL) scenario in CL does not have a task identity given in the testing, while it is expected to infer predictions in the ${\cal Y}=\bigcup\nolimits_{i = 1}^T {{{\cal Y}_i}} $ space, requiring the learner to be adequate for all previous tasks. In general, learners face \textit{challenges} immeasurable to ERM \cite{6,7} in (1) \textit{non-i.i.d stream}: the continuum of tasks breaks i.i.d for $T$ tasks; (2) \textit{catastrophic forgetting}: the offsets of network optima and lacks of correct gradient regularization cause biased weights and ERM failure \cite{3}; (3) \textit{identity absence}: task index is not given in testing, and task order is unpredictable \cite{bib3}. The data from previous tasks are at most partially or not accessible, hindering ERM and misguiding the global gradient.}


Recent research on CIL problems falls into three categories: model-, replay-, and regularization-based methods \cite{bib3,bib4}. However, these approaches are based on some \textit{hypothetical settings} where: (1) \textcolor{black}{the} task amount $T$ is small and \textcolor{black}{sample data} $|{{\cal D}_i}|$ is large. \textcolor{black}{The non-immediate learner can cost multiple training epochs} on each task before responding \cite{7,9}; (2) the whole dataset of new tasks can be observed once tasks switch, i.e. task-wise locally i.i.d; (3) task boundary demarcating the contents of learning reveals the occasions to update and infer the learner. \textcolor{black}{In these settings}, pioneers such as EWC \cite{6} and GEM \cite{7} achieved good results. \textcolor{black}{Unfortunately, some conditions are rarely satisfied in actual real-time scenes such as robotics \cite{zhou2025vcsap}, industrial detection systems \cite{wang2024adaptive,fan2024learning,10654781}, and resource dispatches \cite{10643332}, which demand prompt learning, immediate decision-making, and fewer mistakes in data streams.} Although current CIL methods may manage partial \textit{challenges} well, \textcolor{black}{they require repetition training yet still suffer from learning dissipation on long task streams as in Fig. \ref{fig 0} (a)}, high complexity and storage consumption, and intractable parameter adjustments (see Section \ref{subsec4.5}).

To remove the above hypothesis and adapt to \textcolor{black}{more practical scenarios in which} task data is streamed online \textcolor{black}{without task-wise i.i.d. or boundary index}, our work focuses on the emerging online task-free CIL (OTCIL) scenario \cite{9, 10, 11}. OTCIL concept chart is shown in Fig. \ref{fig 0} (b). \textcolor{black}{It is expected to develop more human-like learners that continually evolve and respond to decisions immediately}, because the online learner should not keep waiting until the entire dataset associated with the current task is completed or the boundary is informed before updating itself \cite{8, 10}. Some classical methods like SI \cite{25} and Online EWC \cite{24} move steps forward to OTCIL, as they compute consolidation weights online and analyze batches one by one from local i.i.d tasks \cite{4, 7}. \textcolor{black}{However, they still need task boundaries and cause more regret, so they are not qualified in OTCIL \cite{10}.}

\textcolor{black}{To achieve favorable decisions about OTCIL tasks in hand, the learner is faced with tough} \textit{challenges} which are summarized as (1) \textit{task-wise non-i.i.d}: data comes as a non-stationary batch sequence where concept drifting occurs, instead of i.i.d being held within each task \cite{59}; (2) \textit{catastrophic forgetting}: the learner should balance stability and plasticity between old and new tasks well \cite{5}; (3) \textit{\textcolor{black}{delayed optimization}}: Most CIL learners are subjected to time-consuming training epochs and intractable hyper-parameters (HP) as updating relies on gradient descent (GD) of \textcolor{black}{non-convex changing ERM. This laborious process struggles to be completed rapidly} before the upcoming task for immediate response \cite{60}. Moreover, continual fitting the ERM leads to a significant decline in accuracy and decision quality on long task streams (e.g. Fig. \ref{fig 0} (a)); (4) \textit{resource consumption}: network volume and memory usage of some CIL methods gradually increase as tasks progress, resulting in subsequent slow learning; (5) \textit{\textcolor{black}{replay on long streams}}: \textcolor{black}{task amount $T$ is relatively large and sample amount $|{{\cal D}_i}|$ is small, while the data requires least replay for privacy protection}; (6) \textit{other expectations} consist of fast updating on new tasks, no growing structure, and no task boundary \cite{7}. Due to these harsh requirements, \textcolor{black}{previous DNN-based CIL methods were ineffective in OTCIL scenarios}. \textcolor{black}{Developing an OTCIL framework that attempts to tackle the above \textit{challenges} is still worthy of a study.}

Randomized neural networks (Randomized NN) present an alternative way of deep learning that attracted attention in large-scale computing due to their effectiveness and efficiency \cite{zhang2020new,xiao2024robust,61}. 
We employ the well-performed ensemble deep random vector functional link network (edRVFL) as the backbone to quickly learn representations, and promote it to the OTCIL scenario \cite{56,62}. Practically, the learner is trained on the current task while the next batch is possibly collected or labeled. This inspired us that, from the online learning perspective, if future unsupervised or reserved prior knowledge can be partially utilized to \textcolor{black}{prematurely rectify} the GD direction of the learner's trainable parameters, it can enhance precognition of future knowledge, which benefits predictions and further reduces learning regrets in OTCIL. 

Forward regularization (-F) can lead to better relative loss bounds than the canonical ridge (-R) in online convex optimization (OCO) \cite{63,64}. This paper proposes a more \textcolor{black}{general -F} with an adjustable $k$ factor to form the -$k$F style, hoping to adopt controllable unsupervised knowledge to improve posterior updates, reduce predictive loss, and tighten the regret boundary for the learner. The $k$ of -$k$F coordinates the intensity of the -F penalty, and -$k$F degenerates into -R when no knowledge is available. Furthermore, we propose -$k$F-Bayes style with $k$ synchronously self-adapted to overcome the complex distribution drifting and intractable tuning of $k$s by regarding the -$k$F incremental updates as a Bayesian learning process during OTCIL. Finally, we present the edRVFL-$k$F and ready-to-work edRVFL-$k$F-Bayes algorithms to respond above \textit{challenges}, which update incrementally, retain knowledge well \textcolor{black}{without replay}, have better immediate responses, and help the learners \textcolor{black}{escape} the dilemmas of plasticity and stability. \textcolor{black}{More excitingly, our algorithms can be made compatible with models with high transferability.}




\section{Related works}\label{sec2}
\textcolor{black}{Catastrophic forgetting is the core problem that needs to be addressed in (OT)CIL scenarios}, first noted in \cite{3}. Current approaches can be categorized into model-, replay-, and regularization-based methods. OTCIL is a more severe context of CIL, so some CIL methods can still be applied to OTCIL. Algorithms specifically designed for OTCIL are labeled by $^\dag $ in the following.

\textbf{Model-based methods.} \textcolor{black}{This branch of CIL algorithms is mainly based on parameter isolation (PI), knowledge distillation (KD), and architecture augmentation (AA) to enable learning across growing joint subspaces. HAT explicitly optimizes a binary mask to select task-specific nodes and parameters \cite{13}. PCL designs multi-head output for seen tasks and limits the drift of network weights on a new task \cite{17}. DYSON$^\dag $ learns new classes by aligning the feature space with geometry \cite{11}. PI is restricted by network capacity and is rarely used alone.} KD and AA require high computing resources for retraining, and the model scale increases with the task, which is suitable for large models but not for OTCIL without i.i.d and boundary premises. Our work adopts a non-growing or non-transferring model with low resource requirements for OTCIL.

\textbf{Replay-based method.} These CIL algorithms are distinguished by storing partial old representations or generative models to overcome forgetting by retraining while learning new tasks. GEM$^\dag $ uses old samples in the buffer pool to bootstrap the current gradient to escape loss increases \cite{7}. GSS$^\dag $ formulates sample selection as a constraint reduction problem in OTCIL \cite{9}. Although RanPAC is partially similar to ours in pre-trained models (PTM) and embedded random layers, \textcolor{black}{it is very different from the theoretical point of view \cite{19}. iCaRL performs KD on old and new training samples \cite{22}. TRAF alternates between standard and intraprocess updates, which requires large memory \cite{21}. Our work includes no previous replay or generators (pool), and allows used data to be discarded for privacy protection.}

\textbf{Regularization-based methods.} This direction is characterized by controlling gradients, which usually collect previous knowledge into explicit regularization terms. \textcolor{black}{A typical implementation is imposing penalties on targets to restrict important weights from wrong changes. The importance can be calculated using the Fisher information matrix (FIM), such as EWC \cite{6}.} Online EWC achieves cycles of active learning followed by consolidation in CIL \cite{24}. \textcolor{black}{SI approximates the importance of the parameter by its update length throughout the training trajectory \cite{25}.} MAS also adopts a similar approach \cite{26}. MAS$^\dag $ is improved to work in OTCIL with assumptions of task boundary and local i.i.d removed \cite{10}. \textcolor{black}{OWM updates the parameter in the orthogonal directions of the previous input space \cite{28}. The emerging CRNet applies the EWC principle to randomized NN and provides closed-form solutions that enable it to learn streaming data \cite{29}.} Our backbone is also a Randomized NN, and previous and future knowledge are all involved in the regularization terms, guiding the learning gradients in OTCIL.

Our work focuses on a general Randomized NN-$k$F framework in OTCIL of long task streams, realizing on-the-fly decision-making and better performance to -R with the help of self-adaptive forward knowledge intervention. It is also concerned with low updating dissipation and regrets, non-replay and non-growing architectures, and delivering competitive results together with \textcolor{black}{multiple PTMs} or fine-tuning techniques on hard image tasks.

\section{Preliminary}\label{sec3}
\subsection{CIL scenario}\label{subsec3.1}
The CIL environment can be described as learning a sequence of $Q$ tasks ${\cal T} = \{ {{\cal T}_1},{{\cal T}_2}, \cdots ,{{\cal T}_Q}\} $ ($Q \to \infty $ for \textcolor{black}{a} lifelong infinite stream). Each task ${{\cal T}_q} = ({{\cal X}_q},{{\cal Y}_q}) $ and ${{\cal T}_q} = \{ (x_q^i,y_q^i)_{i = 1}^{|{{\cal X}_q}|}|x_q^i \in {\mathbb{R}^{  s}},y_q^i \in {\mathbb{R}^{ {m_q}}},1 \le q \le Q\} $, where $q$ is task index, $|{{\cal X}_q}|$ denotes the total number of sample pair $(x_q^i,y_q^i)$ in ${{\cal T}_q}$, and $s$, $m_q=|{{\cal Y}_q}|$ are feature dimension and space dimension in ${{\cal T}_q}$ respectively. This explains the task-wise local i.i.d and task partition by boundary $q$. \textcolor{black}{The feature vector $x_q^i$ can denote a tabular example or an image representation.} Dataset $\cal T$ has a total $\sum\nolimits_{q = 1}^Q {|{{\cal X}_q}|} $ samples belonging to $m=|\bigcup\nolimits_{q = 1}^Q {{{\cal Y}_q}} | $ classes.

During training on ${{\cal T}_q}$ ($q>1$), an expected constraint is the unavailability of previous data $\bigcup\nolimits_{i = 1}^{q-1} {{{\cal T}_i}} $, i.e. $\bigcap\nolimits_{i = 1}^q {{{\cal T}_i}}  = \emptyset $, resulting in catastrophic forgetting on ${{\cal T}_q}$. \textcolor{black}{For CIL, the target (\ref{1}) is trying to continually minimize the learner ${f_q}(\theta )$'s ERM on the seen data while keeping the learned ${{\cal P}({{\cal Y}_{1..q-1}}|{{\cal X}_{1..q-1}})}$ stable.}
\begin{align}\label{1}
{\min_{\theta}}  {\cal L}({f_q}({{\cal X}_q};\theta ),{{\cal Y}_q}) + \mathbb{E}[{\cal L}({f_q}({{\cal X}_{1..(q - 1)}};\theta ),{{\cal Y}_{1..(q - 1)}})]
\end{align}
After training \textcolor{black}{on} ${{\cal T}_q}$, the learner ${f_{q + 1}}(\theta )$ is requested to enable predictions \textcolor{black}{in all classes encountered in} $\bigcup\nolimits_{i = 1}^q {{{\cal Y}_i}} $ without the identifier $1..q$. 
\textcolor{black}{The objective of (OT)CIL is to train a general learner $f_{Q+1}:{\cal X} \to {\cal Y}$ that can be incrementally updated, accumulate knowledge over time, resist past forgetting, and make fast decisions on the entire ${\cal T}$.}


\subsection{Offline edRVFL network}\label{subsec3.2}
\textcolor{black}{As a state-of-the-art Randomized NN, edRVFL stacks multiple RVFL sub-learners, reconnects the original input as residuals, and employs ensemble strategy to improve performance on hard tasks \cite{56}. We first introduce the offline edRVFL design, which will be qualified to work in OTCIL in Section \ref{sec4}.}

Given the non-continual edRVFL scheme with $L$ hidden layers, $N$ nodes per layer, the ${{\cal T}_q}$'s dataset $({{\cal X}_q},{{\cal Y}_q})$, and the projection ${f_{edRVFL}}:{{\cal X}_q} \to {{\cal Y}_q}$, the extracted feature $H$ of the $1$-st hidden layer is defined as:
\begin{equation} \label{2}
{H_{1,q}} = {g_1}({{\cal X}_q} \cdot {W_1}).
\end{equation}
For layer $1 < l \leqslant L$ it is defined as:
\begin{equation} \label{3}
{H_{l,q}} = {g_l}([{H_{l - 1,q}}|{{\cal X}_q}] \cdot {W_l}),
\end{equation}
where $g$ is an activation function (e.g. ReLU, Sigmoid, Swish), innate randomized weights ${W_1} \in {\mathbb{R}^{s \cdot N}}$ and ${W_l} \in {\mathbb{R}^{(s + N) \cdot N}}$, and layer feature ${H_{l,q}} \in {\mathbb{R}^{|{{\cal X}_q}| \cdot N}}$. Due to the differentiable convex target with respect to (w.r.t) learnable $\theta$, edRVFL is trained to the optimum in $L$ snapshots without GD therapy:
\begin{equation} \label{4}
{\theta _{l,q + 1}} = \arg {\min _{\theta_l} }\left\| {[{H_{l,q}}|{{\cal X}_q}] \cdot \theta_l  - {{\cal Y}_q}} \right\|_F^2 + {\lambda _l} \cdot \left\| \theta_l  \right\|_F^2,
\end{equation}
and the closed-form solution for future prediction:
\begin{equation} \label{5}
\begin{split}
primal&: {\theta _{l,q + 1}} = {(D_{l,q}^T{D_{l,q}} + {\lambda _l}I)^{ - 1}}D_{l,q}^T{{\cal Y}_q}\\
dual&: {\theta _{l,q + 1}} = D_{l,q}^T{({D_{l,q}}D_{l,q}^T + {\lambda _l}I)^{ - 1}}{{\cal Y}_q},
\end{split}
\end{equation}
where ${D_{l,q}} = [{H_{l,q}}|{{\cal X}_q}]$, ${\theta _{l,q+1}} \in {\mathbb{R}^{(s + N) \cdot |{{\cal Y}_q}|}}$ denotes the trainable output weight matrix. Final predictions can be obtained by \textcolor{black}{the} ensemble strategy, normally of aggregating averaged or median values of sub-learners for regression and \textcolor{black}{those} after SoftMax for classification tasks, which can be denoted as $En(\{ {D_{l,q}}{\theta _{l,q + 1}}\} _{l = 1}^L)$. For training, $\mathcal{O}({f_{edRVFL}})$ has minima of $\mathcal{O}(L \cdot {(N + s)^3})$ and ${\cal O}(L \cdot |{{\cal X}_q}{|^3})$ time complexity compared to $\inf \mathcal{O}({f_{MLP}}) = \mathcal{O}({T_{it}} \cdot L \cdot {N^2})$ with ${T_{it}}$ iterations every epoch. In fact, edRVFL always spends much less time as MLP needs multistep gradient computations for each layer. \textcolor{black}{edRVFL has been widely used in industry \cite{66p,67p}.}


\subsection{Bregman divergence and OCO problem}\label{subsec3.3}
We regard the OTCIL of edRVFL as a typical OCO problem, where -F can reduce regret compared to using -R \cite{63}. \textcolor{black}{Relative cumulative regret is defined as the cost difference between an online learner using the -F style and an online expert in hindsight}, which is defined as $\mathcal{R}^F = \sum\nolimits_{t} {} \mathcal{L}^F - \mathop {\min}\limits_{\theta  \in \Omega } \mathcal{L}^E(\theta )$ \cite{64}. Although recent papers studied multi-column -F, adaptive OCO, and discounted OCO theories, they have not focused on the more general -$k$F nor on adaptive -$k$F-Bayes, considering the appropriate penalty intensity and distribution drifting in OTCIL \cite{53p,54p,55p}.

We describe OCO's (OT)CIL process using Bregman divergence as it can represent penalties and result in general conclusions.
\begin{lemma}{\bf\cite{63}}\label{def 1} Bregman divergence is used to measure relative projection distance between HP sets or distributions. For a real-valued differentiable convex projection $G:\bm{\beta}  \in \bm{{\rm B}}  \to \mathbb{R}$, Bregman divergence $\Delta $ is defined as:
\begin{equation} \label{6}
{\Delta _G}( \bm{\tilde\beta} ,\bm{\beta} ): = G(\bm{\tilde \beta} ) - G(\bm{\beta} )- {(\bm{\tilde \beta}  - \bm{\beta} )^T}\cdot {\nabla \bm{_\beta} }G(\bm{\beta} ),
\end{equation}
where $\bm{\beta}$ is a vector, and ${\nabla \bm{_\beta }}$ denotes the gradient operator w.r.t $\bm{\beta} $.
\end{lemma}

Some properties of Bregman divergence are listed in \ref{poft1}. Offline learning can be described as follows:
\begin{lemma}{\bf\cite{63,64}}\label{lemma 2} Offline learning 
refers to the learning process of an expert \textcolor{black}{for} global tasks. Assume loss ${{\cal L} _q}$ and ${U_0}(\bm{\beta} )$ are differentiable and convex,  and there always exists a solution:
\begin{equation} \label{7}
{\bm{\beta} _{Q + 1}} = \arg {\min _{\bm{\beta}\in\bm{{\rm B}} }}{\text{}}{U_{Q + 1}}(\bm{\beta} ),
\end{equation}
where ${U_{Q + 1}}(\bm{\beta} ) = {\Delta _{{U_0}}}(\bm{\beta},{\bm{\beta} _0}) + {{\cal L} _{1..Q}}(\bm{\beta} )$, subscript 0 denotes prior distributions, and ${\bm{\beta} _{Q + 1}}$ represents the updated parameter after $Q$ tasks completed.
\end{lemma}

Function $U_0$ in the Bregman divergence can be a $l_2$ quadric form of Gaussian distribution: ${U_0}(\bm{\beta} ) = \frac{1}{2}{\bm{\beta} ^T}\eta _0^{ - 1}\bm{\beta} $ with symmetric positive definite matrix $\eta _0$ for example. (\ref{4}) shows the separate sub-learner working on one task, and (\ref{7}) represents the offline expert on the entire dataset $\cal T$.

\textcolor{black}{Our random network’s closed-form update ensures stable adaptation to new classes, reminiscent of the Neural Tangent Kernel (NTK) \cite{69p}, where wide networks evolve linearly in parameter space. Although our network has finite width, the NTK perspective provides a theoretical lens to interpret its incremental learning behavior \cite{70p, 71p}.}

\section{Methodology}\label{sec4}
We follow the OTCIL setup in \cite{10, 11}, where some restrictions break the original CIL assumptions. (1) Non-i.i.d data and concept \textcolor{black}{drift} in each task. The system observes some randomly drawn samples $(x_q^i,y_q^i)_{i = 1}^b$ in one batch. (2) No task boundary. The distribution ${{\cal P}_q}$ could itself undergo sudden or gradual shifts to ${{\cal P}_{q+1}}$ at any moment, and the system is unaware of when these \textcolor{black}{changes in distribution occur}. OTCIL is more prone to catastrophic forgetting. Our method targets settings where the total number of tasks is large, namely long task streams, while data arrive in batches. 
\subsection{Randomized NN with -$k$F for OTCIL}\label{subsec4.1}
As the learner is requested to comply with OTCIL setups, we formulate as the non-stationary data stream ${\cal T} = \{ ({X_t},{Y_t})_{t = 1}^T| {X_t} \in {\mathbb{R}^{b \cdot s}},{Y_t} \in {\mathbb{R}^{b \cdot m}},1 \le t \le T\} $, where each batch is isolated by unknown boundaries, $m$ can be the designed load of total classes $|{\cal Y}|$, $b$ is the batch size, and $b \cdot T =\sum\nolimits_1^Q |{{\cal X}_q}|$. This demands the model \textcolor{black}{to} learn quickly from the fed data. 

During OTCIL for $1 \le t \le T$, the $X_t$ is extracted to $\{ {D_{l,t}} = [{H_{l,t}}|{X_t}]\} _{l = 1,t = 1}^{L,T}$ using (\ref{2}) and (\ref{3}) \textcolor{black}{due to} randomized weights in $L$ sub-learners shared across tasks. With the edRVFL output regarded as multiple continuous sub-learners $\{ {D_{l,t}} \cdot \theta _{l,t+1} \to {Y_t}\} _{l = 1}^L$ over $T$ times, for the sake of simplicity, the following analysis is based on the feature stream $(D,Y) = \{ ({D_{l,t}},{Y_t})\} _{l = 1,t = 1}^{L,T}$.

\begin{figure}[!t]
\centering
\includegraphics[width=0.7\columnwidth,trim=6mm 6mm 7mm 7mm, clip]{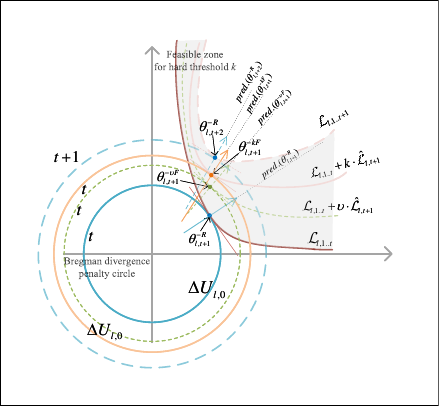}
  \caption{The description to one sub-learner using -R and -$k$F styles in OTCIL process. The concentric circles denote varied Bregman divergence penalties, and the curves denote losses on data. At time $t$, the learner using -R (blue solid circle) tries to match ${\cal L}_{l,1..t}$ (crimson solid curve) and evolve into $\theta_{l,t+1}^{-R}$ and predict. The learner using optimized -$k$F (yellow solid circle) tries to meet ${{\cal L}_{l,1..t}} + k \cdot {\hat {\cal L}_{l,t + 1}}$ (pink solid curve). The -$\upsilon $F ($\upsilon  \ne k$) is an instance of non-optimized -$k$F in $k$ feasible zone (gray area), shown in green dashed style. Then at time $t+1$, the same logic follows for $\theta_{l,t+2}^{-R}$. We recommend -$k$F as it reduces decision regret on new data compared to -R. $k$ should be carefully picked in the current feasible zone for better performance. -$k$F is with hard threshold $k$ and we introduce -$k$F-Bayes with soft adaptive ones later.}
  \label{fig penalty}
\end{figure}

\begin{theorem}\label{theorem 1} Incremental offline learning using -$k$F regularization is defined here. For $0 \leqslant t \leqslant T$, the following equations are optimized:
\begin{align}\label{8}
{\theta} _{t + 1} &= \arg {\min _{\theta} }{\text{}}U_{t + 1}({\theta} ) \\
U_{t + 1}({\theta} ) &= {\Delta _{{U_{0}}}}({\theta} ,{{\theta} _{0}}) + {{\cal L} _{1..t}}({\theta} ) + k \cdot {{\hat {\cal L} }_{t + 1}}({\theta} ),\notag
\end{align}
where $\hat {\cal L} $ is the estimated loss generated by current ${\theta} $ on upcoming data (the entire set is not necessary). 
\end{theorem}

The optimization target of -$k$F is defined as an extra adjustable regularization added to -R.
This shows that -$k$F adopts guidance from future unsupervised knowledge to incur loss and affects the gradients of optimization (also can be seen as a posterior optimum) when updating \textcolor{black}{the} present parameters.

Theorem \ref{theorem 1} repeatedly retrains on all data after observing $t$-th data as Lemma \ref{lemma 2} does. It is a full replay-based method that requires much retrieval and memory storage for cumulative data, resulting in retrospective retraining. We deduce the non-replay version of -$k$F for ensemble sub-learners.

\begin{theorem}\label{theorem 2} (OT)CIL framework of Randomized NN using -$k$F regularization. For $0 \leqslant t \leqslant T$ and $1 \le l \le L$, with previous ${{\cal L} _{l,1..t - 1}}$ preserved by Bregman divergence $\Delta {U_{l,t}}$, the framework is achieved by continuous optimization on data streams without replay.
\begin{align} 
\theta _{l,t + 1} = \arg {\min _{\theta_l} }{\text{}}{\Delta _{U_{l,t}}}(\theta_l ,\theta _{l,t}) + {{\cal L}_{l,t}}(\theta_l ) + k \cdot{\hat {\cal L}_{l,t + 1}}(\theta_l ) - k \cdot{\hat {\cal L}_{l,t}}(\theta_l ) \label{10}
\end{align}
\end{theorem}
\textbf{Proof}: See \ref{poft2}.

The proof also suggests \textcolor{black}{that} the effect of optimizing (\ref{10}) to update the Randomized NN-$k$F in (OT)CIL is equal to employing a synchronous offline expert (\ref{8}) using -$k$F regularization to be trained incrementally on \textcolor{black}{the} global data observed so far. (\ref{10}) can collect foregone knowledge in $\Delta _{U_{l,t}}$ without revisiting previous data, which shows \textcolor{black}{that} our methods are non-replay thoroughly.

\begin{remark}
\label{corollary 2}
There is no learning dissipation for the learner using -$k$F in (OT)CIL because its optimization remains the same as a simultaneous offline one, and non-replay is realized if $\Delta _{U_{l,t}}$ can be reasonably provided. 
\end{remark}

\textcolor{black}{As stated in \cite{65p} and Remark \ref{corollary 2}, the randomized feature of the past data still achieves completeness and universal approximation because the function clusters ${D_{l,t}}$ with smooth non-polynomial $g$ maintaining column-wise linear irrelevance in the Hilbert space for $0 \leqslant t \leqslant T$. On the basis of these features, -F achieves improved optima and less relative loss compared to -R by using novel regularization terms in OCO. The experiments in Section \ref{subsec5.4} will exhibit the above interactions.}

The estimated penalty on unlabeled data is expected to decorate the gradients and learning rates. We will show the details in Section \ref{subsec4.3}. The diagram for one sub-learner using -R and -$k$F strategies is shown in Fig. \ref{fig penalty}.

\subsection{edRVFL-$k$F algorithm}\label{subsec4.3}
\begin{figure*}[!t]
\centering
\includegraphics[width=\textwidth,trim=9.5mm 6mm 15mm 7mm, clip]{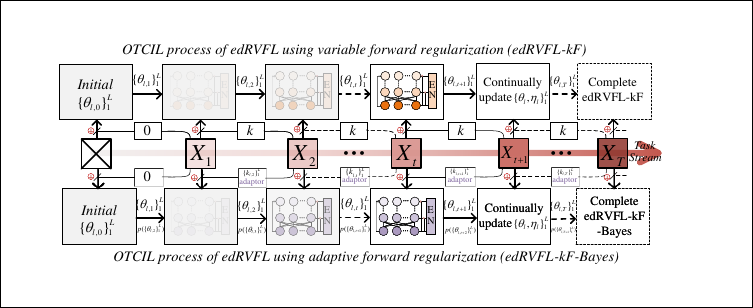}
  \caption{OTCIL processes of edRVFL-$k$F and edRVFL-$k$F-Bayes. Task stream over time is shown in a varied rufous arrow. Representations inside edRVFL multiple sub-learners are painted respectively in varied brightness yellow and purple to distinguish two algorithmic styles. Clustered sub-learners are progressively updated and uncertainty is gradually removed (shown by increasing sharpness) to present improving performance as batches come.\textcolor{red}{$\ \oplus $} denotes unsupervised forward knowledge participates in the current decision, where the $k_l$ is an adjustable constant in -$k$F and adaptive variable in -$k$F-Bayes.}
  \label{fig process}
\end{figure*}

We materialize the Theorem \ref{theorem 2} and derive the edRVFL-$k$F algorithm with the multi-dimensional Vovk-Azoury-Warmuth (VAW) form \cite{53p}.

\begin{theorem}\label{theorem 3} For the OTCIL process of edRVFL-$k$F, as shown in Fig. \ref{fig process}, step-wise updates of learnable weights follow:
\begin{align}\label{th3}
{\theta _{l,t + 1}}={\theta _{l,t}}- {\eta _{l,t + 1}}[(D_{l,t}^TD_{l,t}^{} + k D_{l,t + 1}^TD_{l,t + 1}^{} - k D_{l,t}^TD_{l,t}^{}){\theta _{l,t}} - D_{l,t}^T{Y_t}]
\end{align}
where $k$ is a constant, ${\theta} _{l,1}  = {\theta} _{l,0}=0$ for no prior knowledge, and variable learning rates ${\eta _{l,t + 1}} = {({({\eta _{l,0}})^{ - 1}} + \sum\nolimits_{i = 1}^t {D_{l,i}^T{D_{l,i}}}  + k \cdot D_{l,t + 1}^T{D_{l,t + 1}})^{ - 1}}$ with the initial value $\eta _{l,0} = {({\lambda _l} \cdot I)^{ - 1}}$.
The OTCIL process of edRVFL-$k$F is non-replay and non-growing in architecture and can respond to decisions quickly without forgetting.
\end{theorem}
Proof: See \ref{poft3}.

Detailed algorithmic procedures of Theorem \ref{theorem 3} can be found in \textbf{Algorithm 1}, and the process flow chart is shown in Fig. \ref{fig process}. \textcolor{black}{\textbf{Algorithm 1} can be configured to be directly applicable to tabular data or representations commonly used in industry.}

\textcolor{black}{The iterative weight updating mechanism could be exploited to fine-tune generative models with incremental data (for example, progressively improving deepfake synthesis). However, the reliance of our method on task-bound updates (see experiments in Section \ref{sec5}) inherently limits its direct applicability to generative tasks, as it optimizes for classification rather than data synthesis.}

\begin{algorithm}[ht]
\caption{edRVFL-$k$F for OTCIL.}\label{algorithm 1}
\begin{algorithmic}[1]
\Require data stream ${\cal T} = \{ ({X_t},{Y_t})_{t = 1}^T\} $, edRVFL $\{L, N, g_l,{\lambda _l}\} $, ${k_l}$
\Ensure $\{\theta _{l,T + 1}\}_{l = 1}^L$
\State {\textsc{Initialize}}: $\{ {W_l}\} _{l = 1}^L$, $\{ {\theta _{l,0}} = {\theta _{l,1}}=0\} _{l = 1}^L$, $\eta _{l,0} = {({\lambda _l} \cdot I)^{ - 1}}$.$\;$\footnotemark[1]
\State {\textsc{OTCIL Procedure:}}
\State {\textsc{For}} $1 \leqslant t \leqslant T$, do:$\#$Assume $X_{T+1}$ accessible.
\State \hspace{0.5cm}Observe $({X_{t +1},{Y_t})}$ $\#$${X_t}$ already in-hand.
\State \hspace{0.5cm}{\textsc{For}} $1 \leqslant l \leqslant L$, do:
\State \hspace{1.0cm}Compute ${D_{l,t+1}}$ by (\ref{2})-(\ref{3}), and guess with $\theta _{l,t}$. 
\State \hspace{1.0cm}{\textsc{If}} t==1: $\eta _{l,2}^\dag  = {\eta _{l,0}}$.$\;$\footnotemark[2]
\State \hspace{1.0cm}{\textsc{Else}}: Update $\eta _{l,t+1}^\dag = \eta _{l,t}^\dag- \eta _{l,t}^\dag D_{l,t}^T{(I + {D_{l,t}}\eta _{l,t}^\dag D_{l,t}^T)^{ - 1}}{D_{l,t}}\eta _{l,t}^\dag $ \footnotemark[3].
\State \hspace{1.0cm}Update $\eta _{l,t+1}$ again:
\Statex \hspace{1.5cm}${\eta _{l,t+1}} = \eta _{l,t+1}^\dag  - \eta _{l,t+1}^\dag {k_l}D_{l,t + 1}^T{(I + {k_l}{D_{l,t + 1}}\eta _{l,t+1}^\dag D_{l,t + 1}^T)^{ - 1}}{D_{l,t + 1}}\eta _{l,t+1}^\dag $
\State \hspace{1.0cm}Update $\theta _{l,t + 1}$:
\State \hspace{1.5cm}${\theta _{l,t + 1}} ={\theta _{l,t}} - {\eta _{l,t + 1}}[(D_{l,t}^TD_{l,t}^{} + k \cdot D_{l,t + 1}^TD_{l,t + 1}^{} - k \cdot D_{l,t}^TD_{l,t}^{}){\theta _{l,t}} - D_{l,t}^T{Y_t}] $
\State  \hspace{0.5cm}{\textsc{End For}} 
\State  \hspace{0.5cm}{\textsc{Response on Request:}} Let ${Y_{te}} \subseteq \bigcup\nolimits_1^t {{Y_i}} $
\State  \hspace{0.5cm}Calculate $\{ {D_{l,te}}\} _{l = 1}^L$ by (\ref{2})-(\ref{3}).
\State  \hspace{0.5cm}Decisions: $\widehat {{Y_{te}}}=softmax\{ {D_{l,te}}{\theta _{l,t + 1}}\} _{l = 1}^L$
\State  \hspace{0.5cm}Accuracy: $acc(En(\widehat {{Y_{te}}}),{Y_{te}})$
\State  {\textsc{End For}} 
\end{algorithmic}
\end{algorithm}
\footnotetext[1]{Usually set ${\lambda _1} = {\lambda _2} \cdots  = {\lambda _L}$ and ${k _1} = {k _2} \cdots  = {k _L}$ in sub-learners for simplicity (adaptive $k$ scheme in Section \ref{subsec4.4}).}
\footnotetext[2]{Here we consider a severe but more practical condition: the user is completely unaware of any prior knowledge. \textcolor{black}{So,} $\eta _{l,2}^\dag$ is set to initial ${\eta _{l,0}}$ instead of ${({({\eta _{l,0}})^{ - 1}} + D_{l,1}^T{D_{l,1}})^{ - 1}}$.}
\footnotetext[3]{A pseudo incomplete $\eta_{l,t+1}$, marked by $\eta _{l,t+1}^\dag$, is maintained as an intermediate variable for computation convenience.}

\begin{corollary}
\label{corollary 3}
The -$k$F style is more general. If $k=0$, Theorem \ref{theorem 3} degenerates to -R style:
\begin{align}\label{140}
{\theta} _{l,t + 1} &= \arg {\min _{{\theta}_l} }{\Delta _{U_{l,t}}}({\theta_l} ,{\theta} _{l,t}) + {{\cal L}_{l,t}}({\theta_l} )\\
{{\theta} _{l,t + 1}} &={{\theta} _{l,t}} - {{\eta} _{l,t + 1}^{}}[{D}_{l,t}^{T}{D}_{l,t} {{\theta} _{l,t}} - {D}_{l,t}^{T}{Y_t}] \notag
\end{align}
If $k=1$, Theorem \ref{theorem 3} degenerates to -F style:
\begin{align}\label{150}
{\theta} _{l,t + 1} &= \arg {\min _{{\theta_l}} }{\Delta _{U_{l,t}}}({\theta_l} ,{\theta} _{l,t}) + {{\cal L}_{l,t}}({\theta_l} )+ {\hat {\cal L}_{l,t + 1}}({\theta_l} )- {\hat {\cal L}_{l,t}}({\theta_l} )\\
{{\theta} _{l,t + 1}} &={{\theta} _{l,t}} - {{\eta} _{l,t + 1}^{}}[{D}_{l,t+1}^{T}{D}_{l,t+1} {{\theta} _{l,t}} - {D}_{l,t}^{T}{Y_t}] \notag
\end{align}
This further suggests that the performance of using -$k$F must not be inferior to that of -R or -F if $k$ can be properly designed. Such $k$ values exist with high probability in practice.
\end{corollary}

For the classification of offline learning in Lemma \ref{lemma 2}, penalty factors of diagonal ${\eta _{l,0}}$ come partially from the secondary moment of Bayesian prior distributions and are usually set to the same constant in $L$ sub-learners. This defaults to the learnable weight components being priorly identically distributed in any sub-learner, regardless of what the feature distribution is or whether the correlations in the weight matrix should be considered.

However, not as expected in Lemma \ref{lemma 2} \textcolor{black}{where the entire dataset simultaneously participates in training, the OTCIL process emphasizes CL where separate task data come in batches and the previous is invisible}, exacerbates distribution drifting and transition.
\begin{remark} 
\label{corollary 4}
It is unreasonable to keep the penalty factors $k_l$ fixed in Theorem \ref{theorem 2} for $0 \leqslant t \leqslant T$ and $1 \leqslant l \leqslant L$. The OTCIL process demands the $k_l$ \textcolor{black}{change properly} to accommodate variations in feature distributions in time and structural directions. 
\end{remark}

\textcolor{black}{From the time point of view}, the Bayesian prior distribution of the weight matrix in each sub-learner changes with the learning progress in (\ref{17}), where the covariance matrix is symmetric full instead of static diagonal. It is also affected by task batch capacity, varying non-i.i.d distribution, accumulated data volume, etc. From the network architecture \textcolor{black}{point of view,} $k_l$ remaining constant in sub-learners is unfair because feature distributions vary \textcolor{black}{between layers}. The above facts result in skewing and imbalance in the actual severity of penalties, and impact learning gradients. Unfortunately, optimizing $k_l$ is unrealistic in real-time or based on all observed data, while how to adjust $k_l$ is critical to improving the -$k$F style.

\subsection{Determination of $k_t$ for learners}\label{subsec4.4}
It is obvious from Theorem \ref{theorem 2} and (\ref{17}) that the choice of $k$ is critical, as it determines how much future unsupervised knowledge is involved in each sub-learner. When $k= 0$, the -$k$ F algorithm degrades to -R style in Corollary \ref{corollary 3}. \textcolor{black}{When $0 < k \le 1$, it can be regarded as a partial forward knowledge intervention because the} ERM on $D_{l,t}^{}$ is larger than the forward penalty of $D_{l,t + 1}^{}$. Having $k>1$ leads to excess in forward regularization. 

However, the unchanged $k$ alone cannot sufficiently control the degree of forward knowledge involvement in one sub-learner as the penalty strength on different tasks changes. 
$k$ should not be set as a constant because the prior distribution of learnable weights keeps changing as OTCIL progresses rather than staying the same as offline learning, and the covariance matrix is no longer diagonal. Remark \ref{corollary 4} also suggests \textcolor{black}{that} many facts disturb $k$. To avoid regularization instability changes in OTCIL and rely on costly searching, we propose the edRVFL-$k$F-Bayes with a self-adaptive strategy of penalty coefficient $k_{l,t}$ to maintain the balance of penalties in the statistical models.

From the Bayesian perspective, (\ref{4}) and (\ref{7}) can be represented as:
\begin{align}\label{23}
- \ln p(\theta |{D_q},{{\cal Y}_q}) \propto  - \ln[p({{\cal Y}_q}|{D_q},\theta )p(\theta )] \propto ||{D_q}\theta  - {{\cal Y}_q}||_F^2 + \frac{{\varepsilon _0^2}}{{\varepsilon _1^2}}||\theta ||_F^2
\end{align}
where ${\varepsilon _0^{2}}$ and ${\varepsilon _1^{2}}$ are the variance of the residual error and column of the weight matrix, respectively. In fact, (\ref{23}) implies some assumptions: (1) ${d^i}\theta  - {{ y}^i} \sim {\cal N}(0,\varepsilon _0^2)$ for each sample because of noise i.i.d; (2) it assumes no correlation among weight elements in each $\theta$'s vector and ${\theta} \sim {\cal N}(0,diag(\varepsilon _1^2,...\varepsilon _1^2))$, although they maybe non-i.i.d; (3) it independently analyzes each element of $\theta$ matrix. This paradigm is sound, as the model is trained on the whole dataset. 
Unfortunately, weight distribution $p(\theta)$ changes dynamically at different periods of OTCIL, which can be described by Bayesian learning:
\begin{align}\label{24}
p({\theta _{t + 1}}|{{\cal Y}_1},...{{\cal Y}_t}) \propto \arg {\max _{{\theta}\in{{\theta} _{t + 1}}} }p({{\cal Y}_t}|{\theta _{}},{{\cal Y}_1},...{{\cal Y}_{t - 1}}) \cdot p({\theta _{}}|{{\cal Y}_1},...{{\cal Y}_{t - 1}})
\end{align}
where the last term denotes a prior distribution concerning the current HP. The prediction loss of posterior distribution on two adjacent batches of data is a decreasing trend, which inhibits the growth of regrets. Our OTCIL process using -$k$F can be regarded as a regularization-based method that constantly modifies the posterior distribution $p({\theta _{t + 1}})$ based on available information to minimize ERM without rehearsal.
This is reflected in the Bregman divergence ${\Delta _{{U_{l,t}}}}({\theta _l},{\theta _{l,t}})$ in Theorem \ref{theorem 2}.  

We first improve Theorem \ref{theorem 2} to the following:
\begin{theorem}\label{theorem 4} To enable the -$k$F to be self-adaptive in OTCIL, the rigid coefficient $k$ in the recursive optimization target (\ref{10}) is replaced with soft variables. Let $U_{l,t + 1}({\theta_l} ) = {\Delta _{{U_{l,0}}}}({\theta_l},{{\theta} _{l,0}})+ {{\cal L} _{l,1..t}}({\theta_l} ) + {k_{l,t+1}} \cdot {{\hat {\cal L} }_{l,t + 1}}({\theta_l} )$, the OTCIL processes can be modeled as follows:
\begin{align} 
\theta _{l,t + 1} &= \arg {\min _{\theta_l} }{\text{}}{\Delta _{U_{l,t}}}(\theta_l ,\theta _{l,t}) + {{\cal L}_{l,t}}(\theta_l )+ {k_{l,t + 1}} {\hat {\cal L}_{l,t + 1}}(\theta_l ) - {k_{l,t }} {\hat {\cal L}_{l,t}}(\theta_l )\label{25}
\end{align}
\end{theorem}
Similarly following the steps in Theorem \ref{theorem 3}, we can obtain:
\begin{align} \label{26}
\theta _{l,t + 1} &= \arg {\min _{\theta_l} }\sum\nolimits_{i = 1}^m {\frac{1}{2}{{({\bm{\theta} _{{l_i}}} - {\bm{\theta} _{{l_i},t}})}^T}[{{({\eta _{l,0}})}^{ - 1}} }\notag\\
&+{ \sum\nolimits_{j = 1}^{t - 1} {D_{l,j}^T{D_{l,j}}}  +{  {k_{l,t }} \cdot D_{l,t}^T{D_{l,t}}}]({\bm{\theta} _{{l_i}}} - {\bm{\theta} _{{l_i},t}})} \notag\\
&+\frac{1}{2}||{D_{l,t}}{\theta _l} - {Y_t}||_F^2-\frac{k_{l,t }}{2}||{D_{l,t }}({\theta _l} - {\theta _{l,0}})||_F^2\notag\\
&+\frac{k_{l,t + 1}}{2}||{D_{l,t + 1}}({\theta _l}- {\theta _{l,0}})||_F^2 
\end{align}
where ${({\eta _{l,t}})^{ - 1}}={{({\eta _{l,0}})}^{ - 1}} + \sum\nolimits_{i = 1}^{t - 1} {D_{l,i}^T{D_{l,i}}}  +{  {k_{l,t }} \cdot D_{l,t}^T{D_{l,t}}}$. The recursive updating policy between $\theta _{l,t}$ and $\theta _{l,t+1}$ can be given correspondingly as follows:
\begin{align} \label{27}
{\theta _{l,t + 1}} ={\theta _{l,t}} - {\eta _{l,t + 1}}[({k_{l,t + 1}}  D_{l,t + 1}^TD_{l,t + 1}+(1- {k_{l,t }}) D_{l,t}^TD_{l,t}^{}){\theta _{l,t}} - D_{l,t}^T{Y_t}]  
\end{align}
where ${({\eta _{l,t+1}})^{ - 1}}={{({\eta _{l,0}})}^{ - 1}} + \sum\nolimits_{i = 1}^{t} {D_{l,i}^T{D_{l,i}}}+{  {k_{l,t+1 }}  D_{l,t+1}^T{D_{l,t+1}}}$.

\textcolor{black}{For computational convenience, our algorithm assumes column-wise independence in ${\theta _l}$ and sets the shared variance ${\eta _l}$ within the columns.} 
By redefining \textcolor{black}{the} Bregman projection, one can also set $\{ {\eta _{l,i,0}}\} _{i = 1}^m \in {\mathbb{R}^{m \cdot {{(s + N)}^2}}}$ or ${\eta _{l,0}} \in {\mathbb{R}^{{{[m \cdot (s + N)]}^2}}}$ for joint correlation modeling of \textcolor{black}{the} Bayesian prior distribution.

Here we still maintain the previous setup and analyze the distribution model column-wise $\theta _{l,t}$ in OTCIL. According to the a posterior computation in (\ref{26}), \textcolor{black}{the Bayesian prior distribution for the current batch has changed to ${ p}({\theta _{{l_i},t + 1}}) \sim {\cal N}({\theta _{{l_i},t}},{\eta _{l,t }})$, so that the estimated distribution} $p(D_{l,t + 1}^{}{\theta _{{l_i},t + 1}}) \sim {\cal N}(D_{l,t + 1}^{}{\theta _{{l_i},t}},D_{l,t + 1}^{}{\eta _{l,t }}D_{l,t + 1}^T)$ and $p(D_{l,t}^{}{\theta _{{l_i},t + 1}}) \sim {\cal N}(D_{l,t}^{}{\theta _{{l_i},t}},D_{l,t}^{}{\eta _{l,t}}D_{l,t}^T)$. This is because the last two terms can be regarded as \textcolor{black}{a Bayesian prior distribution w.r.t.} $D_{l,t+1}^{}{\theta _{{l},t + 1}}$ and $D_{l,t}^{}{\theta _{{l},t + 1}}$ respectively. The ${k_{l,t + 1}}$ and ${k_{l,t }}$ can be set to:
\begin{align} \label{28}
{k_{l,t + 1}}&={(\frac{{Tr[{{(D_{l,t+1}^{}{\eta _{l,t}}D_{l,t+1}^T)}^{ - 1}}]}}{b})^{ - 1}}\notag\\
{k_{l,t }}&={(\frac{{Tr[{{(D_{l,t}^{}{\eta _{l,t}}D_{l,t}^T)}^{ - 1}}]}}{b})^{ - 1}}
\end{align}

${k_{l,t}} = {(random\_pick{(D_{l,t}^{}{\eta _{l,t}}D_{l,t}^T)^{ - 1}})^{ - 1}}$ or ${k_{l,t}} = Tr[(D_{l,t}^{}{\eta _{l,t}}D_{l,t}^T)]/b$ could be another two choices for quick computation.
To avoid the ill-posed problem and allow manual adjustment, (\ref{28}) is calculated by Moore-Penrose as follows:
\begin{align} \label{29}
{k_{l,t + 1}}&=\kappa \cdot {(\frac{{Tr[{{(D_{l,t+1}^{}{\eta _{l,t}}D_{l,t+1}^T+\sigma I)}^{ - 1}}]}}{b})^{ - 1}}\notag\\
{k_{l,t }}&=\kappa \cdot {(\frac{{Tr[{{(D_{l,t}^{}{\eta _{l,t}}D_{l,t}^T+\sigma I)}^{ - 1}}]}}{b})^{ - 1}}
\end{align}
where $\sigma $ is a small positive number (e.g. ${10^{ - 5}}$) and $\kappa $ is a positive constant (e.g. 1.0). edRVFL-$k$F-Bayes algorithm can be found in $\textbf{Algorithm 2}$, and the process \textcolor{black}{flow} chart is shown in Fig. \ref{fig process}. \textcolor{black}{\textbf{Algorithm 2} can also be applied to tabular data or other representations in the industry, and $k$ \textcolor{black}{no longer has to} be manually set. The new HP (i.e., $\kappa$, $\sigma$) are less sensitive and affect performance only to a limited extent, so this algorithm can be used in a \textit{plug-and-play} fashion.}

\begin{algorithm}[t]
\caption{edRVFL-$k$F-Bayes for OTCIL.}\label{algorithm 2}
\begin{algorithmic}[1]
\Require data stream ${\cal T} = \{ ({X_t},{Y_t})_{t = 1}^T\} $, edRVFL $\{L, N, g_l,{\lambda _l}\} $
\Ensure $\{\theta _{l,T + 1}\}_{l = 1}^L$
\State {\textsc{Initialize}}: $\{ {W_l}\} _{l = 1}^L$, $\{ {\theta _{l,0}} = {\theta _{l,1}}=0\} _{l = 1}^L$, $\eta _{l,0} = {({\lambda _l} \cdot I)^{ - 1}}$.
\State {\textsc{OTCIL Procedure:}}
\State {\textsc{For}} $1 \leqslant t \leqslant T$, do:$\#$Assume $X_{T+1}$ accessible.
\State \hspace{0.5cm}Observe $({X_{t +1},{Y_t})}$ $\#$${X_t}$ already in-hand.
\State \hspace{0.5cm}{\textsc{For}} $1 \leqslant l \leqslant L$, do:
\State \hspace{1.0cm}Compute ${D_{l,t+1}}$ by (\ref{2})-(\ref{3}), and guess with $\theta _{l,t}$.
\State \hspace{1.0cm}{\textsc{If}} t==1: 
\State \hspace{1.5cm}$\eta _{l,1}^\dag  = {\eta _{l,0}}$.
\State \hspace{1.5cm}Get $k_{l,t+1}(\eta _{l,1}^\dag)$ and $k_{l,t}(\eta _{l,1}^\dag)$ by (\ref{29})

\State \hspace{1.0cm}{\textsc{Else}}: 
\State \hspace{1.5cm}Update $\eta _{l,t}^\dag = \eta _{l,t-1}^\dag- \eta _{l,t-1}^\dag D_{l,t}^T{(I + {D_{l,t}}\eta _{l,t-1}^\dag D_{l,t}^T)^{ - 1}}{D_{l,t}}\eta _{l,t-1}^\dag $.
\State \hspace{1.5cm}Get $k_{l,t+1}(\eta _{l,t})$ and $k_{l,t}(\eta _{l,t})$ by (\ref{29})
\State \hspace{1.0cm}Update $\eta _{l,t+1}$:
\Statex \hspace{1.5cm}${\eta _{l,t+1}} = \eta _{l,t}^\dag  - \eta _{l,t}^\dag {k_{l,t+1}}D_{l,t + 1}^T{(I + {k_{l,t+1}}{D_{l,t + 1}}\eta _{l,t}^\dag D_{l,t + 1}^T)^{ - 1}}{D_{l,t + 1}}\eta _{l,t}^\dag $
\State \hspace{1.0cm}Update $\theta _{l,t + 1}$:
\Statex \hspace{1.5cm}${\theta _{l,t + 1}} ={\theta _{l,t}} - {\eta _{l,t + 1}}[({k_{l,t + 1}}  D_{l,t + 1}^TD_{l,t + 1}+(1- {k_{l,t }}) D_{l,t}^TD_{l,t}^{}){\theta _{l,t}} - D_{l,t}^T{Y_t}]$
\State  \hspace{0.5cm}{\textsc{End For}} 
\State  \hspace{0.5cm}{\textsc{Response on Request:}} Let ${Y_{te}} \subseteq \bigcup\nolimits_1^t {{Y_i}} $
\State  \hspace{0.5cm}Calculate $\{ {D_{l,te}}\} _{l = 1}^L$ by (\ref{2})-(\ref{3}).
\State  \hspace{0.5cm}Decisions: $\widehat {{Y_{te}}}=softmax\{ {D_{l,te}}{\theta _{l,t + 1}}\} _{l = 1}^L$
\State  \hspace{0.5cm}Accuracy: $acc(En(\widehat {{Y_{te}}}),{Y_{te}})$
\State  {\textsc{End For}} 
\end{algorithmic}
\end{algorithm}

\subsection{Comparisons with CRNet}\label{subsec4.5}
The novel CRNet proposed in \cite{29} is a strong competitor to ours, which is based on the EWC principle and protects the performance of previous tasks by constraining current optimized weights within a low-error region using consolidation factors in Randomized NN. Specifically, it adopts \textcolor{black}{the} Laplacian approximation to match a posterior distribution $p({\theta _{t + 1}}|{{\cal Y}_1},...{{\cal Y}_{t - 1}})$ (assumed by \textcolor{black}{the} Gaussian distribution) at the latest optimized weights. In this way, Gaussian covariance of the regularization term is estimated by:
\begin{align} \label{FIM}
\mathbb{E}({\sigma ^{ - 2}}) =  - \mathbb{E}(\nabla _\theta ^2\log p(\theta |{{\cal Y}_{t - 1}})) = \mathbb{E}({\nabla _\theta }\log p{(\theta |{{\cal Y}_{t - 1}})^2})
\end{align}
and consolidation factors are given by the empirical Fisher information matrix (EFIM) finally.

Although the CRNet is also based on Randomized NN and updated one-shot using closed-form solution, it has some obvious differentiae compared to our methods such as: (1) only for CIL scenario with task-wise i.i.d. and boundary but not working in OTCIL; (2) trade-off $\lambda _q$ measuring task relative importance greatly affects performance, while properly setting them is hard, considering of large $Q$ and other HP; (3) although it uses closed-form updates, the numerical fitting loss of a posterior $p({\theta}|{\cal Y})$ with EFIM and its Taylor series exists, as shown in Fig. \ref{fig 0}, which increases as more tasks come. Our proposed methods focus on more practical OTCIL with long task stream and batch non-i.i.d premises, enable crucial adaptive factors, and maintain a posterior $p({\theta _{t + 1}}|{{\cal Y}_1},...{{\cal Y}_{t - 1}})$ in Bregman divergence.

\section{Experiments}\label{sec5}
This section documents experimental results on public image datasets. We used multiple measurements to assess the static and dynamic performance of our methods and compare them with SOTA baselines. We also explored the efficacy of edRVFL-$k$F and edRVFL-$k$F-Bayes in different scenarios and further studied ablation testing.

\subsection{Implementation details}\label{subsec5.0}
(1) \textbf{Evaluation metrics}:
\textcolor{black}{An in-depth analysis of the experimental results was performed by introducing six metrics that characterized the accuracy of the immediate test set, the accuracy of incremental tasks, the ability to retain knowledge, the degree of knowledge loss, the immediate regret and the Kullback-Leibler divergence (KL).}

\textit{Average task accuracy} ($ACC$) is defined in CL literature as the average accuracy of all previously learned tasks.
\begin{align} \label{30}
ACC = \frac{1}{{|Q|}}\sum\nolimits_{q = 1}^Q {{R_{Q,q}}} 
\end{align}
where ${R_{Q,q}}$ is the classification accuracy of the learner on task $q$ after learning on task $Q (Q \ge q)$. It reflects the task-wise accuracy variation in (OT)CIL processes.

\textit{Backward transfer} ($BWT$) is defined as the average difference between the accuracy of all task completions and the first learning of one task:
\begin{align} \label{31}
BWT = \frac{1}{{|Q| - 1}}\sum\nolimits_{q = 1}^{Q - 1} {{R_{Q,q}} - {R_{q,q}}} 
\end{align}
BWT indicates the knowledge retention ability of the algorithm \textcolor{black}{when} larger values are desired.

\textit{Forward transfer} ($FWT$) is defined as the average difference between the accuracy of first learning of one task and using an independent expert on it:
\begin{align} \label{32}
FWT = \frac{1}{{|Q| - 1}}\sum\nolimits_{q = 2}^{Q} {{R_{q,q}}-{R_{q}^{ind}}} 
\end{align}
where ${R_{q}^{ind}}$ denotes the testing accuracy of an independent expert trained only on task $q$. Higher FWT indicates \textcolor{black}{that} the learner can acquire more knowledge from newly seen tasks.

\textit{Immediate accuracy} ($acc.(t)$) denotes the immediate testing accuracy on the entire testset after the learner finishes $t-th$ learning on ${\cal T}$ in OTCIL.
\begin{align} \label{33}
acc.(t)={R_{t,1..Q}}\;\;\;\;1 \le t \le T
\end{align}
It is used to study the dynamic learning performance of using -$k$F and -$k$F-Bayes styles precisely. 

\textit{Immediate regret} ($regret(t)$) is a common metric in online stream learning and is defined as the real-time cost incurred by the learner: 
\begin{align} \label{34}
regre{t}(t) = ||\frac{{\sum\nolimits_{l = 1}^L {soft\max({D_{l,te}}{\theta _{l,t}})}  - L \cdot {Y_{te}}}}{{L \cdot |{{\cal X}_{te}}|}}||_F^2
\end{align}
\textcolor{black}{where $1 \le t \le T$ and the subscript denotes Frobenius norm. We also offer a cumulative regret to describe the total loss incurred on the testset in the OTCIL process.}

\textit{Immediate KL} ($KL(t)$) is another metric for measuring the learner's loss in OTCIL from the perspective of polynomial distribution.
\begin{align} \label{35}
KL(t) = \frac{{\sum\nolimits_m {{Y_{te}} \odot \ln(\frac{{L \cdot {Y_{te}}}}{{\sum\nolimits_{l = 1}^L {soft\max({D_{l,te}}{\theta _{l,t}})} }})} }}{{|{{\cal X}_{te}}|}} 
\end{align}
where $1 \le t \le T$. Algorithms that incurred lower losses in OTCIL are thought to perform better. The two indices are used to study the dynamic and entire performance in OTCIL.

(2) \textbf{Dataset}:
To verify the proposed framework of edRVFL-$k$F and study the efficacy of edRVFL-$k$F-Bayes in OTCIL scenarios, we conducted experiments on 
Fashion-MNIST\footnotemark[4] and CIFAR-100\footnotemark[5]. The terminology [DATASET]-$|\cal Y|/Q$ denotes the $\cal T$ with $|\cal Y|$ classes is divided into $Q$ tasks, namely $m/Q$ classes in each task \cite{29}. In addition, $\cal T$ could be split into batches to learn as our proposed methods worked without task boundaries. Data chunks of tasks were sequentially fed to the learner and the previous revisit was not allowed except for replay-based methods.

Fewer tasks, such as FashionMNIST-10/5 and CIFAR-100/5, tended to examine the learner's ability to learn difficult tasks continually, while long task streams paid attention to knowledge retention like FashionMNIST-10/10 and CIFAR-100/100. Implementation details are recorded in subsections.
\footnotetext[4]{https://www.kaggle.com/datasets/zalando-research/fashionmnist}
\footnotetext[5]{https://www.cs.toronto.edu/$\sim$kriz/cifar.html}

(3) \textbf{Compared methods}: The compared methods included model-, replay- and regularization-based approaches. In each experimental part, classical and latest methods, such as EWC (2017), CRNet (2023), and NICE (2024), would be involved. We also added specially designed methods for OTCIL, such as GEM (2017) and DYSON (2024). We used official open-source codes or popular third-party codes for all methods. 

\subsection{FashionMNIST dataset}\label{subsec5.2}
For the FashionMNIST-10/5, each task containing 2 classes of approximate difficulty was sequentially learned for CIL methods, and \textcolor{black}{the number} of tasks would be extended to more for OTCIL. We carried out experiments on a wide range of methods, including classical (e.g. EWC \cite{6}, MAS \cite{10}, PCL \cite{17}, GEM \cite{7}) and SOTA algorithms (e.g. CRNet \cite{29}, RanPAC \cite{19}, NICE \cite{50}, CGAN+ResNet-50 \cite{41}). The backbone architectures and HP selection followed the setups in \cite{29}, or adhered to open source codes in the original papers (see the footnotes in Table \ref{tab5} for details). The task order was randomly sorted and baselines were tested 10 times for consistent results. Only the current task/batch data were present, and the task boundary was merely revealed to the CIL methods. Our methods still learned tasks using one-pass and without revisits, and had no prior update before incoming data. 

\begin{table}[!htbp]

\caption{Testset accuracy comparisons of CIL algorithms on FashionMNIST}\label{tab5}
\begin{adjustbox}{width=1.0\textwidth}
\centering
\begin{tabular}{lccccc}
\hline%
& & &\multicolumn{3}{@{}c@{}}{Metric} \\
\cmidrule{4-6}
Category & Algorithm &Task batch sum.& $ACC\pm std.$ & $BWT\pm std.$ & $FWT\pm std.$  \\
\midrule
\multirow{3}{*}{Non-CL}&offline edRVFL\cite{56}&1&0.9876$\pm$1.09e-4&-&- \\
&separate edRVFL&1&0.9899$\pm$0.0043&-&- \\
&fine-tuning edRVFL&1&0.2655$\pm$0.0202&-&- \\
&non-incremental edRVFL&1&0.1974$\pm$0.0030&-&- \\
\midrule
\multirow{3}{*}{Regularization}&EWC$^*$\cite{6}&5&0.4021$\pm$0.0422&-0.6550$\pm$0.0798&-0.0596$\pm$0.0296\\
&SI$^*$\cite{25}&5&0.4151$\pm$0.0424&-0.6525$\pm$0.0608&-0.0379$\pm$0.0329\\
&\multirow{2}{*}{MAS$^*$$^\dag $\cite{10}}&5&0.3883$\pm$0.0258&-0.7077±0.1074&-0.0893±0.0395\\
&&10&0.3564$\pm$0.0237&-0.7235±0.0857&-0.1023±0.0284\\
&Online EWC$^*$\cite{24}&5&0.4535$\pm$0.0706&-0.5808±0.0811&-0.0592±0.0438\\
&DMC$^*$\cite{42}&5&0.7081$\pm$0.0229&-0.3327±0.3748&-0.2793±0.0478\\
&OWM$^*$\cite{28}&5&0.7931$\pm$0.0103&-0.1791±0.0230&-0.0670±0.0119\\
&CRNet-I$^*$$^\diamondsuit$\cite{29}&5&0.9472$\pm$0.0075&-0.0328±0.0086&-0.0184±0.0025\\
&CRNet-II$^*$\cite{29}&5&0.9205$\pm$0.0054&-0.0534±0.0071&-0.0299±0.0034\\
&IF2NET$^\diamondsuit$\cite{43}&5&0.9501$\pm$0.0047&-0.0189±0.0074&-0.0267±0.0112\\
\midrule
\multirow{3}{*}{Model}&EFT$^*$\cite{44}&5&0.7826±0.0205&-0.1571±0.0624&-0.0598±0.0234\\
&PCL$^*$\cite{17}&5&0.8327±0.0049&-0.1258±0.0301&-0.0681±0.0131\\
&PathNet$^*$\cite{46}&5&0.6957±0.0482&-&-\\
&HAT$^*$\cite{13}&5&0.7092±0.0184&-&-\\
&CAT$^*$\cite{45}&5&0.7774±0.0410&-&-\\
&FS-DGPM\cite{47}&5&0.8245±0.0055&-0.1471±0.0237&-0.1340±0.0220\\
&RanPAC\cite{19}&5&0.8745±0.0353&-0.0526±0.0245&-0.0423±0.0065\\
&RanDumb$^\diamondsuit$\cite{48}&5&0.9364±0.0130&-0.0456±0.0083&-0.0303±0.0052\\
&NICE\cite{50}&5&0.7538±0.0263&-0.3735±0.0412&-0.0324±0.0185\\
&\multirow{2}{*}{DYSON$^\dag$\cite{11}}&5&0.8333±0.0192&-0.1300±0.0217&-0.0196±0.0211\\
&&10&0.7950±0.0232&-0.1547±0.0225&-0.0357±0.0228\\
&AUTOACTIVATOR$^\diamondsuit$\cite{49}&5&0.8846±0.0006&$\overline{-0.0850}\pm\underline{0.0750}$&-\\
\midrule
\multirow{3}{*}{Replay}&RPS-Net$^*$\cite{51}&5&0.7984±0.0192&-0.0228±0.0131&-0.0326±0.0412\\
&\multirow{2}{*}{GEM$^*$$^\dag $\cite{7}}&5&0.8282±0.0107&-0.0689±0.0392&-0.0915±0.0399\\
&&10&0.7526±0.0152&-0.0935±0.0238&-0.1103±0.0343\\
&LOGD$^*$\cite{52}&5&0.8316±0.0191&-0.1001±0.0077&-0.0546±0.0215\\
&IL2M$^*$\cite{53}&5&0.8687±0.0235&-0.0604±0.0313&-0.0166±0.0028\\
&\multirow{2}{*}{GSS$^\dag $\cite{9}}&5&0.8636±0.0523&-0.0435±0.0157&-0.0201±0.0088\\
&&10&0.8246±0.0360&-0.0634±0.0133&-0.0364±0.0122\\
&ARI\cite{54}&5&0.8376±0.0950&-0.0871±0.0238&-0.0653±0.0254\\
&\textcolor{black}{DER++\cite{68p}}&\textcolor{black}{5}&\textcolor{black}{0.7157±0.0233}&\textcolor{black}{-0.2609±0.0523}&\textcolor{black}{-0.0746±0.0203}\\
&X-DER\cite{55}&5&0.8345±0.0126&-0.0935±0.0075&-0.0398±0.0124\\
&CVAE+ResNet-50$^\diamondsuit$\cite{41}&5&$\overline{0.9750}\pm\underline{\textbf{0.0000}}$&$\overline{-0.0250}$&-\\
&CGAN+ResNet-50$^\diamondsuit$\cite{41}&5&$\overline{0.9350}\pm\underline{0.0050}$&$\overline{-0.1050}$&-\\
\midrule
\multirow{3}{*}{Ours} &\multirow{3}{*}{edRVFL-R$^\dag$}&5&0.4136±0.0891&-0.0067±0.0336&-0.6008±0.1243\\
&&10&0.5789±0.1427&0.0883±0.1183&-0.4645±0.1639\\
&&20&0.9358±0.0267&0.0895±0.0838&-0.1180±0.0747\\

&\multirow{3}{*}{edRVFL-$k$F$^\dag $$^\diamondsuit$}
&5&0.9787±0.0145&0.3206±0.1144&-0.3389±0.1250\\
&&10&0.9867±0.0002&0.0064±\textbf{0.0036}&-0.0065±0.0027\\
&&20& 0.9876$\pm$\textbf{7.07e-5}&0.0077$\pm$0.0110&\textbf{-0.0062}$\pm$0.0024\\

&\multirow{3}{*}{edRVFL-$k$F-Bayes$^\dag $$^\diamondsuit$}&5&0.9804±0.0066&\textbf{0.3581}±0.0437&-0.4130±0.0475\\
&&10&\textbf{0.9879}±0.0008&0.0063±\textbf{0.0036}&-0.0066±0.0028\\
&&20&0.9872$\pm$2.24e-4&0.0108$\pm$0.0093&-0.0066$\pm$\textbf{0.0023}\\
\hline
\end{tabular}
\end{adjustbox}
\scriptsize
Note: the $\dag$ denotes this algorithm can be applied to OTCIL scenario, so that we add additional experiments on varied task stream length; algorithm with $^*$ denotes the setups follow in \cite{29}, and without $^*$ means using recommendations; $^\diamondsuit$ signifies the best top-2 benchmarks in the groups; higher metrics with lower $std.$ is better; the best value of each metric is shown in \textbf{bold}; the \underline{underline} and $\overline{overline}$ indicate the upper and lower bound respectively; the HP of our proposed methods can be found in Table \ref{tab6}; low $BWT$ indicates learning new tasks severely impacts the performance of previous ones; low $FWT$ indicates learning new tasks hardly benefits from previous tasks.
\end{table}

Our proposed methods adopted a simple sparse autoencoder-based PTM $\cal F( \cdot )$ without a convolution layer, which was introduced in CRNet (see Fig. 2 in \cite{29}). We used the same PTMs for other algorithms for a fair comparison unless one already had a specific PTM. The metric values with $std.$ are also reported in Table \ref{tab5}. In the non-CL group, the offline and separate edRVFL can be seen as the accuracy upper bound for the entire dataset and every single task, respectively. In contrast, the non-incremental version which forgets previous task information is the lower bound of $ACC$.

\textcolor{black}{For the group based on regularization, SOTA CRNet and IF2NET outperform others, which can be attributed to effective PTM and closed-form solutions to adjust elastic weight penalties, reducing retention loss during CIL.} These methods have \textcolor{black}{a} good resistance to random task order. Current OTCIL methods, such as MAS, DYSON, GEM, and GSS, experience performance degradation when encountering long task streams. RanDumb, the best model-based method, has similarities to ours in using the randomized projector and linear learner, but ours introduces unsupervised knowledge and ensemble learning. CVAE+ResNet-50 and CGAN+ResNet-50 in the replay-based group also achieve high accuracy because they use strong PTM ResNet-50 and allow revisits. \textcolor{black}{As shown in Table \ref{tab5}, our methods based on -F improve $ACC$ to \textcolor{black}{more than} 0.97, compared to 0.71 for DER++ and 0.83 for DYSON. This demonstrates that our approaches mitigate catastrophic forgetting better and without relying on replay buffers (unlike DER++) or complex meta-learning components (unlike DYSON).}

\begin{figure}[!htbp]
\centering
\includegraphics[width=0.67\columnwidth,trim=7mm 2mm 12mm 5mm, clip]{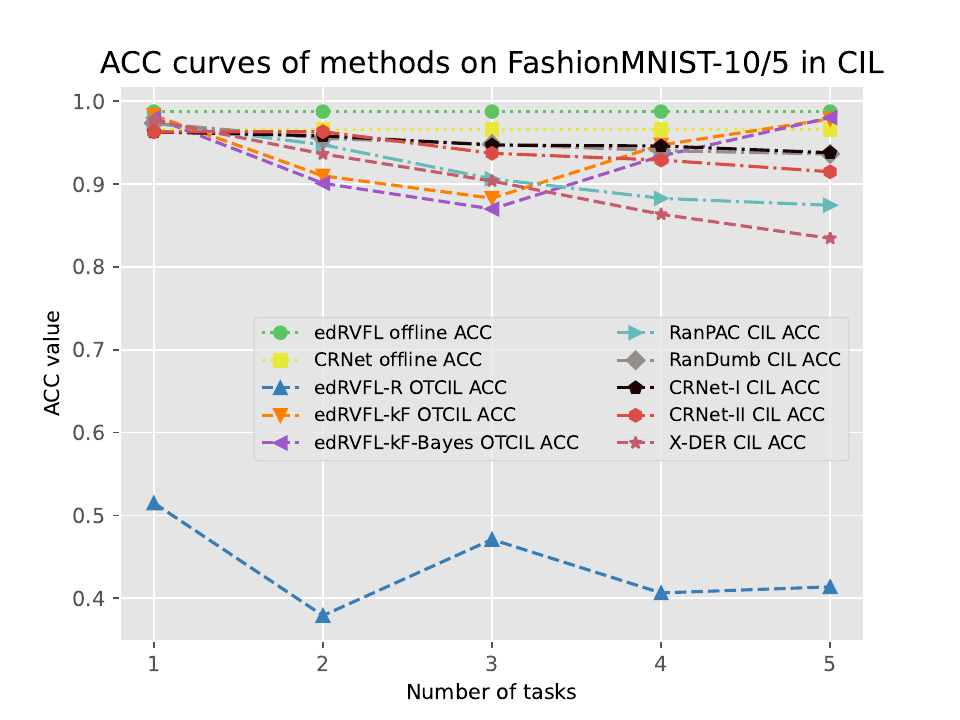}
  \caption{ACC curves of partial methods of Table \ref{tab5} on FashionMNIST-10/5. Methods learned 2 classes in each task. The edRVFL-R, edRVFL-$k$F, and edRVFL-$k$F-Bayes learned under OTCIL scenario without task boundary.}
  \label{fig a4}
\end{figure}

\begin{figure}[!htbp]
\centering
\includegraphics[width=0.67\columnwidth,trim=2mm 2mm 12mm 5mm, clip]{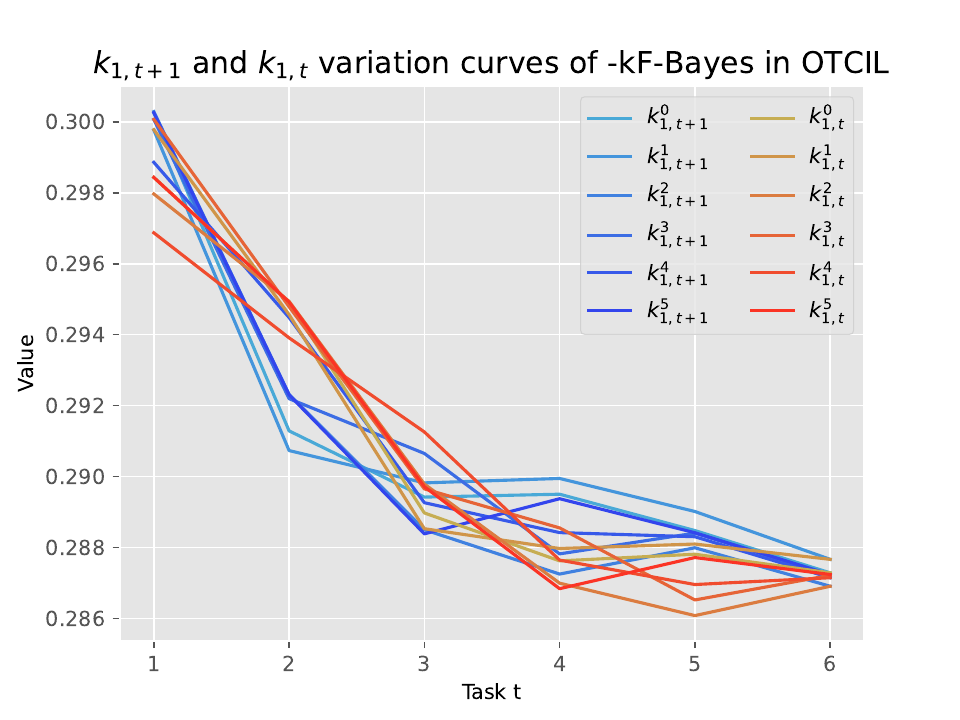}
  \caption{$k_{l,t+1}$ and $k_{l,t}$ variation curves of -$k$F-Bayes learning 5 tasks in OTCIL. The results of the first 6 experiments are shown here, marked by superscripts. It shows the dynamic adjustments and a phase shift between the two variables. Trends illustrate the effect of maintaining the balance of penalties in changing optimization targets.}
  \label{fig a5}
\end{figure}

The edRVFL-$k$F and edRVFL-$k$F-Bayes exhibit excellence in metrics. The following conclusions can be drawn from Table \ref{tab5}: (1) -$k$F-Bayes obtain a relatively superior and stable accuracy without intractable $k$ optimization in 3 varied task streams, \textcolor{black}{in contrast to} -R and -$k$F. The -R style is greatly affected when $T=5$ because of no boundary and no prior knowledge premises, which proves the efficacy of -$k$F and -$k$F-Bayes; (2) the performance of the proposed methods is remarkably close to the offline expert's. This, together with positive $BWT$ and small $FWT$, also suggests performance recovery and little learning dissipation in OTCIL even when starting with a naive model; (3) -$k$F is not worse than using -R, \textcolor{black}{under} the premise of optimized $k$; (4) based on above analysis, we recommend employing edRVFL-$k$F-Bayes with PTM in OTCIL. The $ACC$ curves on FashionMNIST-10/5 can be found in Fig. \ref{fig a4}, and the dynamic adaptive process of $k$ in -$k$F-Bayes is shown in Fig. \ref{fig a5}. \textcolor{black}{In practical industrial applications, tabular, visual, or textual data can be extracted as numerical features by specific PTMs and then fed into our algorithms for representation learning.}

\begin{table}[h]
\centering
\begin{threeparttable}
\small
\caption{Partial HP values in FashionMNIST tests}\label{tab6}%
\begin{tabular}{@{}lcccc@{}}
\toprule
$\text{Parameters}$  &
\multirow{2}{*} {$N$}&  
\multirow{2}{*} {$L$}&
\multirow{2}{*} {${\log _2}({\lambda ^{ - 1}})$}&
\multirow{2}{*} {$k$}\\
Algorithms$ \downarrow $ &  & &&\\
\midrule
edRVFL-R   & 2   & 1 & 0&-  \\
edRVFL-$k$F    &  2  & 1  & 2&1.4142127133 \\
edRVFL-$k$F-Bayes  &2  & 1 &  6 &\textit{dynamic} $k_{l}$  \\
\bottomrule
\end{tabular}
\footnotesize 
Note: we did not optimize the networks' HP by SMAC3 except for $\lambda$ and the $k$ of -$k$F style; instead, we used minimum architectures, namely $N=2$ and $L=1$ to realize competitive results in Table \ref{tab5}; $\cal F(\cdot)$ was with $N1=10$, $N2=10$,$N3=670$ as the same with CRNet.
\end{threeparttable}
\end{table}

Next, we want to explore the dynamic performance and ablation tests of edRVFL-$k$F and edRVFL-$k$F-Bayes. The methods were first configured by SMAC3 with the best HP sets to \textcolor{black}{achieve} optimal performance, where the $N$ and $L$ in Table \ref{tab6} were increased to 512 and 5 respectively, and each class was learned \textcolor{black}{in two batches}. The following describes the conducted ablation experiments.

\begin{figure}[!htbp]
\centering
\includegraphics[width=0.67\columnwidth,trim=7mm 2mm 12mm 5mm, clip]{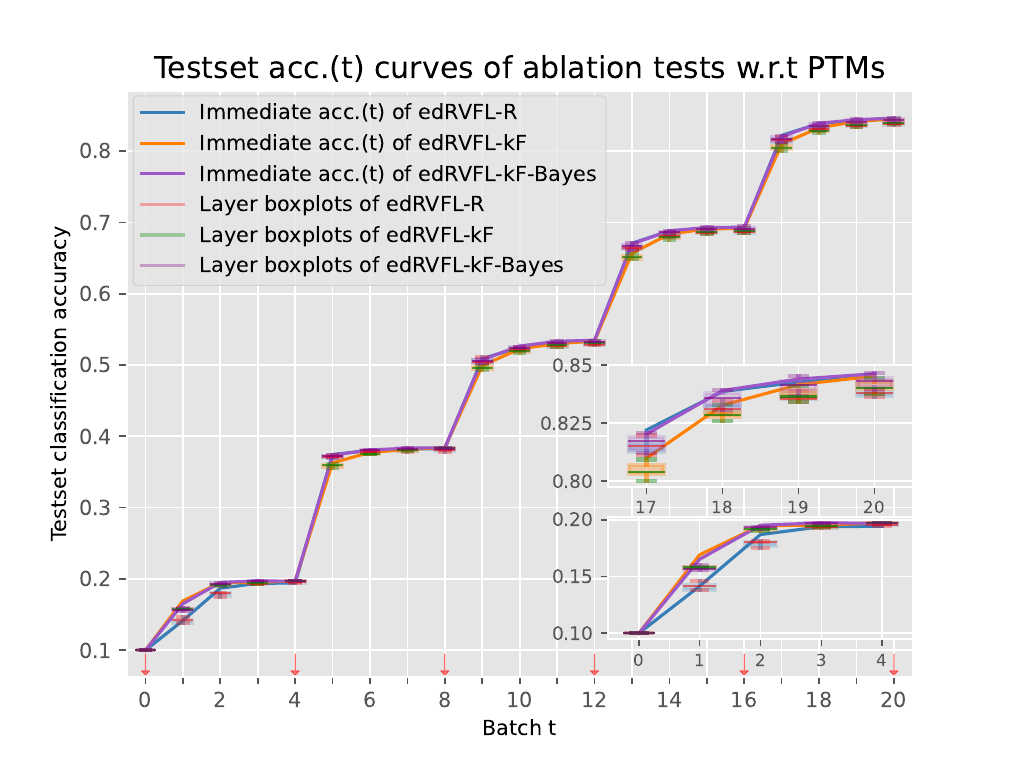}
  \caption{The FashionMNIST testset immediate $acc.(t)$ curves of edRVFL using -R, -$k$F, and -$k$F-Bayes styles without PTM in OTCIL process. Accuracy increases in steps with time. X-axis denotes batches where algorithms learn one class in two batches, and Y-axis shows accuracy values. Boxplot shows statistics of immediate accuracy for interior sub-learners. The \textcolor{red}{\(\downarrow\)} reveals task boundary. X-axis is locally enlarged in the inlaid subfigures.}
  \label{fig a6}
\end{figure}

\begin{figure}[!htbp]
\centering
\includegraphics[width=0.67\columnwidth,trim=1.5mm 2mm 8mm 2mm, clip]{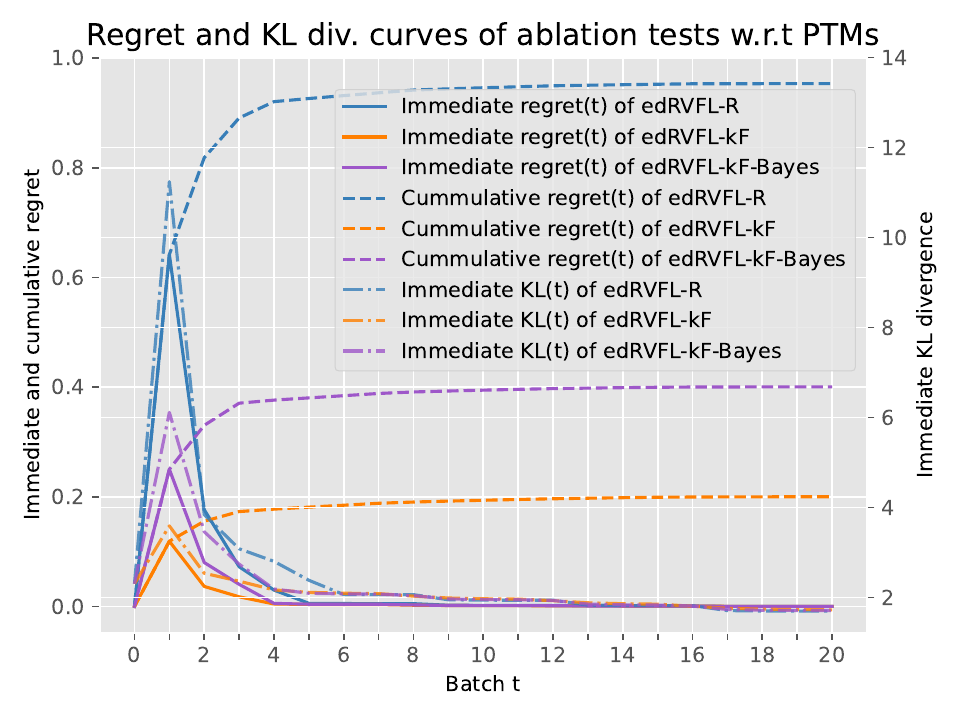}
  \caption{The FashionMNIST testset immediate $regret(t)$ and $KL(t)$ curves of edRVFL using -R, -$k$F, and -$k$F-Bayes styles in OTCIL process. -$k$F has the minimum loss at the beginning. X-axis denotes batches, Y-axis (left) is regret and cumulative regret, and Y-axis (right) is KL divergence.}
  \label{fig a7}
\end{figure}

(1) \textit{PTM-less operation} demands \textcolor{black}{that} our methods work without a feature extractor or PTM. As shown in Fig. \ref{fig a6}, removing PTM affects the final immediate testset $acc.(t)$ where all accuracies drop to over 10$\%$. The -$k$F and -$k$F-Bayes respond quickly, and in the end, -R and -$k$F-Bayes obtain higher $acc.(T)$ as they grasp nearly full knowledge halfway through each task. Combined with Fig. \ref{fig a7}, \textcolor{black}{although -$k$F starts with a smaller error}, it does not perform as well overall as -$k$F-Bayes. This reveals the difficulties in balancing all aspects of achieving good performance by \textcolor{black}{simply} adjusting $k$ in -$k$F. 

\begin{figure}[!htbp]
\centering
\includegraphics[width=0.67\columnwidth,trim=8mm 3mm 13mm 5mm, clip]{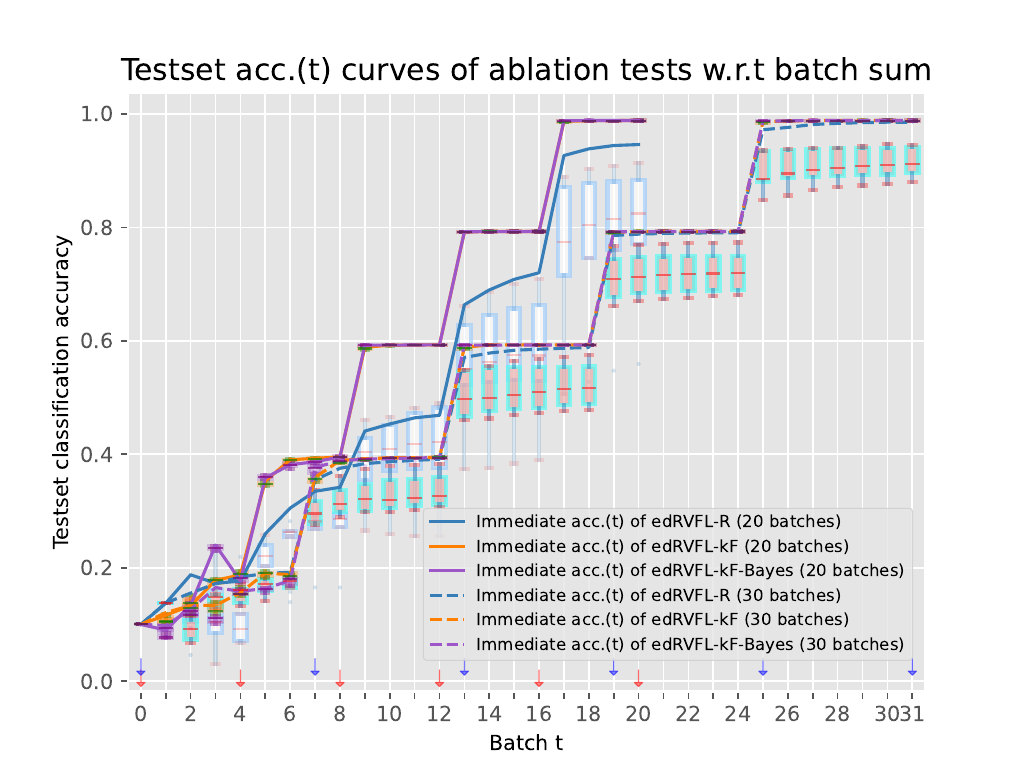}
  \caption{The FashionMNIST testset immediate $acc.(t)$ curves of edRVFL with -R, -$k$F, and -$k$F-Bayes styles learning on varied task streams. X-axis denotes batches, and Y-axis is $acc.(t)$ value. Boxplot shows statistics of immediate accuracy for interior sub-learners. The \textcolor{red}{\(\downarrow\)} and \textcolor{black}{\(\downarrow\)} respectively reveal varied task boundaries of 20- and 30-batch task streams.}
  \label{fig a8}
\end{figure}

\begin{figure}[!htbp]
\centering
\includegraphics[width=0.67\columnwidth,trim=9.5mm 2mm 13mm 5mm, clip]{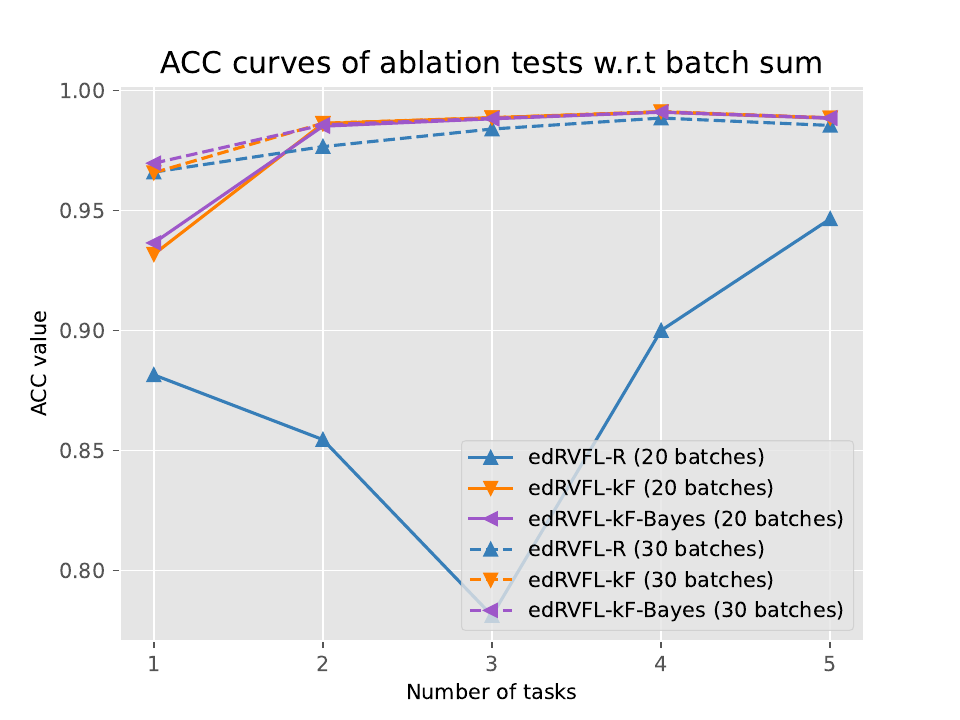}
  \caption{The FashionMNIST testset $ACC$ curves of edRVFL with -R, -$k$F, and -$k$F-Bayes styles learning on varied task streams. X-axis denotes the task number, and Y-axis is the $ACC$ value. Updating in more batches can improve our method performance.}
  \label{fig a9}
\end{figure}

(2) \textit{Varying the number of tasks split} investigates performance variation on learning task streams \textcolor{black}{of} different lengths. As shown in Fig. \ref{fig a8}, edRVFL-R suffers from large accuracy loss, and sub-learners fluctuate widely on the 20-batch task stream, while this situation is alleviated when learning on the 30-batch task stream; even sub-learners still vary greatly in their behavior. By contrast, little loss is incurred in using -$k$F and -$k$F-Bayes, and all sub-learners work well. The related $ACC$ curves can be found in Fig. \ref{fig a9}. More batches of updates improve their performance. \textcolor{black}{Given that} the methods are learning from a disturbed start, normal R needs more batches to recover the learning process and increase accuracy, while the sub-learners using -$k$F and -$k$F-Bayes can be more stable in harsh contexts and maintain expected results.



\subsection{CIFAR-100 dataset}\label{subsec5.3}
For the CIFAR-100 dataset, we split it into 3 CIL environments: CIFAR-100/5, CIFAR-100/10, and CIFAR-100/100 to implement experiments focusing on continual learning and knowledge retention, and also explored ablation tests on network architecture. We selected well-behaved methods from Table \ref{tab5}, such as EWC \cite{6}, MAS \cite{10}, CRNet \cite{29} for the regularization-based group, RanPAC \cite{19}, NICE \cite{50}, DYSON \cite{11} for the model-based methods, and GEM \cite{7}, GSS \cite{9} for the replay-based branch. A standard resnet-56 was employed as PTM and used to extract features from CIFAR-100, which inherited the setup in \cite{17,28,29} for fair comparison, and to enable benchmarks to learn image representations. We offered the same PTM for all algorithms unless they are already equipped with one, such as for NICE and DYSON. After resnet-56, the PTM $\cal F(\cdot)$ was followed in our proposed methods for enhancement. The backbone architecture and HP setups adhered to original papers as CIFAR-100 was a common dataset in the CL community. The task order was randomly sorted and baselines were tested 10 times. No task boundary was given to OTCIL methods. Our methods still learned every task in one-pass, had no revisits, and we abode by severe no prior update before the data arrival.

\begin{figure}[!htbp]
\centering
\includegraphics[width=0.67\columnwidth,trim=7mm 3mm 16mm 5mm, clip]{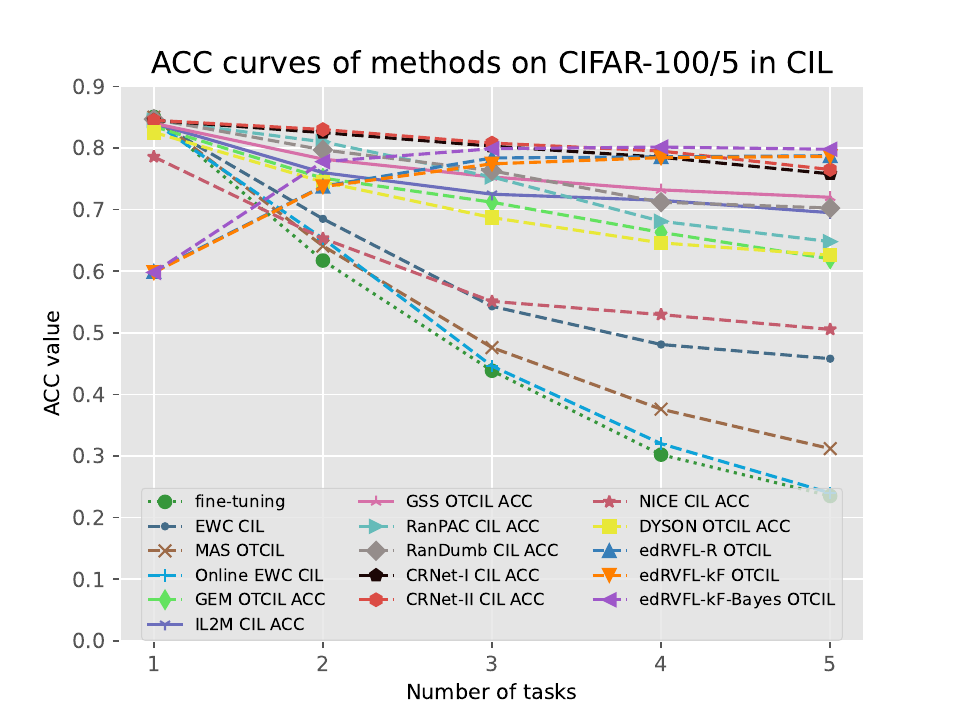}
  \caption{ACC curves of methods on CIFAR-100/5 in CIL. Each task contained 20 classes. The edRVFL-R, edRVFL-$k$F, and edRVFL-$k$F-Bayes learned tasks in OTCIL.}
  \label{fig b}
\end{figure}

\begin{figure}[!htbp]
\centering
\includegraphics[width=0.67\columnwidth,trim=7mm 3mm 13mm 5mm, clip]{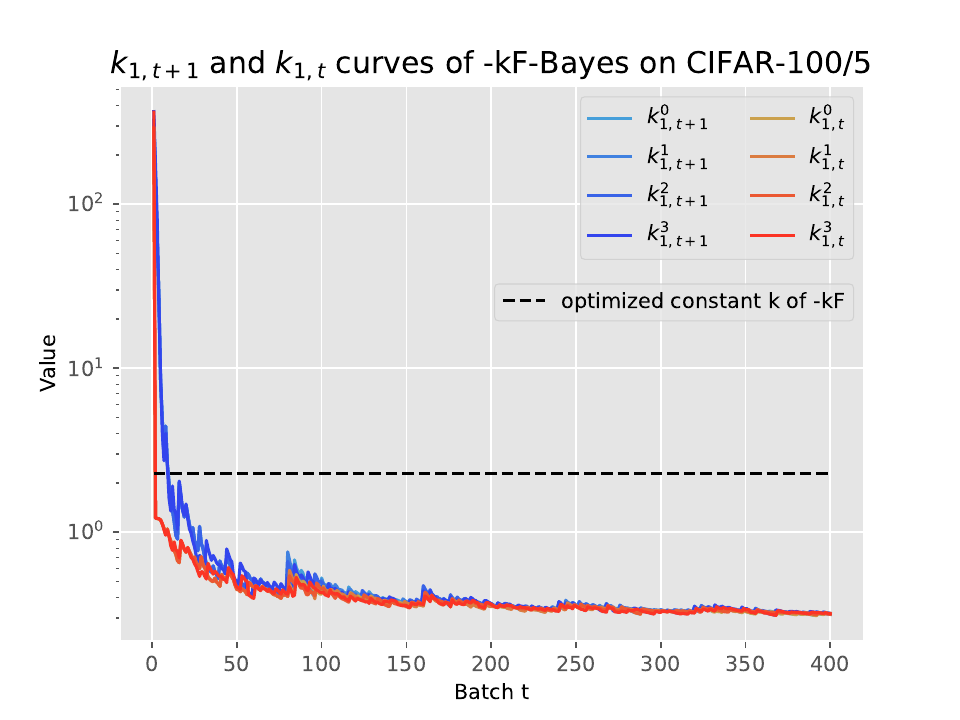}
  \caption{$k_{l,t+1}$ and $k_{l,t}$ variation curves of -$k$F-Bayes in OTCIL on CIFAR-100/5 task stream. X-axis denotes batch number, Y-axis is the logarithmic value. The results of the first 4 experiments are shown here, marked by superscripts. The phase shift between the two $k_l$ is observed again. Similar trends illustrate the effect of maintaining the balance of penalties in changing optimization targets.}
  \label{fig c}
\end{figure}

\begin{figure}[tb]
\centering
\includegraphics[width=0.67\columnwidth,trim=7mm 3mm 13mm 5mm, clip]{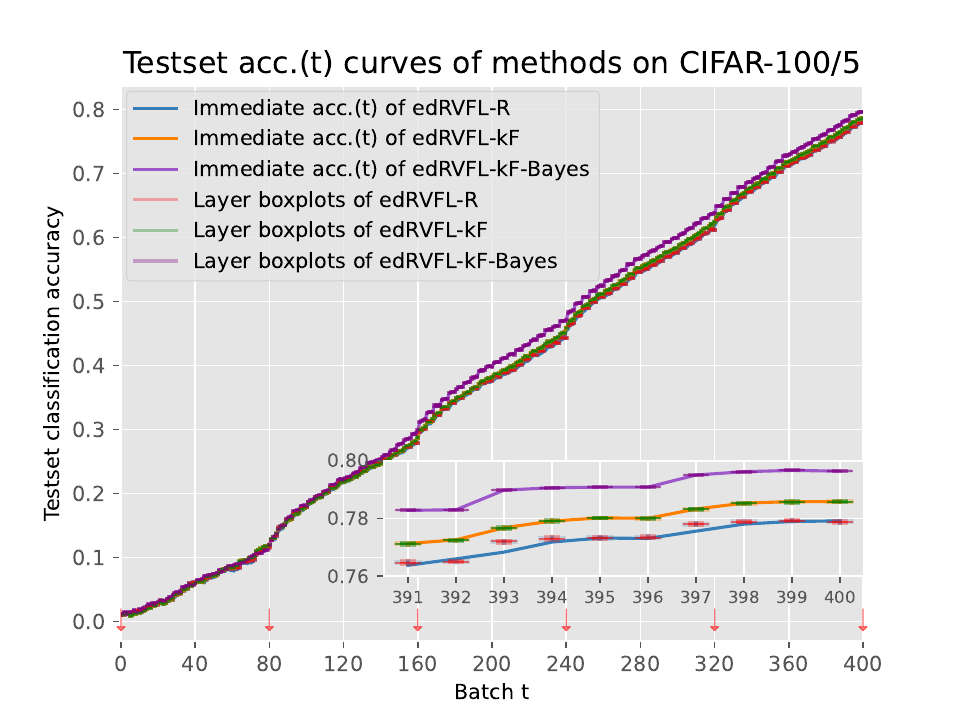}
  \caption{The CIFAR-100/5 testset immediate $acc.(t)$ curves of edRVFL using -R, -$k$F, and -$k$F-Bayes styles in OTCIL process. Accuracy steps up as more batches arrive. X-axis denotes batch number, and Y-axis represents accuracy value. Boxplot shows statistics of immediate accuracy for interior sub-learners. The \textcolor{red}{\(\downarrow\)} reveals task boundary. X-axis is locally enlarged in the inlaid subfigures.}
  \label{fig c.1}
\end{figure}

For CIFAR-100/5, the methods were subjected to learning 20 classes in each task. As the $ACC$ results are shown in Fig. \ref{fig b}, fine-tuning experienced the lowest accuracy because it destroyed past learned weights in the current tuning. Other methods (e.g. MAS, EWC) exhibit clear signs of forgetfulness when continually learning tough tasks. The IL2M and Randumb achieve remarkable accuracy and GSS also shows better performance supported by the replay buffer. CRNet-I/II obtains around 0.76 on $ACC$. Our methods were roughly adjusted in SMAC3, the number of trials was 200, $T=400$ meant each class was finished in 4 batches. Other HP are listed in Table \ref{tab7}. Our method edRVFL-$k$F-Bayes realizes the best accuracy at 0.80, and it also demonstrates a fast increase. The edRVFL-R and edRVFL-$k$F get 0.7863 and 0.7875 accuracy, respectively. The corresponding variation of $k_t$ of edRVFL-$k$F-Bayes is shown in Fig. \ref{fig c}, which shows the process of tracking during OTCIL and being adjusted automatically to optimal values, and the OTCIL dynamic performance described by $acc.(t)$ is drawn in Fig. \ref{fig c.1}. In contrast, the constant $k$ of -$k$F is only set to a rigid intermediate value by SMAC3.

\begin{table}[h]
\centering
\begin{threeparttable}
\small
\caption{Structural HP values on CIFAR-100 tests}\label{tab7}%
\begin{tabular}{@{}lccc@{}}
\toprule
Algorithms&
\multirow{2}{*} {edRVFL-R}&  
\multirow{2}{*} {edRVFL-$k$F}&
\multirow{2}{*} {edRVFL-$k$F-Bayes}\\
Main HP$ \downarrow $ &  & &\\
\midrule
$N1$$^*$  & 20   & 30 & 20  \\
$N2$$^*$  & 10   & 10 & 10  \\
$N3$$^*$  & 1000   & 1000 & 900  \\

$N$  & 512   & 8 & 512  \\
$L$ &  1  & 5  & 1 \\
${\log _2}({\lambda ^{ - 1}})$  &4$^\dag $  & 4$^\dag $  &4$^\dag $\\
$g$&Leaky ReLU&ReLU&tanh\\
$k$&-&2.2867217520&\textit{dynamic} $k_{l}$$^\dag $  \\
$\kappa$&-&-&1.0$^\dag $\\
\bottomrule
\end{tabular}
\footnotesize Note: HP marked by $^*$ is belonging to PTM $\cal F(\cdot)$; HP value marked by $^\dag $ denotes it is not included in SMAC3 optimization; the dimension of HP configuration space of -$k$F is one higher than others; subsequently we continued to use these HP for CIFAR-100 dataset.
\end{threeparttable}
\end{table}

Then we extended the task amount to 10 by creating CIFAR-100/10. The task order and choice of classes in each task were randomized. The well-behaved algorithms in Fig. \ref{fig b} were selected to be studied further in Fig. \ref{fig d}. Combined with Fig. \ref{fig e} below, the proposed edRVFL-$k$F-Bayes still maintains an obvious advantage in performance. The ultimate experiment was CIFAR-100/100, where this pattern required single CIL and presented a rigorous inspection of knowledge retention skills. We only compared with CRNets and gave some observations based on Fig. \ref{fig f}: (1) the initial harsh environments slowed down our algorithms' learning, and the impacts were gradually moderated in OTCIL process. Finally, the $ACC(T)$ of ready-to-work edRVFL-$k$F-Bayes reached 0.829; (2) edRVFL-$k$F performed badly in the new CIFAR-100/100 scenario while we still used the previous HP in Table \ref{tab7}, which demonstrated -$k$F was poorly adapted and demanded precise adjustments in different tasks. Unfortunately, it is impractical to set $k$ of -$k$F simultaneously as that requires previous data. So we need to reduce the dependence on the exact configuration of sensitive $k$; (3) the CRNets underwent large oscillations during CIL and got around 0.76 accuracy. We likewise point out that CRNet-I took over 20 times longer than ours.

\begin{figure}[!htbp]
\centering
\includegraphics[width=0.67\columnwidth,trim=7mm 3mm 16mm 5mm, clip]{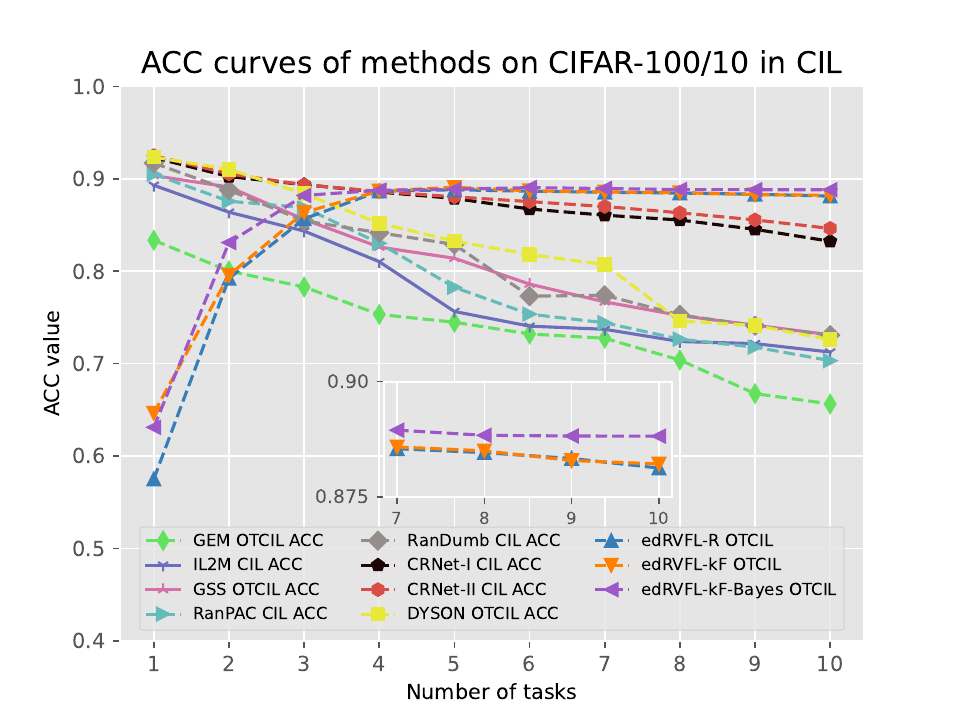}
  \caption{ACC curves of methods on CIFAR-100/10 in CIL. Each task contained 10 classes. The edRVFL-R, edRVFL-$k$F, and edRVFL-$k$F-Bayes learned tasks in OTCIL. X-axis is locally enlarged in the inlaid subfigures.}
  \label{fig d}
\end{figure}

\begin{figure}[!htbp]
\centering
\includegraphics[width=0.67\columnwidth,trim=3mm 2mm 3mm 1mm, clip]{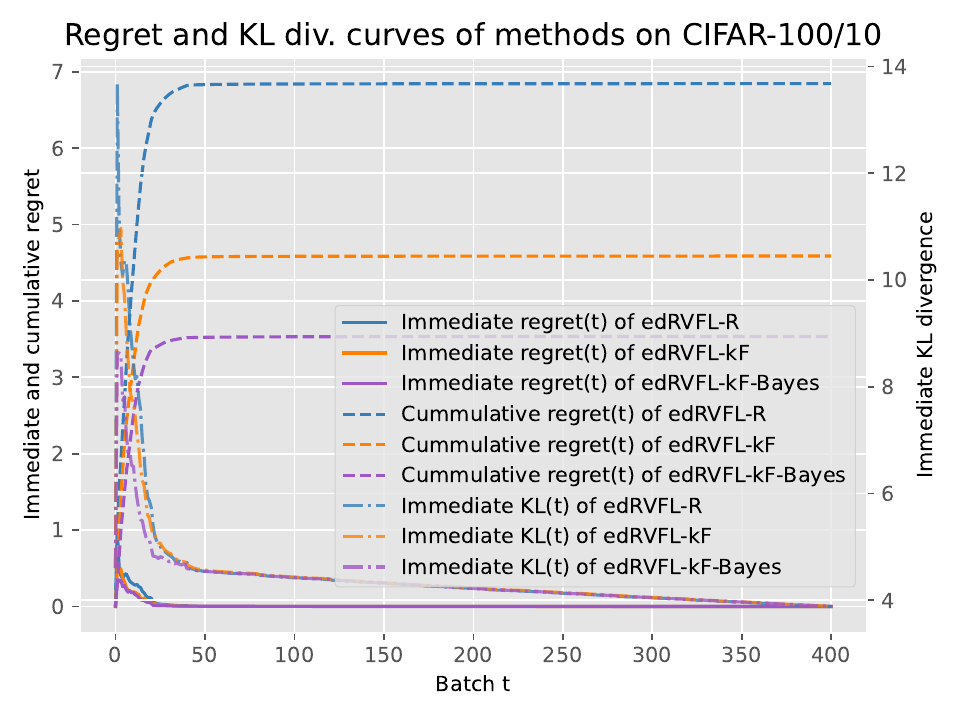}
  \caption{The CIFAR-100/10 testset immediate $regret(t)$ and $KL(t)$ curves of edRVFL using -R, -$k$F, and -$k$F-Bayes styles in OTCIL process. -$k$F-Bayes has the minimum loss in the whole process. X-axis denotes batch number, Y-axis (left) is regret and cumulative regret, and Y-axis (right) is KL.}
  \label{fig e}
\end{figure}


\begin{figure}[!htbp]
\centering
\includegraphics[width=0.67\columnwidth,trim=7mm 3mm 16mm 5mm, clip]{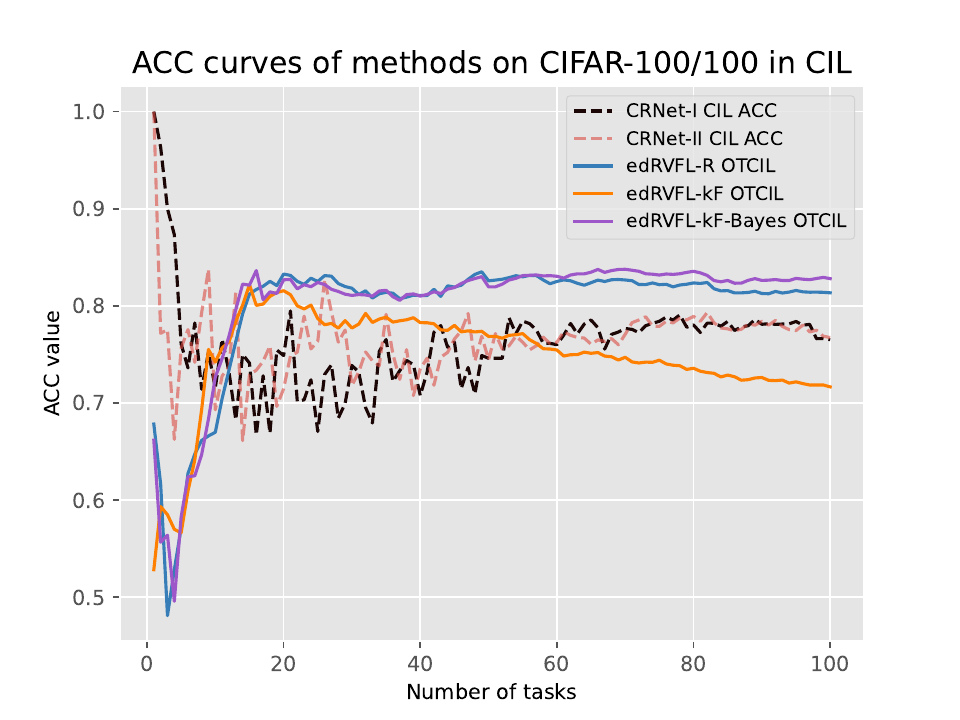}
  \caption{ACC curves of methods on CIFAR-100/100 in CIL. Each task contained 1 class. The edRVFL-R, edRVFL-$k$F, and edRVFL-$k$F-Bayes learned tasks in OTCIL, and CRNet learned tasks in CIL.}
  \label{fig f}
\end{figure}

\begin{figure}[!htbp]
\centering
\includegraphics[width=0.67\columnwidth,trim=7mm 3mm 16mm 5mm, clip]{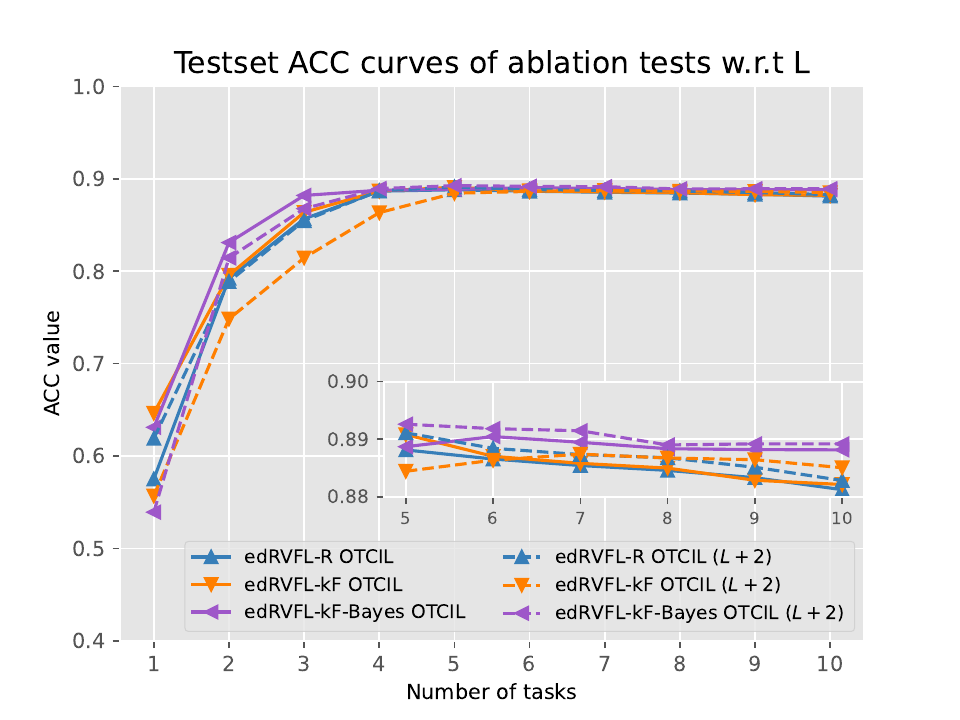}
  \caption{CIFAR-100/10 $ACC$ curves of ablation tests w.r.t $L$. $L$ was increased by 2 in all methods, and other HP were held. The edRVFL-$k$F-Bayes gets the best performance.}
  \label{fig g}
\end{figure}

\begin{figure}[!htbp]
\centering
\includegraphics[width=0.67\columnwidth,trim=7mm 3mm 16mm 5mm, clip]{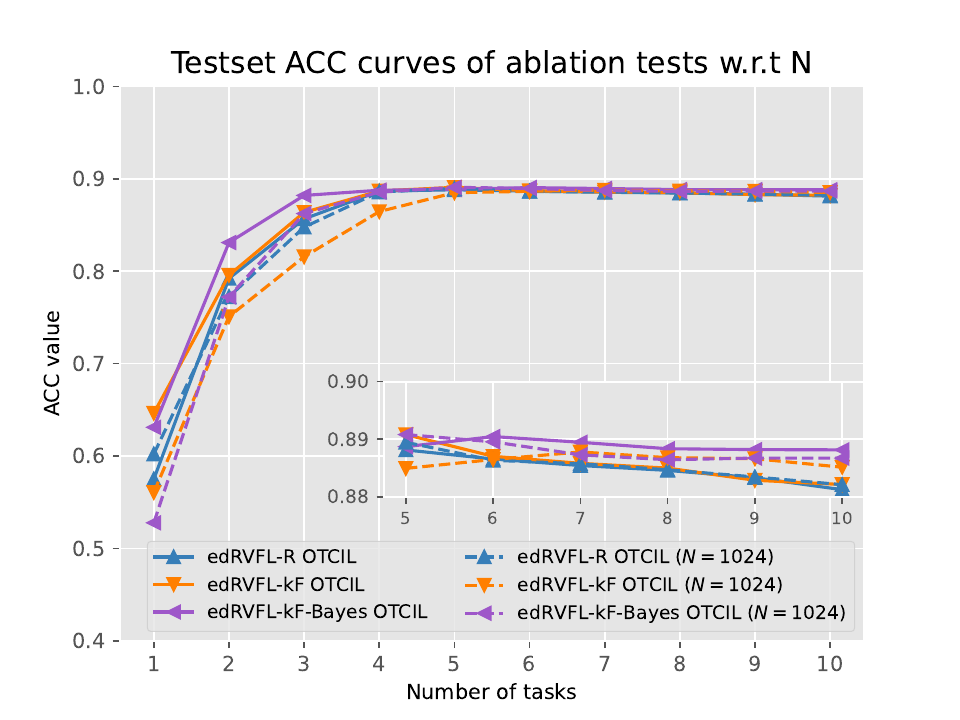}
  \caption{CIFAR-100/10 $ACC$ curves of ablation tests w.r.t $N$. $N$ was set to 1024 in all methods, and other HP were held. The edRVFL-$k$F-Bayes gets the best performance.}
  \label{fig h}
\end{figure}

The ablation experiments for CIFAR-100/10 were concentrated on network architectures. The structural HP mainly refers to the number of sub-learners' layers $L$ with $N$ neurons inside each. For the ablation tests w.r.t $L$ or $N$, we separately increased all the layers $L$ by 2 and set $N$ to 1024, and then retrained the 3 models on CIFAR-100/10. The resulting $ACC$ curves of ablation tests on $L$ and $N$ are shown in Fig. \ref{fig g} and Fig. \ref{fig h} respectively. The results suggest that the advantages of using -$k$F-Bayes still remain, and the methods mainly differ in the dynamic response speed. The -$k$F and -$k$F-Bayes strategies were relatively stable in the ablation tests.


\begin{figure}[!htbp]
\centering
\includegraphics[width=0.67\columnwidth,trim=7mm 2mm 8mm 5mm, clip]{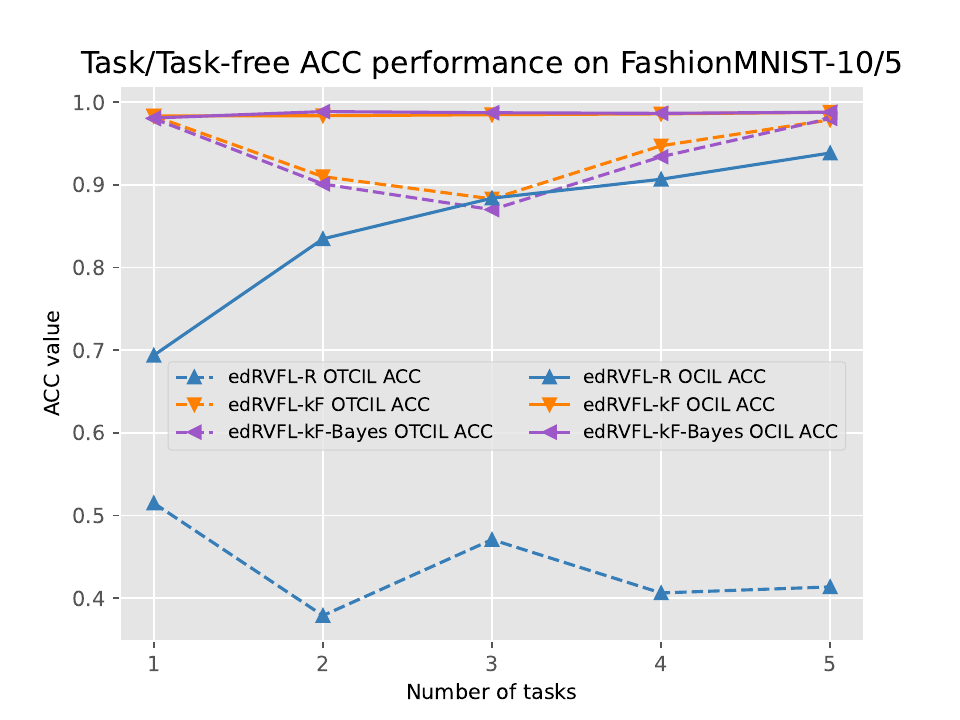}
  \caption{Task/Task-free CIL $ACC$ performance on FashionMNIST-10/5. Compare the $ACC$ changes with Fig. \ref{fig a4}. This experiment is also an ablation test that reveals the influence of removing task boundary information.}
  \label{fig a}
\end{figure}

In addition, we would like to explain why $BWT$ is positive and the reason behind the increasing trend at the start of our algorithms, which can be attributed to two factors: (1) the severe environment. To simulate the practical environment, we did not assume any prior knowledge or initialization before data arrival, so our initial HP setups were unaware of any incoming task information, but only relied on the regular OTCIL process to reasonably update. This mitigation behavior can be observed in the dashed lines of Fig. \ref{fig a}, which also shows why algorithms perform better when one task is split into more batches; (2) no task boundary information. We strictly adhered to the rules in OTCIL, and consequently, the learnable weights would be scattered among other unseen classes, resulting in misjudgment. This phenomenon was more pronounced in CIL when one task contained more classes and fewer batches. A small batch or task capacity was beneficial to the stability of the solution, and a given boundary limited the scattering of weights, further improving the accuracy. When task boundaries were known, this led to enhanced results are shown in Fig. \ref{fig a}'s solid lines.

\subsection{\textcolor{black}{Generalizability on large ViT}}\label{subsec5.4}
\textcolor{black}{Our algorithms with preposed PTMs achieved good results in image classification experiments, confirming the validity of using PTMs. Recently released official models were trained on massive datasets with handcrafted tricks, which have brought strong generalizability and made PTMs more accessible for downstream tasks \cite{56p}. Popular work on PTM-based CIL can be found in \cite{57p,58p}.}

\textcolor{black}{As the SimpleCIL combining large ViT with similarity measurements outperformed SOTA baselines \cite{56p}, we explore next whether our algorithms could still work effectively with large PTMs. Two issues were important to note: (1) PTMs should be frozen during (OT)CIL, which guaranteed consistent convex optimization and closed-form solutions in our algorithm. (2) A significant domain gap still existed between pre-trained datasets and incremental streams (e.g., ImageNet and CIFAR), requiring our algorithms to bridge distribution drift for downstream tasks. Based on the setups in \cite{56p}, we integrated a ViT from timm\footnotemark[6] into our proposed algorithms to generate generalizable embeddings for feasibility and transferability studies. }

\footnotetext[6]{\textcolor{black}{Use timm.create\_model('vit\_base\_patch16\_224\_in21k',pretrained = True, num\_classes ${\rm{ = }}$ 0) with timm${\rm{ == }}$0.6.12 or https://huggingface.co/google/vit-base-patch16-224-in21k to load PTM.}}

\textcolor{black}{The backbone ViT-B/16-IN21K with over 85M parameters was pre-trained on ImageNet-21k (14M images, more than 21K classes) at a resolution of 224x224. This PTM $\cal F(\cdot)$ was designated as [CLS] tokens of the Transformer instead of patches, which were fed into our algorithms. In this case, edRVFL-kF and edRVFL-kF-Bayes could still avoid the exemplar replay and catastrophic forgetting. The downstream task was CIFAR-100/20, which aligns with CIFAR B0 Inc5 in \cite{56p}.}

\begin{table}[h]
\centering
\begin{threeparttable}
\small
\caption{\textcolor{black}{Testset accuracy comparisons of CIL algorithms with large ViT PTM on CIFAR-100/20}}\label{tableViT}%
\begin{tabular}{@{}lccc@{}}
\hline
\textcolor{black}{Algorithms} &\textcolor{black}{$\bar {\cal A}$ }&\textcolor{black}{${\cal A}_B$}&\textcolor{black}{$ACC$}\\
\hline
\textcolor{black}{Finetune} & \textcolor{black}{38.90}& \textcolor{black}{20.17}&\textcolor{black}{ -}\\
\textcolor{black}{Finetune Adapter\cite{64p}} & \textcolor{black}{60.51}&\textcolor{black}{49.32} & \textcolor{black}{-}\\
\textcolor{black}{LwF\cite{27}} & \textcolor{black}{46.29}& \textcolor{black}{41.07}  & \textcolor{black}{-}\\
\textcolor{black}{SDC\cite{59p}} & \textcolor{black}{68.21}& \textcolor{black}{63.05}  & \textcolor{black}{-}\\
\textcolor{black}{L2P\cite{60p}} & \textcolor{black}{85.94}& \textcolor{black}{79.93 } &\textcolor{black}{ -}\\
\textcolor{black}{DualPrompt\cite{61p}} & \textcolor{black}{87.87}& \textcolor{black}{81.15} & \textcolor{black}{-}\\
\textcolor{black}{CODA-Prompt\cite{62p}} & \textcolor{black}{89.11}& \textcolor{black}{81.96} & \textcolor{black}{-}\\
\textcolor{black}{CPP\cite{63p}} &\textcolor{black}{85.21}& \textcolor{black}{78.64} & \textcolor{black}{-}\\
\textcolor{black}{SimpleCI}L &\textcolor{black}{87.57}& \textcolor{black}{81.26}&\textcolor{black}{81.28}\\
\textcolor{black}{APER w/ Finetune} &\textcolor{black}{87.67}& \textcolor{black}{81.27}&\textcolor{black}{81.27}\\
\textcolor{black}{APER w/ VPT-Shallow} &\textcolor{black}{90.43} &\textcolor{black}{84.57}&\textcolor{black}{81.80}\\
\textcolor{black}{APER w/ VPT-Deep}& \textcolor{black}{88.46} &\textcolor{black}{82.17}&\textcolor{black}{82.65}\\
\textcolor{black}{APER w/ SSF} &\textcolor{black}{87.78} &\textcolor{black}{81.98}&\textcolor{black}{82.01}\\
\textcolor{black}{APER w/ Adapter} &\textcolor{black}{90.65} &\textcolor{black}{85.15}&\textcolor{black}{85.14}\\
\hline
\textcolor{black}{edRVFL-R (ViT)}   &\textcolor{black}{93.17} & \textcolor{black}{86.17}&\textcolor{black}{90.13} \\
\textcolor{black}{edRVFL-$k$F (ViT)}   & \textcolor{black}{93.72}&\textcolor{black}{86.14}&\textcolor{black}{90.32}\\
\textcolor{black}{edRVFL-$k$F-Bayes  (ViT)} &\textbf{\textcolor{black}{94.48}}&\textbf{\textcolor{black}{86.28}}&\textbf{\textcolor{black}{90.92}}\\

\hline
\end{tabular}
\textcolor{black}{Note: APER-based methods and defined accuracy ($\bar {\cal A}$, ${\cal A}_B$) for comparison could be found in \cite{56p}; the best value of each metric is shown in \textbf{bold}}.
\end{threeparttable}
\end{table}

\textcolor{black}{Table \ref{tableViT} records the results of the above experiments. Our methods used common parameter configurations rather than special tuning: $N=256$, $L=2$, $g=Sigmoid$ for all backbone edRVFLs, $k=2.0$ for -$k$F style, and $\sigma ={10^{ - 3}}$ and $\kappa = 128$ for -$k$F-Bayes style. We only fine-tuned the $\lambda$ of precarious edRVFL-R to ensure robust results, as some singularities dropped its accuracy by more than 30\%. The following conclusions can be drawn: (1) under the condition that large PTMs were accessible, our proposed methods still yielded remarkable results in all algorithms. (2) edRVFL-$k$F-Bayes outperforms others, and edRVFL-$k$F is the second best in our group, though they shared the same backbone structure. These prove that edRVFL-$k$F-Bayes remains efficient for large PTMs.}

\begin{figure}[htbp]
    \centering
    \begin{subfigure}[]
        \centering
        \includegraphics[trim=21mm 14mm 16mm 15mm, clip,width=0.138\textwidth]{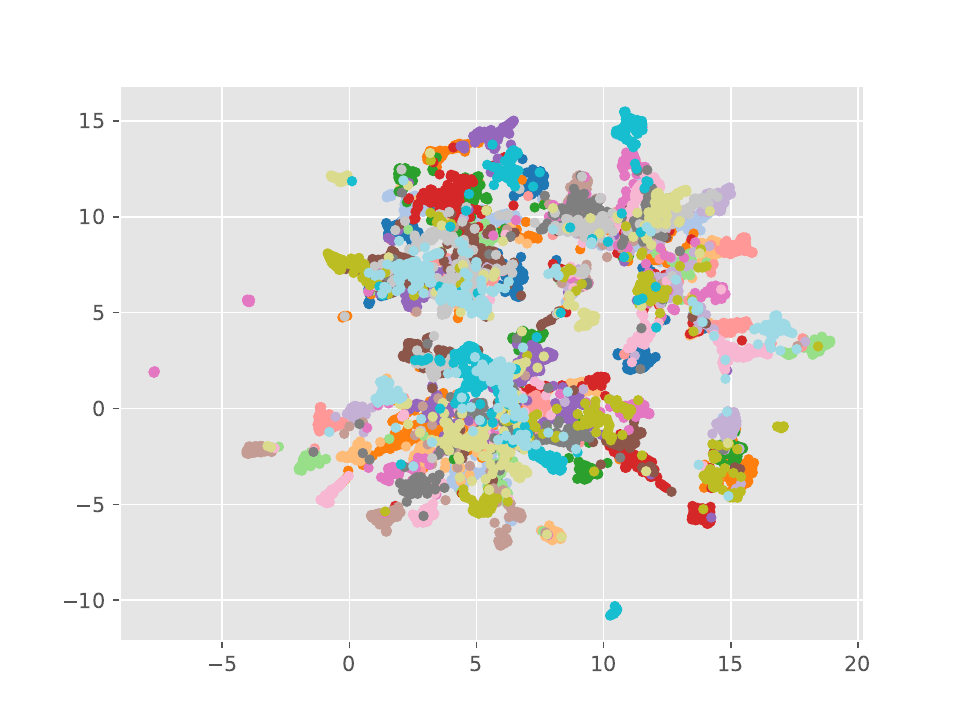} 
        \label{figt1}
    \end{subfigure}
    \hspace{-4mm}
    \begin{subfigure}[]
        \centering
        \includegraphics[trim=21mm 14mm 16mm 15mm, clip,width=0.138\textwidth]{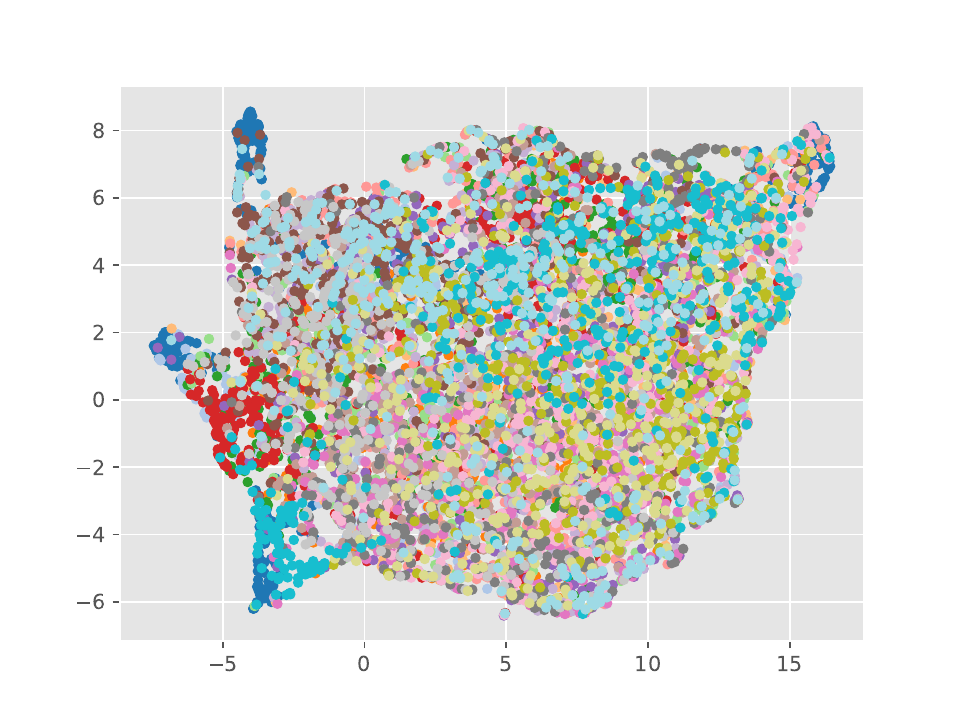} 
        \label{figt2}
    \end{subfigure}
    \hspace{-4mm}
    \begin{subfigure}[]
        \centering
        \includegraphics[trim=21mm 14mm 16mm 15mm, clip,width=0.138\textwidth]{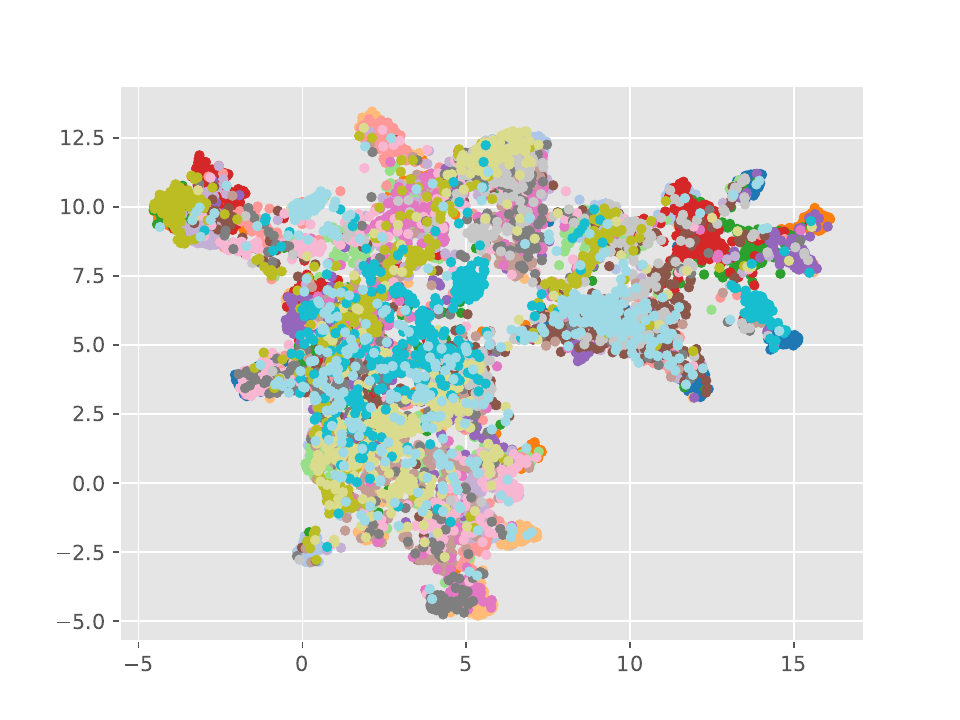} 
        \label{figt3}
    \end{subfigure}
    \hspace{-4mm}
    \begin{subfigure}[]
        \centering
        \includegraphics[trim=21mm 14mm 16mm 15mm, clip,width=0.138\textwidth]{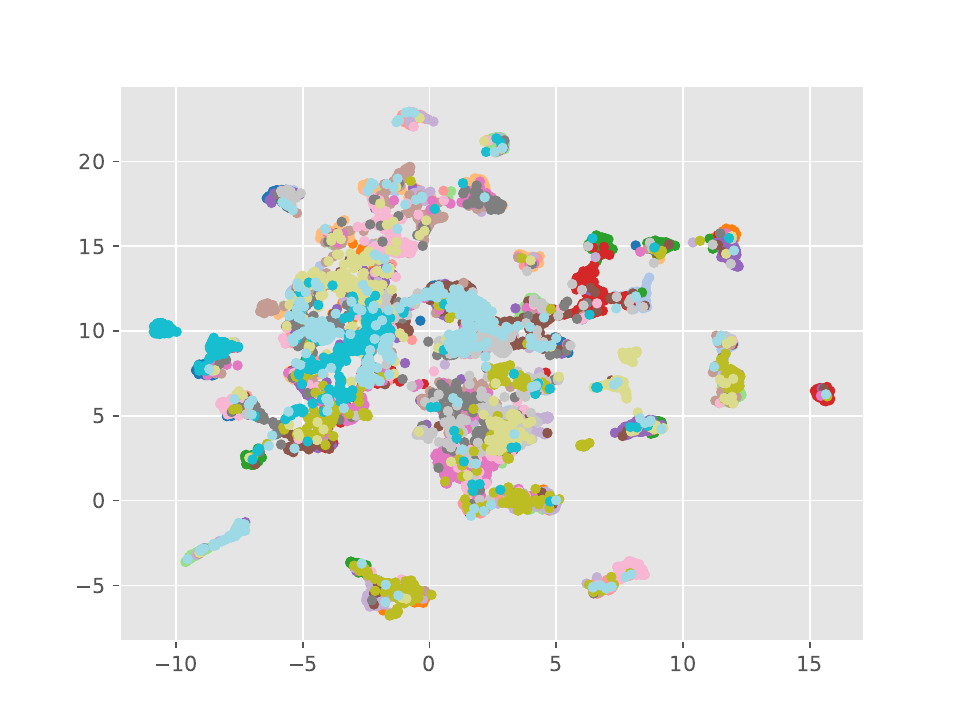} 
        \label{figt4}
    \end{subfigure}
    \hspace{-4mm}
    \begin{subfigure}[]
        \centering
        \includegraphics[trim=21mm 14mm 16mm 15mm, clip,width=0.138\textwidth]{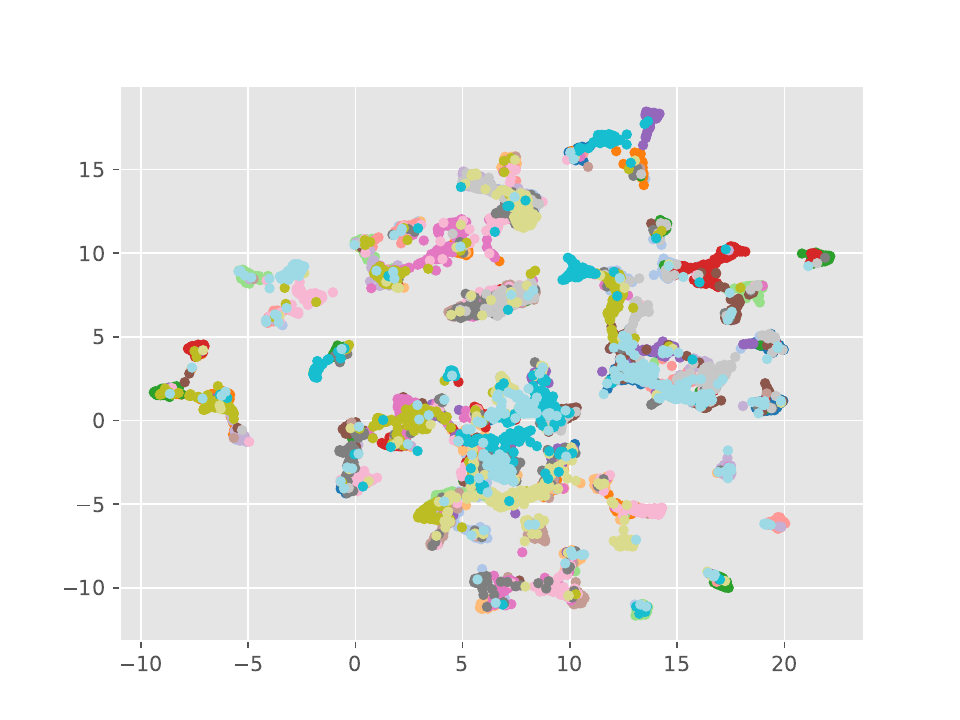} 
        \label{figt5}
    \end{subfigure}
    \hspace{-4mm}
    \begin{subfigure}[]
        \centering
        \includegraphics[trim=21mm 14mm 16mm 15mm, clip,width=0.138\textwidth]{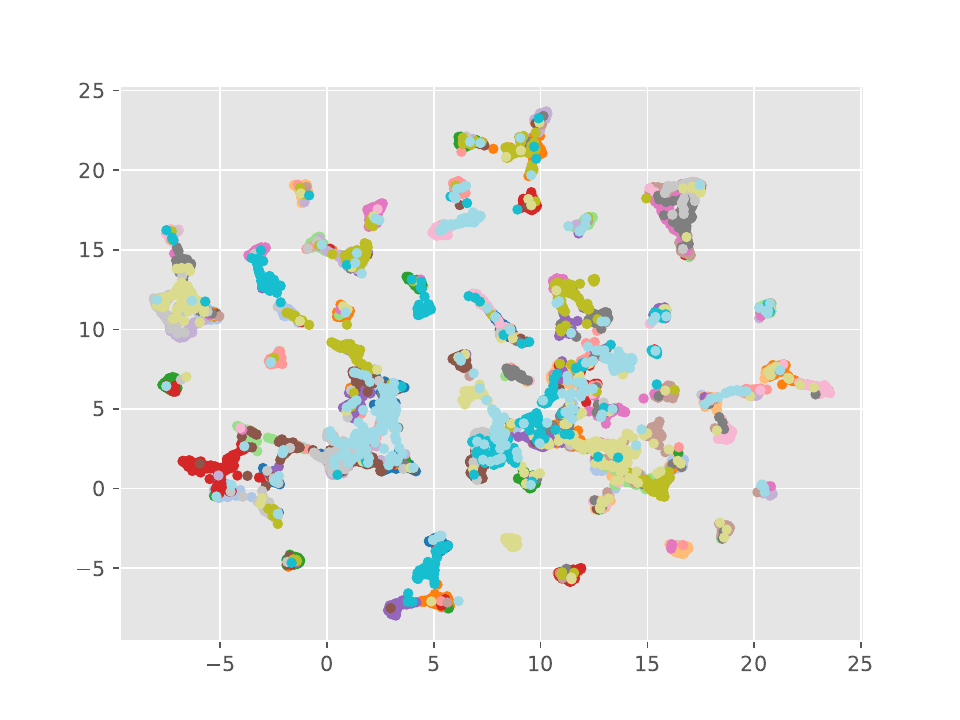} 
        \label{figt6}
    \end{subfigure}
    \hspace{-4mm}
    \begin{subfigure}[]
        \centering
        \includegraphics[trim=21mm 14mm 16mm 15mm, clip,width=0.138\textwidth]{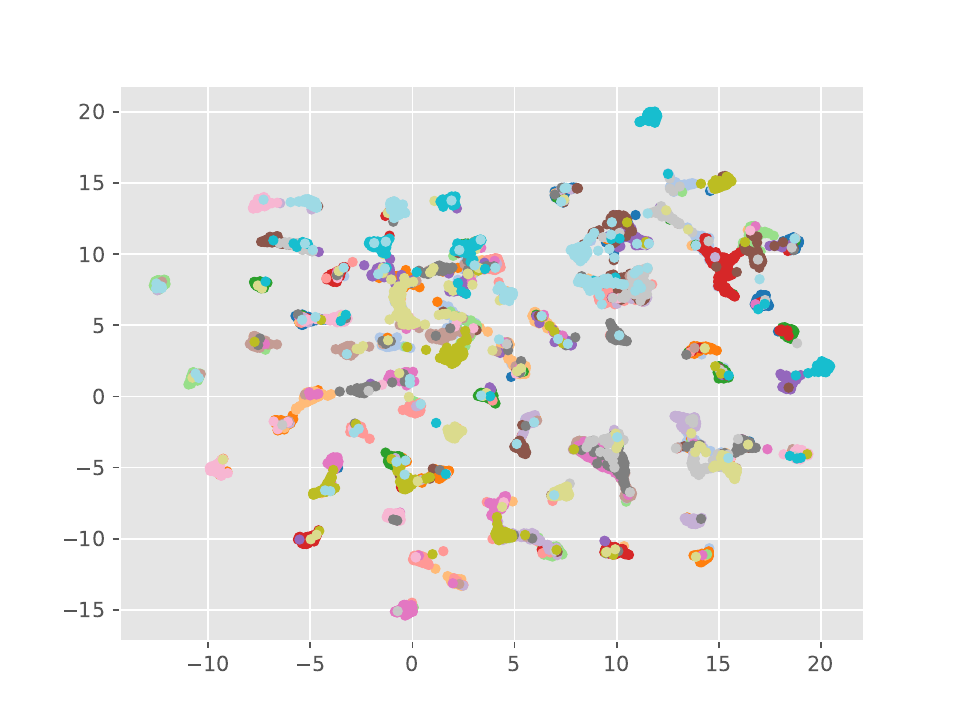} 
        \label{figt7}
    \end{subfigure}
     \\
    \begin{subfigure}[]
        \centering
        \includegraphics[trim=21mm 14mm 16mm 15mm, clip,width=0.14\textwidth]{Figure_0.pdf} 
        \label{figt8}
    \end{subfigure}
    \hspace{-4mm}
    \begin{subfigure}[]
        \centering
        \includegraphics[trim=21mm 14mm 16mm 15mm, clip,width=0.14\textwidth]{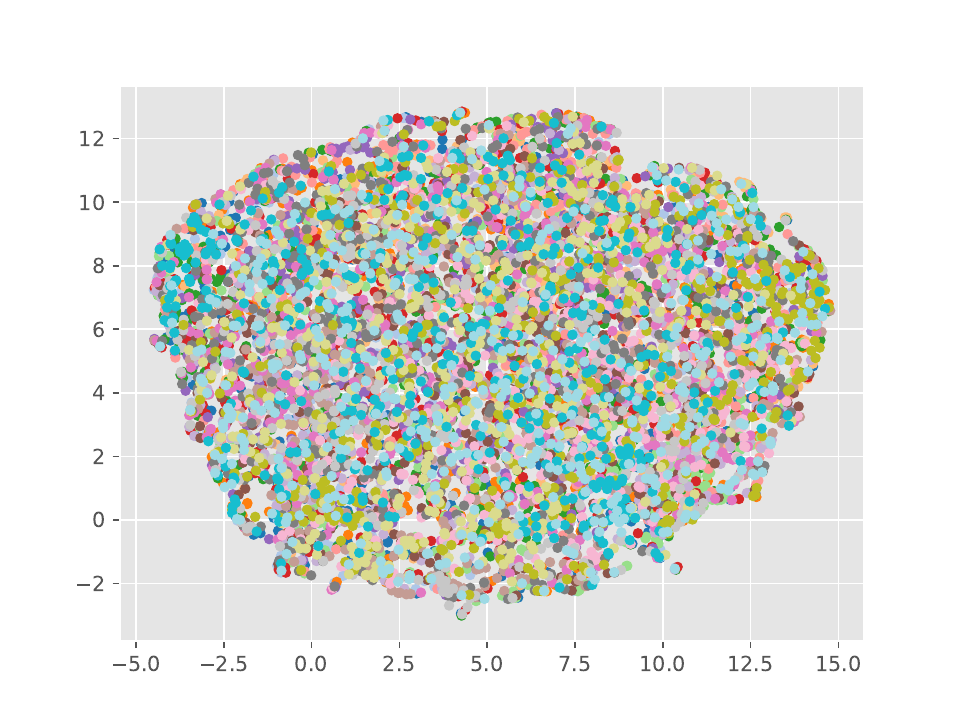} 
        \label{figt9}
    \end{subfigure}
    \hspace{-4mm}
    \begin{subfigure}[]
        \centering
        \includegraphics[trim=21mm 14mm 16mm 15mm, clip,width=0.14\textwidth]{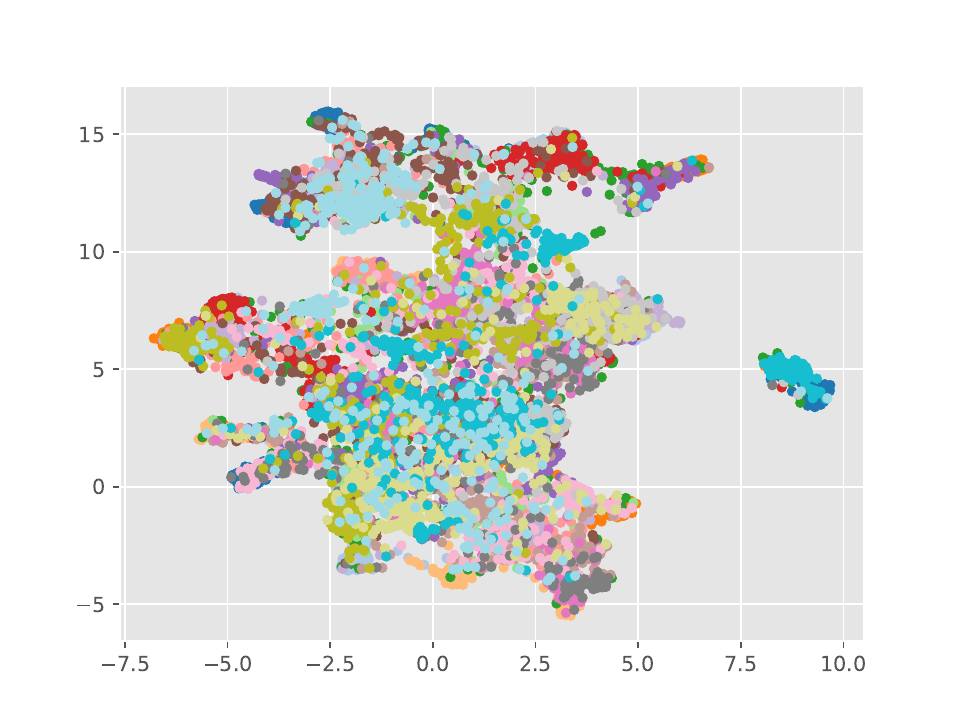} 
        \label{figt10}
    \end{subfigure}
    \hspace{-4mm}
    \begin{subfigure}[]
        \centering
        \includegraphics[trim=21mm 14mm 16mm 15mm, clip,width=0.14\textwidth]{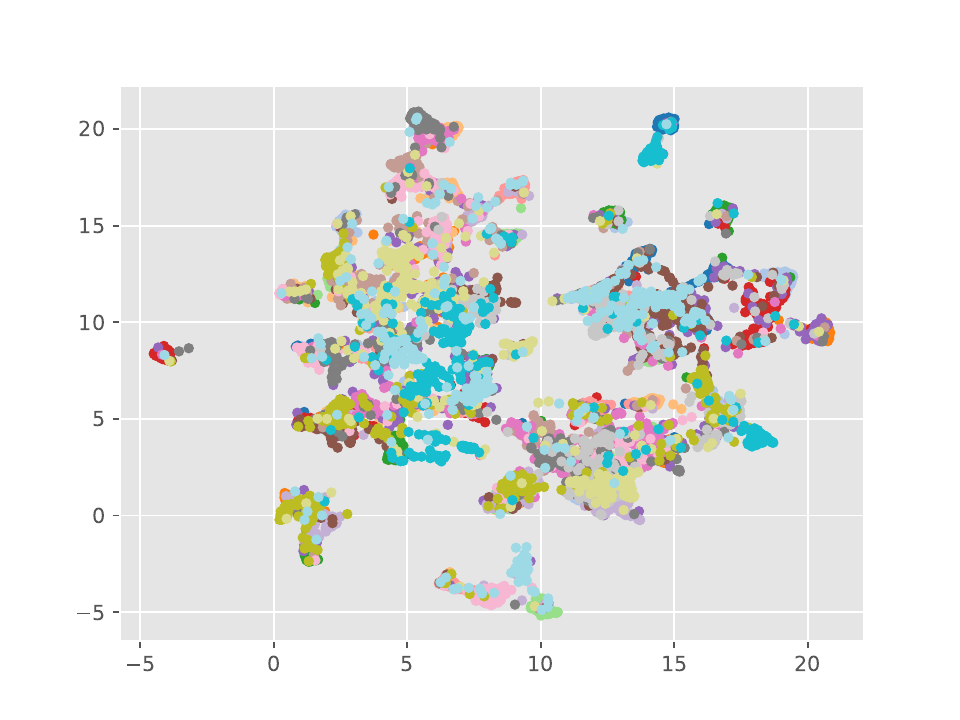} 
        \label{figt11}
    \end{subfigure}
    \hspace{-4mm}
    \begin{subfigure}[]
        \centering
        \includegraphics[trim=21mm 14mm 16mm 15mm, clip,width=0.14\textwidth]{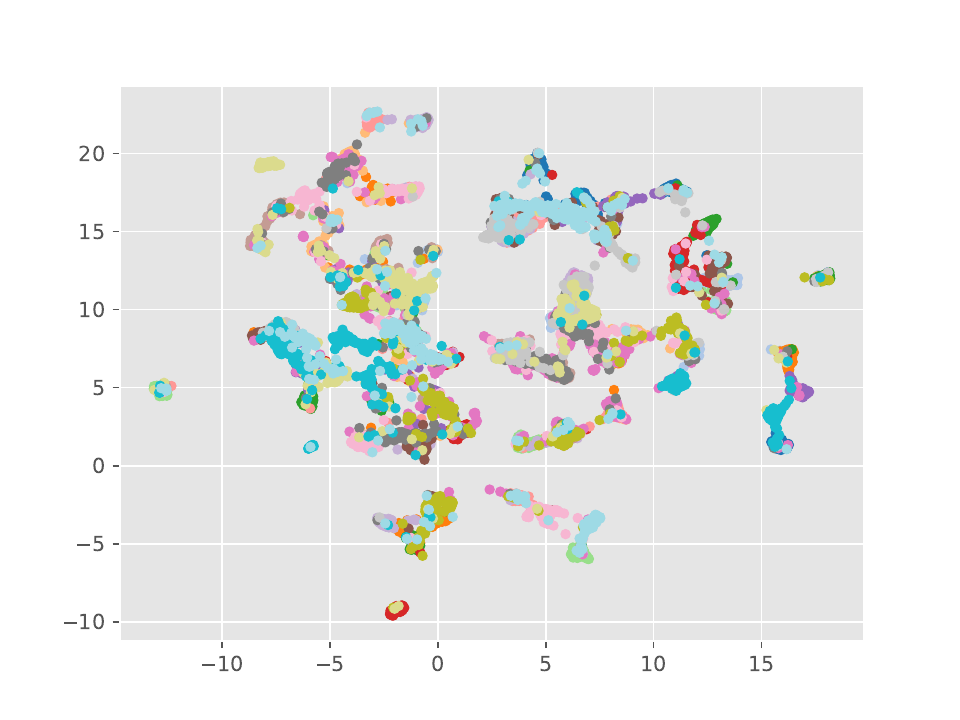} 
        \label{figt12}
    \end{subfigure}
    \hspace{-4mm}
    \begin{subfigure}[]
        \centering
        \includegraphics[trim=21mm 14mm 16mm 15mm, clip,width=0.14\textwidth]{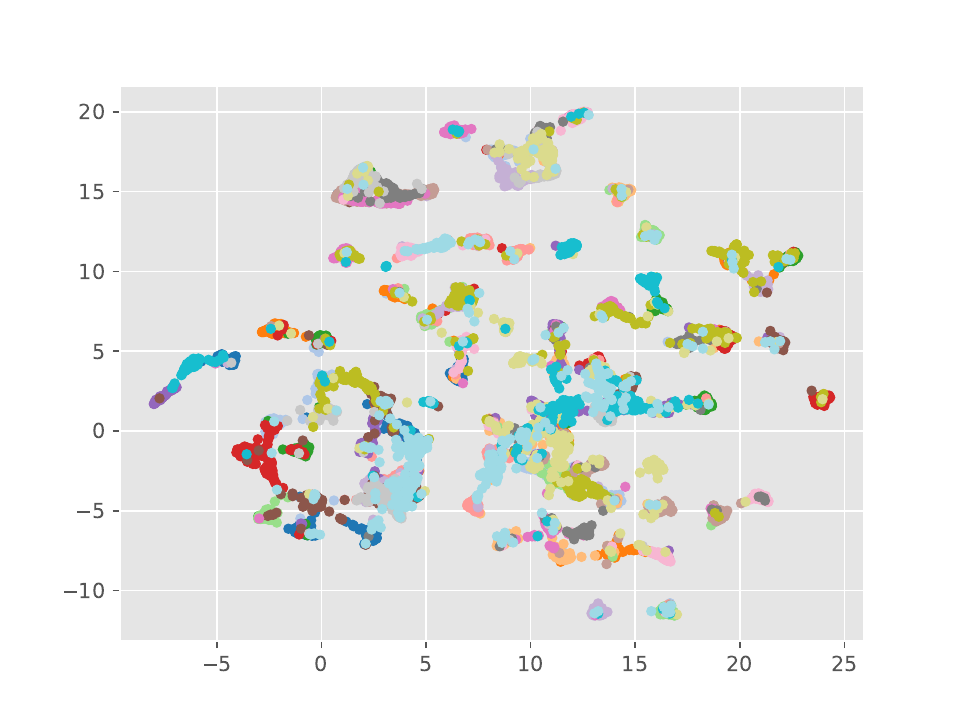} 
        \label{figt13}
    \end{subfigure}
    \hspace{-4mm}
    \begin{subfigure}[]
        \centering
        \includegraphics[trim=21mm 14mm 16mm 15mm, clip,width=0.14\textwidth]{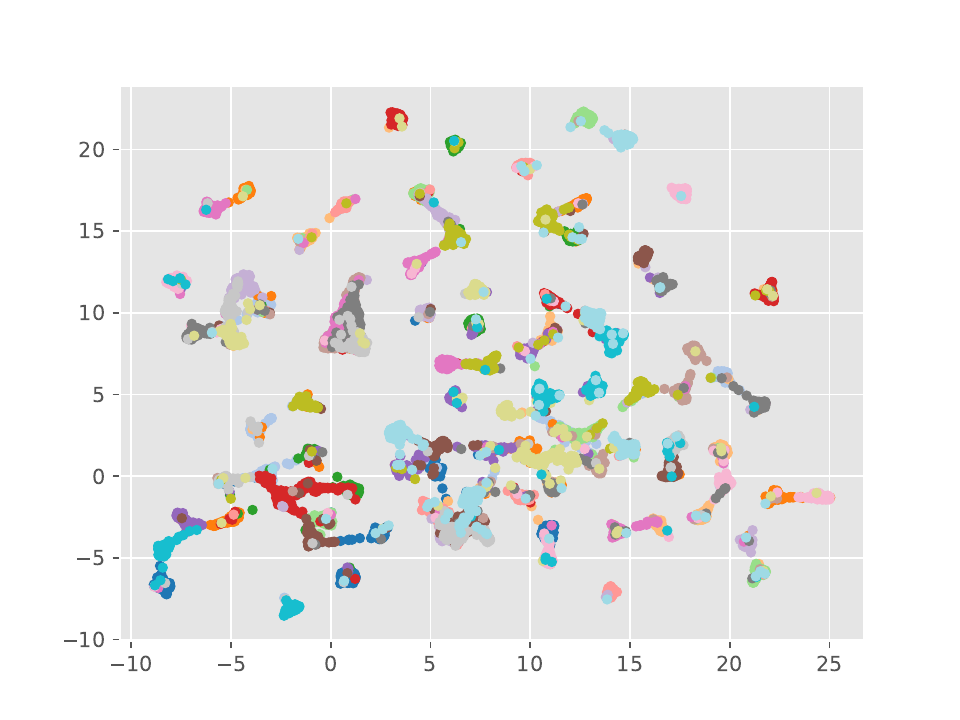} 
        \label{figt14}
    \end{subfigure}

    \caption{\textcolor{black}{t-SNE visualization of the decision boundary on CIFAR-100/20, where colorful dots denote data representations after ViT and randomization. (a)-(g): boundary evolution of using edRVFL-R (ViT). (h)-(n): boundary evolution of using edRVFL-$k$F-Bayes (ViT). (a), (h) are the initial features; (b)-(g) represent the results after learning 0, 20, 40, 60, 80, 100 classes respectively, as do (i)-(n).}}
    \label{figtsne}
 
\end{figure}

\textcolor{black}{Next, we investigated how the random projections interacted with the forward style by visualizing the decision boundaries using t-SNE during the OTCIL process. In Fig. \ref{figtsne}, we compared edRVFL-R (ViT) and edRVFL-$k$F-Bayes (ViT). It can be observed that: (1) the random mapping provides complete representations, and the regularized loss functions can gradually partition the boundary of the random projections as OTCIL progresses, proving that algorithms work. (2) edRVFL-$k$F-Bayes (ViT) employs enhanced gradients that result in lower future regrets in OTCIL, leading to sparser between-class margins and improved classification results. Visualizations reveal the evolution mechanism and effectiveness of using edRVFL-$k$F-Bayes (ViT) in OTCIL.}

\begin{figure}[htbp]
    \centering
    \begin{subfigure}[]
        \centering
        \includegraphics[trim=5mm 2mm 5mm 5mm, clip,width=0.43\textwidth]{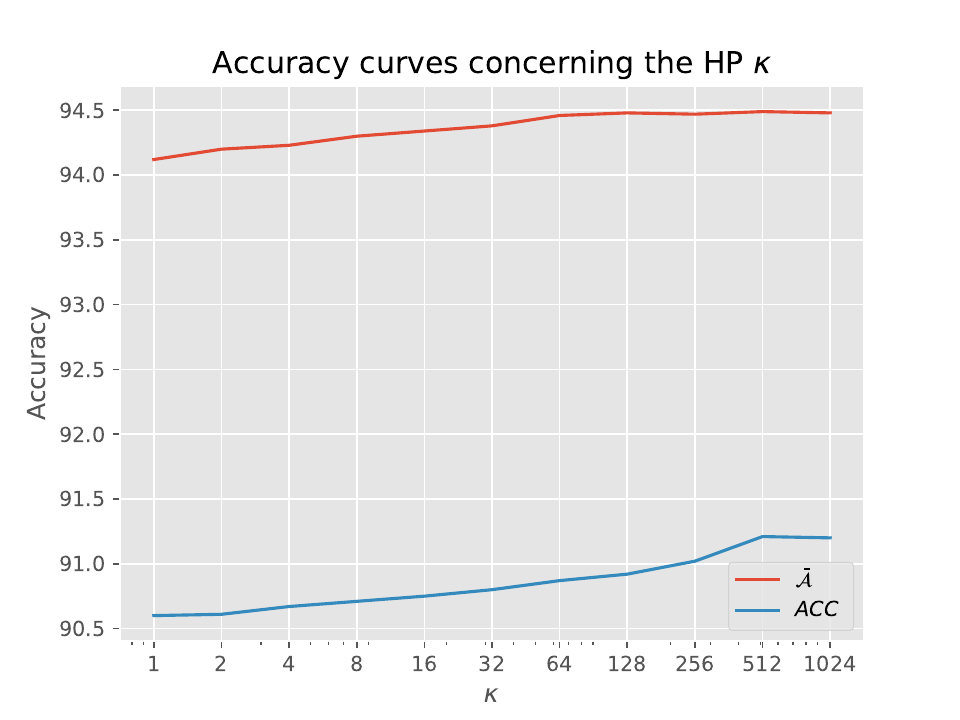} 
        \label{fighp1}
    \end{subfigure}
    \hspace{-4mm}
    \begin{subfigure}[]
        \centering
        \includegraphics[trim=5mm 2mm 5mm 5mm, clip,width=0.43\textwidth]{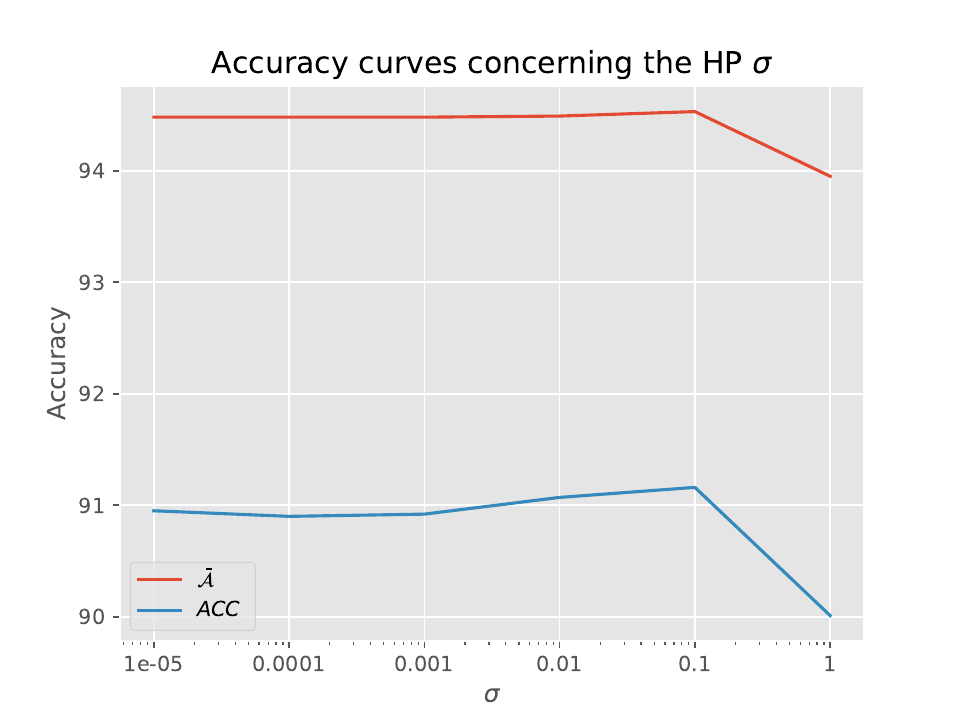} 
        \label{fighp2}
    \end{subfigure}
    
    \caption{\textcolor{black}{Effect of main HP on edRVFL-$k$F-Bayes on CIFAR-100/20. (a): accuracy variation under HP $\kappa$ changes. (b): accuracy variation under HP $\sigma$ changes.}}
    \label{fighp}
 
\end{figure}

\textcolor{black}{Finally, we would like to explain the specific HP (i.e., $\sigma$, $\kappa$) selections of edRVFL-$k$F-Bayes. Although this appears to introduce extra HP over edRVFL-$k$F, it is actually more stable and easier to tune because the performance of edRVFL-$k$F-Bayes is less sensitive and dependent on these two HP. In Fig. \ref{fighp}, the magnitudes of the performance variations due to different $\kappa$ are less than 0.5\% for $\bar {\cal A}$ and 0.7\% for $ACC$; the magnitudes of variations in performance due to different $\sigma$ are less than 0.5\% for $\bar {\cal A}$ and 1.2\% for $ACC$. The results demonstrate that edRVFL-$k$F-Bayes will maintain similar superiority and robustness over wide ranges of its HP. Therefore, we recommend manual adjustment or linear search to adjust them.}

\section{Conclusion}\label{sec13}
To achieve immediate decision-making and reduce loss of knowledge retention in OTCIL scenarios, especially containing long task streams, a framework of multi-layer Randomized NN with controllable unsupervised knowledge is proposed to enhance precognition learning and guide gradients of continuous optimization, hoping to outperform the canonical -R and have \textcolor{black}{lower regret}. Regarding the framework as the OCO, we theoretically derive the algorithm of edRVFL-$k$F with adjustable $k$ forward regularization, which features one-pass closed-form incremental updates and variable learning rate. It also effectively resists catastrophic forgetting and replay. Moreover, to avoid unstable penalties in OTCIL and mitigate intractable tuning of hard $k$, we propose the ready-to-work edRVFL-$k$F-Bayes with soft $k$ self-adapted, based on Bayesian learning and \textcolor{black}{distribution retaining} in the ever-changing optimization target of non-i.i.d streams. We carried out comprehensive experiments in multiple scenarios, from which we observed the superiority of using edRVFL-$k$F-Bayes in OTCIL processes. Our future work will focus on the theoretical foundation of regret bounds using the -$k$F style in OTCIL.

\textcolor{black}{Beyond technical merits, it is important to note the potential misuse stemming from the method's incremental efficiency. Since the closed-form incremental update enables rapid adaptation to new facial identities or behavioral patterns, deployment without proper governance can lead to unauthorized surveillance systems that continuously integrate unseen individuals' data.}

\section*{Acknowledgements}
The authors thank the National Natural Science Foundation of China (No. 62273230, 62203302) and the State Scholarship Fund of China Scholarship Council (No. 202206230182) for their financial support. We also thank the editors and reviewers for their valuable comments and suggestions. 

\section*{Declaration of Interest}
The authors declare that they have no competing interests or personal relationships that could influence the work reported in this paper.

\section*{CRediT Author Statement}
Junda Wang: Methodology, Software, Writing - Original Draft. Minghui Hu: Conceptualization, Validation. Ning Li: Supervision, Funding acquisition. Abdulaziz Al-Ali: Supervision, Resources. Ponnuthurai Nagaratnam Suganthan: Supervision, Project administration.

\FloatBarrier
\bibliographystyle{elsarticle-num} 
  \bibliography{sn-bibliography}

\begin{thebibliography}{10}
\expandafter\ifx\csname url\endcsname\relax
  \def\url#1{\texttt{#1}}\fi
\expandafter\ifx\csname urlprefix\endcsname\relax\def\urlprefix{URL }\fi
\expandafter\ifx\csname href\endcsname\relax
  \def\href#1#2{#2} \def\path#1{#1}\fi

\bibitem{6}
J.~Kirkpatrick, R.~Pascanu, N.~Rabinowitz, J.~Veness, G.~Desjardins, A.~A. Rusu, K.~Milan, J.~Quan, T.~Ramalho, A.~Grabska-Barwinska, et~al., Overcoming catastrophic forgetting in neural networks, Proceedings of the national academy of sciences 114~(13) (2017) 3521--3526.

\bibitem{29}
D.~Li, Z.~Zeng, Crnet: A fast continual learning framework with random theory, IEEE Transactions on Pattern Analysis and Machine Intelligence 45~(9) (2023) 10731--10744.

\bibitem{bib1}
M.~Kim, Theoretical bounds of generalization error for generalized extreme learning machine and random vector functional link network, Neural Networks 164 (2023) 49--66.

\bibitem{bib11}
Q.~Wei, W.~Zhang, Class-incremental learning with balanced embedding discrimination maximization, Neural Networks 179 (2024) 106487.

\bibitem{bib2}
R.~M. French, Catastrophic forgetting in connectionist networks, Trends in cognitive sciences 3~(4) (1999) 128--135.

\bibitem{bib22}
J.~Kim, W.~Lee, M.~Eo, W.~Rhee, Improving forward compatibility in class incremental learning by increasing representation rank and feature richness, Neural Networks 183 (2025) 106969.

\bibitem{4}
F.~Zenke, B.~Poole, S.~Ganguli, Continual learning through synaptic intelligence, in: International conference on machine learning, PMLR, 2017, pp. 3987--3995.

\bibitem{7}
D.~Lopez-Paz, M.~Ranzato, Gradient episodic memory for continual learning, Advances in neural information processing systems 30 (2017).

\bibitem{3}
M.~McCloskey, N.~J. Cohen, Catastrophic interference in connectionist networks: The sequential learning problem, in: Psychology of learning and motivation, Vol.~24, Elsevier, 1989, pp. 109--165.

\bibitem{bib3}
G.~Sun, B.~Ji, L.~Liang, M.~Chen, Cecr: Cross-entropy contrastive replay for online class-incremental continual learning, Neural Networks 173 (2024) 106163.

\bibitem{bib4}
S.~Tian, L.~Li, W.~Li, H.~Ran, X.~Ning, P.~Tiwari, A survey on few-shot class-incremental learning, Neural Networks 169 (2024) 307--324.

\bibitem{9}
R.~Aljundi, M.~Lin, B.~Goujaud, Y.~Bengio, Gradient based sample selection for online continual learning, Advances in neural information processing systems 32 (2019).

\bibitem{zhou2025vcsap}
R.~Zhou, W.~Zhu, S.~Han, M.~Kang, S.~L{\"u}, Vcsap: Online reinforcement learning exploration method based on visitation count of state-action pairs, Neural Networks (2025) 107052.

\bibitem{wang2024adaptive}
R.~Wang, J.~Wu, X.~Cheng, X.~Liu, H.~Qiu, Adaptive expert fusion model for online wind power prediction, Neural Networks (2024) 107022.

\bibitem{fan2024learning}
J.~Fan, Y.~Ge, X.~Zhang, Z.~Wang, H.~Wu, J.~Wu, Learning the feature distribution similarities for online time series anomaly detection, Neural Networks 180 (2024) 106638.

\bibitem{10654781}
Y.~Li, X.~Yang, Q.~Gao, H.~Wang, J.~Zhang, T.~Li, Cross-regional fraud detection via continual learning with knowledge transfer, IEEE Transactions on Knowledge and Data Engineering 36~(12) (2024) 7865--7877.
\newblock \href {https://doi.org/10.1109/TKDE.2024.3451161} {\path{doi:10.1109/TKDE.2024.3451161}}.

\bibitem{10643332}
L.~Yang, Z.~Luo, S.~Zhang, F.~Teng, T.~Li, Continual learning for smart city: A survey, IEEE Transactions on Knowledge and Data Engineering 36~(12) (2024) 7805--7824.
\newblock \href {https://doi.org/10.1109/TKDE.2024.3447123} {\path{doi:10.1109/TKDE.2024.3447123}}.

\bibitem{10}
R.~Aljundi, K.~Kelchtermans, T.~Tuytelaars, Task-free continual learning, in: Proceedings of the IEEE/CVF conference on computer vision and pattern recognition, 2019, pp. 11254--11263.

\bibitem{11}
Y.~He, Y.~Chen, Y.~Jin, S.~Dong, X.~Wei, Y.~Gong, Dyson: Dynamic feature space self-organization for online task-free class incremental learning, in: Proceedings of the IEEE/CVF Conference on Computer Vision and Pattern Recognition, 2024, pp. 23741--23751.

\bibitem{8}
Y.~Zhang, B.~Gao, D.~Yang, W.~L. Woo, H.~Wen, Online learning of wearable sensing for human activity recognition, IEEE Internet of Things Journal 9~(23) (2022) 24315--24327.

\bibitem{25}
F.~Zenke, B.~Poole, S.~Ganguli, Continual learning through synaptic intelligence, in: International conference on machine learning, PMLR, 2017, pp. 3987--3995.

\bibitem{24}
J.~Schwarz, W.~Czarnecki, J.~Luketina, A.~Grabska-Barwinska, Y.~W. Teh, R.~Pascanu, R.~Hadsell, Progress \& compress: A scalable framework for continual learning, in: International conference on machine learning, PMLR, 2018, pp. 4528--4537.

\bibitem{59}
M.~Masana, X.~Liu, B.~Twardowski, M.~Menta, A.~D. Bagdanov, J.~Van De~Weijer, Class-incremental learning: survey and performance evaluation on image classification, IEEE Transactions on Pattern Analysis and Machine Intelligence 45~(5) (2022) 5513--5533.

\bibitem{5}
H.~An, J.~Yang, X.~Zhang, X.~Ruan, Y.~Wu, S.~Li, J.~Hu, A class-incremental learning approach for learning feature-compatible embeddings, Neural Networks 180 (2024) 106685.

\bibitem{60}
Y.~Ghunaim, A.~Bibi, K.~Alhamoud, M.~Alfarra, H.~A. Al~Kader~Hammoud, A.~Prabhu, P.~H. Torr, B.~Ghanem, Real-time evaluation in online continual learning: A new hope, in: Proceedings of the IEEE/CVF Conference on Computer Vision and Pattern Recognition, 2023, pp. 11888--11897.

\bibitem{zhang2020new}
P.-B. Zhang, Z.-X. Yang, A new learning paradigm for random vector functional-link network: Rvfl+, Neural Networks 122 (2020) 94--105.

\bibitem{xiao2024robust}
Y.~Xiao, M.~Adegoke, C.-S. Leung, K.~W. Leung, Robust noise-aware algorithm for randomized neural network and its convergence properties, Neural Networks 173 (2024) 106202.

\bibitem{61}
A.~K. Malik, R.~Gao, M.~Ganaie, M.~Tanveer, P.~N. Suganthan, Random vector functional link network: recent developments, applications, and future directions, Applied Soft Computing 143 (2023) 110377.

\bibitem{56}
Q.~Shi, R.~Katuwal, P.~N. Suganthan, M.~Tanveer, Random vector functional link neural network based ensemble deep learning, Pattern Recognition 117 (2021) 107978.

\bibitem{62}
R.~Gao, R.~Li, M.~Hu, P.~N. Suganthan, K.~F. Yuen, Online dynamic ensemble deep random vector functional link neural network for forecasting, Neural Networks (2023).

\bibitem{63}
K.~S. Azoury, M.~K. Warmuth, Relative loss bounds for on-line density estimation with the exponential family of distributions, Machine learning 43 (2001) 211--246.

\bibitem{64}
R.~Ouhamma, O.-A. Maillard, V.~Perchet, Stochastic online linear regression: the forward algorithm to replace ridge, Advances in Neural Information Processing Systems 34 (2021) 24430--24441.

\bibitem{13}
J.~Serra, D.~Suris, M.~Miron, A.~Karatzoglou, Overcoming catastrophic forgetting with hard attention to the task, in: International conference on machine learning, PMLR, 2018, pp. 4548--4557.

\bibitem{17}
W.~Hu, Q.~Qin, M.~Wang, J.~Ma, B.~Liu, Continual learning by using information of each class holistically, in: Proceedings of the AAAI Conference on Artificial Intelligence, Vol.~35, 2021, pp. 7797--7805.

\bibitem{19}
M.~D. McDonnell, D.~Gong, A.~Parvaneh, E.~Abbasnejad, A.~van~den Hengel, Ranpac: Random projections and pre-trained models for continual learning, Advances in Neural Information Processing Systems 36 (2024).

\bibitem{22}
S.-A. Rebuffi, A.~Kolesnikov, G.~Sperl, C.~H. Lampert, icarl: Incremental classifier and representation learning, in: Proceedings of the IEEE conference on Computer Vision and Pattern Recognition, 2017, pp. 2001--2010.

\bibitem{21}
Y.~Kong, L.~Liu, M.~Qiao, Z.~Wang, D.~Tao, Trust-region adaptive frequency for online continual learning, International Journal of Computer Vision 131~(7) (2023) 1825--1839.

\bibitem{26}
R.~Aljundi, F.~Babiloni, M.~Elhoseiny, M.~Rohrbach, T.~Tuytelaars, Memory aware synapses: Learning what (not) to forget, in: Proceedings of the European conference on computer vision (ECCV), 2018, pp. 139--154.

\bibitem{28}
G.~Zeng, Y.~Chen, B.~Cui, S.~Yu, Continual learning of context-dependent processing in neural networks, Nature Machine Intelligence 1~(8) (2019) 364--372.

\bibitem{66p}
R.~Cheng, R.~Gao, M.~Hu, P.~N. Suganthan, K.~F. Yuen, Wind speed forecasting using an ensemble deep random vector functional link neural network based on parsimonious channel mixing, in: 2024 International Joint Conference on Neural Networks (IJCNN), 2024, pp. 1--8.
\newblock \href {https://doi.org/10.1109/IJCNN60899.2024.10650817} {\path{doi:10.1109/IJCNN60899.2024.10650817}}.

\bibitem{67p}
A.~Bhambu, R.~Gao, P.~N. Suganthan, Recurrent ensemble random vector functional link neural network for financial time series forecasting, Applied Soft Computing 161 (2024) 111759.

\bibitem{53p}
M.~Hihat, G.~Garrigos, A.~Fermanian, S.~Bussy, Multivariate online linear regression for hierarchical forecasting, arXiv preprint arXiv:2402.14578 (2024).

\bibitem{54p}
Z.~Zhang, D.~Bombara, H.~Yang, Discounted adaptive online learning: Towards better regularization, in: Forty-first International Conference on Machine Learning, 2024.

\bibitem{55p}
A.~Jacobsen, A.~Cutkosky, Online linear regression in dynamic environments via discounting, in: Forty-first International Conference on Machine Learning, 2024.

\bibitem{69p}
A.~Jacot, F.~Gabriel, C.~Hongler, Neural tangent kernel: Convergence and generalization in neural networks, Advances in neural information processing systems 31 (2018).

\bibitem{70p}
J.~Liu, Z.~Ji, Y.~Yu, J.~Cao, Y.~Pang, J.~Han, X.~Li, Parameter-efficient fine-tuning for continual learning: A neural tangent kernel perspective, arXiv preprint arXiv:2407.17120 (2024).

\bibitem{71p}
A.~S. Benjamin, C.~Pehle, K.~Daruwalla, Continual learning with the neural tangent ensemble, Advances in Neural Information Processing Systems 37 (2024) 58816--58840.

\bibitem{65p}
D.~Needell, A.~A. Nelson, R.~Saab, P.~Salanevich, O.~Schavemaker, Random vector functional link networks for function approximation on manifolds, Frontiers in Applied Mathematics and Statistics 10 (2024) 1284706.

\bibitem{50}
M.~B. Gurbuz, J.~M. Moorman, C.~Dovrolis, Nice: Neurogenesis inspired contextual encoding for replay-free class incremental learning, in: Proceedings of the IEEE/CVF Conference on Computer Vision and Pattern Recognition, 2024, pp. 23659--23669.

\bibitem{41}
A.~Krawczyk, A.~Gepperth, An analysis of best-practice strategies for replay and rehearsal in continual learning, in: Proceedings of the IEEE/CVF Conference on Computer Vision and Pattern Recognition, 2024, pp. 4196--4204.

\bibitem{42}
J.~Zhang, J.~Zhang, S.~Ghosh, D.~Li, S.~Tasci, L.~Heck, H.~Zhang, C.-C.~J. Kuo, Class-incremental learning via deep model consolidation, in: Proceedings of the IEEE/CVF winter conference on applications of computer vision, 2020, pp. 1131--1140.

\bibitem{43}
D.~Li, T.~Wang, B.~Xu, K.~Kawaguchi, Z.~Zeng, P.~N. Suganthan, If2net: Innately forgetting-free networks for continual learning, arXiv preprint arXiv:2306.10480 (2023).

\bibitem{44}
V.~K. Verma, K.~J. Liang, N.~Mehta, P.~Rai, L.~Carin, Efficient feature transformations for discriminative and generative continual learning, in: Proceedings of the IEEE/CVF conference on computer vision and pattern recognition, 2021, pp. 13865--13875.

\bibitem{46}
C.~Fernando, D.~Banarse, C.~Blundell, Y.~Zwols, D.~Ha, A.~A. Rusu, A.~Pritzel, D.~Wierstra, Pathnet: Evolution channels gradient descent in super neural networks, arXiv preprint arXiv:1701.08734 (2017).

\bibitem{45}
Z.~Ke, B.~Liu, X.~Huang, Continual learning of a mixed sequence of similar and dissimilar tasks, Advances in neural information processing systems 33 (2020) 18493--18504.

\bibitem{47}
D.~Deng, G.~Chen, J.~Hao, Q.~Wang, P.-A. Heng, Flattening sharpness for dynamic gradient projection memory benefits continual learning, Advances in Neural Information Processing Systems 34 (2021) 18710--18721.

\bibitem{48}
A.~Prabhu, S.~Sinha, P.~Kumaraguru, P.~H. Torr, O.~Sener, P.~K. Dokania, Randumb: A simple approach that questions the efficacy of continual representation learning, Proceedings of the IEEE/CVF conference on computer vision and pattern recognition (2024).

\bibitem{49}
D.~Li, T.~Wang, J.~Chen, W.~Dai, Z.~Zeng, Harnessing neural unit dynamics for effective and scalable class-incremental learning, in: Forty-first International Conference on Machine Learning, 2024.

\bibitem{51}
J.~Rajasegaran, M.~Hayat, S.~H. Khan, F.~S. Khan, L.~Shao, Random path selection for continual learning, Advances in neural information processing systems 32 (2019).

\bibitem{52}
S.~Tang, D.~Chen, J.~Zhu, S.~Yu, W.~Ouyang, Layerwise optimization by gradient decomposition for continual learning, in: Proceedings of the IEEE/CVF conference on Computer Vision and Pattern Recognition, 2021, pp. 9634--9643.

\bibitem{53}
E.~Belouadah, A.~Popescu, Il2m: Class incremental learning with dual memory, in: Proceedings of the IEEE/CVF international conference on computer vision, 2019, pp. 583--592.

\bibitem{54}
R.~Wang, Y.~Bao, B.~Zhang, J.~Liu, W.~Zhu, G.~Guo, Anti-retroactive interference for lifelong learning, in: European Conference on Computer Vision, Springer, 2022, pp. 163--178.

\bibitem{68p}
P.~Buzzega, M.~Boschini, A.~Porrello, D.~Abati, S.~Calderara, Dark experience for general continual learning: a strong, simple baseline, Advances in neural information processing systems 33 (2020) 15920--15930.

\bibitem{55}
M.~Boschini, L.~Bonicelli, P.~Buzzega, A.~Porrello, S.~Calderara, Class-incremental continual learning into the extended der-verse, IEEE transactions on pattern analysis and machine intelligence 45~(5) (2022) 5497--5512.

\bibitem{56p}
D.-W. Zhou, Z.-W. Cai, H.-J. Ye, D.-C. Zhan, Z.~Liu, Revisiting class-incremental learning with pre-trained models: Generalizability and adaptivity are all you need, International Journal of Computer Vision 133~(3) (2025) 1012--1032.

\bibitem{57p}
D.-W. Zhou, Q.-W. Wang, Z.-H. Qi, H.-J. Ye, D.-C. Zhan, Z.~Liu, Class-incremental learning: A survey, IEEE Transactions on Pattern Analysis and Machine Intelligence (2024).

\bibitem{58p}
D.-W. Zhou, H.-L. Sun, J.~Ning, H.-J. Ye, D.-C. Zhan, Continual learning with pre-trained models: A survey, arXiv preprint arXiv:2401.16386 (2024).

\bibitem{64p}
S.~Chen, C.~Ge, Z.~Tong, J.~Wang, Y.~Song, J.~Wang, P.~Luo, Adaptformer: Adapting vision transformers for scalable visual recognition, Advances in Neural Information Processing Systems 35 (2022) 16664--16678.

\bibitem{27}
Z.~Li, D.~Hoiem, Learning without forgetting, IEEE transactions on pattern analysis and machine intelligence 40~(12) (2017) 2935--2947.

\bibitem{59p}
L.~Yu, B.~Twardowski, X.~Liu, L.~Herranz, K.~Wang, Y.~Cheng, S.~Jui, J.~van~de Weijer, Semantic drift compensation for class-incremental learning. 2020 ieee, in: CVF Conference on Computer Vision and Pattern Recognition (CVPR), 2020, pp. 6980--6989.

\bibitem{60p}
Z.~Wang, Z.~Zhang, C.-Y. Lee, H.~Zhang, R.~Sun, X.~Ren, G.~Su, V.~Perot, J.~Dy, T.~Pfister, Learning to prompt for continual learning, in: Proceedings of the IEEE/CVF conference on computer vision and pattern recognition, 2022, pp. 139--149.

\bibitem{61p}
Z.~Wang, Z.~Zhang, S.~Ebrahimi, R.~Sun, H.~Zhang, C.-Y. Lee, X.~Ren, G.~Su, V.~Perot, J.~Dy, et~al., Dualprompt: Complementary prompting for rehearsal-free continual learning, in: European conference on computer vision, Springer, 2022, pp. 631--648.

\bibitem{62p}
J.~S. Smith, L.~Karlinsky, V.~Gutta, P.~Cascante-Bonilla, D.~Kim, A.~Arbelle, R.~Panda, R.~Feris, Z.~Kira, Coda-prompt: Continual decomposed attention-based prompting for rehearsal-free continual learning, in: Proceedings of the IEEE/CVF conference on computer vision and pattern recognition, 2023, pp. 11909--11919.

\bibitem{63p}
Z.~Li, L.~Zhao, Z.~Zhang, H.~Zhang, D.~Liu, T.~Liu, D.~N. Metaxas, Steering prototypes with prompt-tuning for rehearsal-free continual learning, in: Proceedings of the IEEE/CVF Winter Conference on Applications of Computer Vision, 2024, pp. 2523--2533.

\end{thebibliography}
  
\FloatBarrier
\newpage
\appendix
\allowdisplaybreaks
\section*{\textbf{ Supplementary material}}
\section{Bregman divergence property} 
\label{poft1}
\begin{lemma}{\cite{63}}\label{lemma 1} Bregman divergence property.  
\begin{enumerate}
\item The divergence is a linear operator. $\forall \mu  \geqslant 0$, ${\Delta _{{G_1} + \mu {G_2}}}(\bm{\tilde \beta} ,\bm{\beta} ) = {\Delta _{{G_1}}}(\bm{\tilde \beta} ,\bm{\beta} ) + \mu {\Delta _{{G_2}}}(\bm{\tilde \beta} ,\bm{\beta} )$.
\item If ${G_1}(\bm{\beta} ) - {G_2}(\bm{\beta} ) = \omega \bm{\beta}  + \upsilon, \omega  \in {\mathbb{R}^{|\bm{\beta}|}}, \upsilon  \in \mathbb{R}$, then ${\Delta _{{G_1}}}(\bm{\tilde \beta} ,\bm{\beta} ) = {\Delta _{{G_2}}}(\bm{\tilde \beta} ,\bm{\beta} )$.
\item If $G = \frac{1}{2}|| \cdot ||_2^2$, $G' = \frac{1}{2}|| \cdot ||_F^2$, $A$ and $B$ are matrices of the same size, ${\bm{a_i}}$ and ${\bm{b_i}}$ are respective column vectors, then $\sum {{\Delta _G}(\bm{a_i},\bm{b_i})}  = {\Delta _{G'}}(A,B)$.
\end{enumerate}
\end{lemma}
Proof: The proofs for the first two can be found in \cite{63}. For the third item, \textcolor{black}{we prove here:}
\begin{align}\label{appa}
{\Delta _{G'}}(A,B)&=\frac{1}{2}||A||_F^2 - \frac{1}{2}||B||_F^2 - {\left\langle {(A - B),B} \right\rangle _F}\notag\\
&= \sum {\frac{1}{2}} ||{a_i}||_2^2 - \sum {\frac{1}{2}} ||{b_i}||_2^2 - \sum {\left\langle {({a_i} - {b_i}),{b_i}} \right\rangle }\\
&=\sum {{\Delta _G}(\bm{a_i},\bm{b_i})}\notag
\end{align}
(\ref{appa}) shows that the Bregman divergence of a matrix measured by the Frobenius norm can be computed separately by column vectors, which ensures that we generalize our -$k$F algorithm to the multi-dimensional Vovk-Azoury-Warmuth (VAW) form \cite{53p}.\\
Proof finished.\hfill$\square$ 

\section{Theorem \ref{theorem 2} proof} 
\label{poft2}
Proof: (\ref{10}) can be expanded to:
\begin{align} \label{appb}
{\theta _{l,t + 1}} &= \arg {\min _{\theta_l} }{U_{l,t}}({\theta _l}) - {U_{l,t}}({\theta _{l,t}}) + {{\cal L}_{l,t}}({\theta _l})\\
& - \sum\nolimits_{ij} {({\theta _l} - {\theta _{l,t}}) \otimes \nabla {U_{l,t}}{{({\theta _{l,t}})}_{ij}}} + k \cdot {\hat {\cal L}_{l,t + 1}}({\theta _l}) - k \cdot {\hat {\cal L}_{l,t}}({\theta _l}),\notag
\end{align}
where the $ \otimes $ denotes Hadamard product. In fact, the $\nabla {U_{l,t}}{{({\theta _{l,t}})}_{ij}}=0$ because of the latest convex optimization. So that (\ref{appb}) can be simplified to:
\begin{align} \label{12}
{\theta _{l,t + 1}} &= \arg {\min _{\theta_l} }{U_{l,t + 1}}({\theta _l}) - {U_{l,t}}({\theta _{l,t}})\\
&= \arg {\min _{\theta_l} }{\Delta _{{U_{l,0}}}}({\theta_l} ,{{\theta} _{l,0}}) + {{\cal L} _{l,1..t}}({\theta_l} ) + k \cdot {{\hat {\cal L} }_{l,t + 1}}({\theta_l} )-\text{\textit{const.}}\notag
\end{align}
Solutions to both models (i.e. online learner (\ref{10}) and offline expert (\ref{8})) remain the same at each time point $t$ if designating the last target as Bregman divergence function. This also suggests that -$k$F suffers no learning dissipation compared to the offline expert because they have similar optima at each step.\\
Proof finished.\hfill$\square$

\section{Theorem \ref{theorem 3} proof}
\label{poft3}
Proof:
Following the loss function in (\ref{4}) expressed by MLE of Gaussian distribution, we define the forms of ${\Delta _{{U_{l,0}}}}$, ${{\cal L}_{l,t}}$, and ${{\hat {\cal L} }_{l,t + 1}}$. The immediate incurred loss on ${D_{l,t}}$ is denoted by ${{\cal L}_{l,t}}(\theta_l ) = \frac{1}{2}||{D_{l,t}}{\theta _l} - {Y_t}||_F^2$, and ${{\cal L}_{l,1..t}}(\theta_l ) = \sum\nolimits_{i = 1}^t {\frac{1}{2}||{D_{l,i}}{\theta _l} - {Y_i}||_F^2} $ is the cumulative loss on $t$ data batches. Similarly define the forward predictive loss to ${{\hat {\cal L} }_{l,t + 1}}(\theta_l)=\frac{1}{2}||{D_{l,t + 1}}({\theta _l} - {\theta _{l,0}})||_F^2$.

Let the initial Bregman function be ${U_{l,0}}(\theta_l ) = Tr(\frac{1}{2}{\theta_{l} ^T}{({\eta _{l,0}})^{ - 1}}\theta _l)$, where $Tr$ signifies matrix trace, ${({\eta _{l,0}})^{ - 1}}$ is a symmetric positive definite matrix. Based on Lemma \ref{lemma 1}, initial Bregman divergence is defined as ${\Delta _{{U_{l,0}}}}({\theta _l},{\theta _{l,0}}) = \sum\nolimits_{i = 1}^m {{\Delta _{{U_{{l_i},0}}}}({\theta _{{l_i}}},{\theta _{{l_i},0}})} =\sum\nolimits_{i = 1}^m {\frac{1}{2}{{({\bm{\theta} _{{l_i}}} - {\bm{\theta} _{{l_i},0}})}^T}{{({\eta _{l,0}})}^{ - 1}}({\bm{\theta} _{{l_i}}} - {\bm{\theta} _{{l_i},0}})} $ because the ${U_{l,0}}$ is a Frobenius norm projector w.r.t $\theta_l$. Define the learning processes of sub-learners of edRVFL-$k$F in OTCIL to the same form as Theorem \ref{theorem 1}. For $0 \leqslant t \leqslant T$ we have:
\begin{align}\label{13}
{\theta} _{l,t + 1} &= \arg {\min _{\theta_l} }{\text{}}U_{l,t + 1}({\theta_l} ),1 \le l \le L\\
U_{l,t + 1}({\theta_l} ) &= \sum\nolimits_{i = 1}^m {\frac{1}{2}{{({\bm{\theta} _{{l_i}}} - {\bm{\theta} _{{l_i},0}})}^T}{{({\eta _{l,0}})}^{ - 1}}({\bm{\theta} _{{l_i}}} - {\bm{\theta} _{{l_i},0}})}\notag\\
&+\sum\nolimits_{i = 1}^t {\frac{1}{2}||{D_{l,i}}{\theta _l} - {Y_i}||_F^2}+\frac{k}{2}||{D_{l,t + 1}}({\theta _l} - {\theta _{l,0}})||_F^2\notag
\end{align}

Convert (\ref{13}) into the form of (\ref{10}) described by Theorem \ref{theorem 2} to avoid the retrospective retraining, and derive variable learning rate via Lemma \ref{lemma 1}:
\begin{align}\label{14}
&{U_{l,t}}({\theta _l}) + \frac{1}{2}||{D_{l,t}}{\theta _l} - {Y_t}||_F^2  +\frac{k}{2}||{D_{l,t + 1}}({\theta _l} - {\theta _{l,0}})||_F^2 - \frac{k}{2}||{D_{l,t}}({\theta _l} - {\theta _{l,0}})||_F^2\notag\\
&={U_{l,0}}({\theta _l}) - {U_{l,0}}({\theta _{l,0}})-{\sum\nolimits_{ij} {({\theta _l} - {\theta _{l,0}}) \otimes \nabla {U_{l,0}}({\theta _{l,0}})} _{ij}}  \notag\\
&+\sum\nolimits_{i = 1}^t {\frac{1}{2}||{D_{l,i}}{\theta _l} - {Y_i}||_F^2}  + \frac{k}{2}||{D_{l,t + 1}}({\theta _l} - {\theta _{l,0}})||_F^2\notag\\
& \Rightarrow{U_{l,t}}({\theta _l}) - {U_{l,0}}({\theta _l})\notag\\
&=\sum\nolimits_{i = 1}^{t - 1} {\frac{1}{2}||{D_{l,i}}{\theta _l} - {Y_i}||_F^2 + \frac{k}{2}||{D_{l,t}}({\theta _l} - {\theta _{l,0}})||_F^2} \notag\\
& - Tr(\frac{1}{2}\theta _{l,0}^T{({\eta _{l,0}})^{ - 1}}{\theta _{l,0}}) - \sum\nolimits_{i = 1}^m {{{({\theta _{{l_i}}} - {\theta _{{l_i},0}})}^T} \cdot {{({\eta _{l,0}})}^{ - 1}}{\theta _{{l_i},0}}} \\
&=  \sum\nolimits_{i = 1}^{t - 1} {\frac{1}{2}||{D_{l,i}}{\theta _l} - {Y_i}||_F^2 + \frac{k}{2}||{D_{l,t}}({\theta _l} - {\theta _{l,0}})||_F^2}  \notag\\
&+ Tr(\frac{1}{2}\theta _{l,0}^T{({\eta _{l,0}})^{ - 1}}{\theta _{l,0}}) - Tr(\theta _l^T{({\eta _{l,0}})^{ - 1}}{\theta _{l,0}})\notag\\
&=Tr(\frac{1}{2}\theta _l^T\sum\nolimits_{i = 1}^{t - 1} {D_{l,i}^T{D_{l,i}}} {\theta _l}) + Tr(\frac{1}{2}\sum\nolimits_{i = 1}^{t - 1} {Y_i^T{Y_i}} )+ Tr(\frac{1}{2}\theta _{l,0}^T{({\eta _{l,0}})^{ - 1}}{\theta _{l,0}}) \notag\\
&- Tr(\sum\nolimits_{i = 1}^{t - 1} {Y_i^T{D_{l,i}}{\theta _l}} )- Tr(\theta _l^T{({\eta _{l,0}})^{ - 1}}{\theta _{l,0}}) + Tr(\frac{k}{2}{({\theta _l} - {\theta _{l,0}})^T}D_{l,t}^T{D_{l,t}}({\theta _l} - {\theta _{l,0}}))\notag
\end{align}

Based on Lemma \ref{lemma 1}, (\ref{14}) can be simplified to:
\begin{align}\label{15}
&{U_{l,t}}({\theta _l}) - ({U_{l,0}}({\theta _l})+Tr(\frac{1}{2}\theta _l^T\sum\nolimits_{i = 1}^{t - 1} {D_{l,i}^T{D_{l,i}}} {\theta _l}) 
+ Tr(\frac{k}{2}\theta _l^TD_{l,t}^T{D_{l,t}}{\theta _l})  )\\
&=Tr(\omega  \cdot {\theta _l}) + const.\notag
\end{align}
where $\omega  \in {\mathbb{R}^{m \cdot (s + N)}}$. If set ${V_{l,t}} = Tr(\frac{1}{2}\theta _l^T(\sum\nolimits_{i = 1}^{t - 1} {D_{l,i}^T{D_{l,i}} + k \cdot D_{l,t}^T{D_{l,t}})} {\theta _l})$, relation can be deduced from (\ref{15}) and Lemma \ref{lemma 1}:
\begin{align}\label{16}
&{\Delta _{{U_{l,t}}}}({\theta _l},{\theta _{l,t}}){\rm{ }} = {\Delta _{U_{l,0}^{} + V_{l,t}^{}}}({\theta _l},{\theta _{l,t}}) 
= {\Delta _{U_{l,0}^{}}}({\theta _l},{\theta _{l,t}}) + {\Delta _{V_{l,t}^{}}}({\theta _l},{\theta _{l,t}})\\
=&\sum\nolimits_{i = 1}^m {\frac{1}{2}{{({\bm{\theta} _{{l_i}}} - {\bm{\theta} _{{l_i},t}})}^T}[{{({\eta _{l,0}})}^{ - 1}} + \sum\nolimits_{j = 1}^{t - 1} {D_{l,j}^T{D_{l,j}}}  }
+{{  k \cdot D_{l,t}^T{D_{l,t}}}]({\bm{\theta} _{{l_i}}} - {\bm{\theta} _{{l_i},t}})} \notag
\end{align}

The above derivation applies well to the ERM paradigm combined with Gaussian MLE and differentiable convex regularization terms, which could bring beneficial effects and enhancement to the learning process in OTCIL. (\ref{16}) shows that it alleviates the previous data revisit with all the knowledge absorbed in Bregman divergence. The solution to (\ref{16}) that needs a stepwise solver during the OTCIL process will change with ${({\eta _{l,t}})^{ - 1}}={{({\eta _{l,0}})}^{ - 1}} + \sum\nolimits_{i = 1}^{t - 1} {D_{l,i}^T{D_{l,i}}}  +{  k \cdot D_{l,t}^T{D_{l,t}}}$, also termed the variable learning rate. The essence of the edRVFL-$k$F algorithm is also to provide time-varying optimization targets to guide the optimal updates of the network in task streams. This optimization objective in Theorem \ref{theorem 2} can be expressed further as:
\begin{align} \label{17}
&\theta _{l,t + 1} = \arg {\min _{\theta_l} }{\text{}}{\Delta _{U_{l,t}}}(\theta_l ,\theta _{l,t}) + {{\cal L}_{l,t}}(\theta_l )+ k \cdot{\hat {\cal L}_{l,t + 1}}(\theta_l ) - k \cdot{\hat {\cal L}_{l,t}}(\theta_l )\\
&= \arg {\min _{\theta_l} }\sum\nolimits_{i = 1}^m {\frac{1}{2}{{({\bm{\theta} _{{l_i}}} - {\bm{\theta} _{{l_i},t}})}^T}[{{({\eta _{l,0}})}^{ - 1}} }+{ \sum\nolimits_{j = 1}^{t - 1} {D_{l,j}^T{D_{l,j}}}  +{  k \cdot D_{l,t}^T{D_{l,t}}}]({\bm{\theta} _{{l_i}}} - {\bm{\theta} _{{l_i},t}})} \notag\\
&+\frac{1}{2}||{D_{l,t}}{\theta _l} - {Y_t}||_F^2+k\cdot\frac{1}{2}||{D_{l,t + 1}}({\theta _l}- {\theta _{l,0}})||_F^2 -k\cdot\frac{1}{2}||{D_{l,t }}({\theta _l} - {\theta _{l,0}})||_F^2\notag
\end{align}

In order to maintain the generality of the above derivation steps to any OCO problems, (\ref{17}) is transformed to the following constrained optimization problem:
\begin{align} \label{18}
&\min\limits_{{\theta _l} \in {\theta _{l,t + 1}}} \sum\nolimits_{i = 1}^m {\frac{1}{2}{{({\bm{\theta} _{{l_i}}} - {\bm{\theta} _{{l_i},t}})}^T}{{({\eta _{l,t}})}^{ - 1}}({\bm{\theta} _{{l_i}}} - {\bm{\theta} _{{l_i},t}})}  \\
&\phantom{{}= \hspace{0.8cm}}+ \frac{1}{2}||{\xi _1}||_F^2 + k \cdot \frac{1}{2}||{\xi _2}||_F^2 - k \cdot \frac{1}{2}||{\xi _3}||_F^2\notag \\
&\phantom{{}= \hspace{0.3cm}}s.t.\;{D_{l,t}}{\theta _l} - {Y_t} = {\xi _1};{D_{l,t + 1}}({\theta _l} - {\theta _{l,0}}) = {\xi _2};{D_{l,t}}({\theta _l} - {\theta _{l,0}}) = {\xi _3},\forall t\notag 
\end{align}
where ${\xi _{\{ 1,2,3\} }}$ denotes the gap between ground truth and prediction. The Lagrangian function of problem (\ref{18}) is:
\begin{align} \label{19}
&\ell ({\theta _l},{\xi _{\{ 1,2,3\} }},{\mu _{\{ 1,2,3\} }})= \sum\nolimits_{i = 1}^m \frac{1}{2}{{({\bm{\theta} _{{l_i}}} - {\bm{\theta} _{{l_i},t}})}^T}{{({\eta _{l,t}})}^{ - 1}}({\bm{\theta} _{{l_i}}} - {\bm{\theta} _{{l_i},t}}) \\
&+ \frac{1}{2}||{\xi _1}||_F^2 + k \cdot \frac{1}{2}||{\xi _2}||_F^2 - k \cdot \frac{1}{2}||{\xi _3}||_F^2 + Tr(\mu _1^T( {D_{l,t}}{\theta _l} - {Y_t} - {\xi _1})) \notag\\
&+ Tr(\mu _2^T({D_{l,t + 1}}({\theta _l} - {\theta _{l,0}}) - {\xi _2})) + Tr(\mu _3^T({D_{l,t}}({\theta _l} - {\theta _{l,0}}) - {\xi _3})),\notag
\end{align}
where $\mu$ denotes the Lagrangian multiplier and is the same size as $\xi$. The Lagrangian function (\ref{19}) can be tackled through Karush-Kuhn-Tucker (KKT) conditions, which can be built into the following formulation:
\begin{align} \label{20.1}
&\frac{{\partial \ell ({\theta _l},\xi ,\mu )}}{{\partial {\theta _l}}} = 0 \Rightarrow {({\eta _{l,t}})^{ - 1}}({\theta _l} - {\theta _{l,t}}) + D_{l,t}^T\mu _1^{} + D_{l,t + 1}^T\mu _2^{} + D_{l,t}^T\mu _3^{} = 0\notag \\
&\frac{{\partial \ell ({\theta _l},\xi ,\mu )}}{{\partial {\xi _1}}} = 0 \Rightarrow {\xi _1} = \mu _1^{}\notag \\
&\frac{{\partial \ell ({\theta _l},\xi ,\mu )}}{{\partial {\xi _2}}} = 0 \Rightarrow k{\xi _2} = \mu _2^{}\notag \\
&\frac{{\partial \ell ({\theta _l},\xi ,\mu )}}{{\partial {\xi _3}}} = 0 \Rightarrow  - k{\xi _3} = \mu _3^{} \\
&\frac{{\partial \ell ({\theta _l},\xi ,\mu )}}{{\partial {\mu _1}}} = 0 \Rightarrow {D_{l,t}}{\theta _l} - {Y_t} = {\xi _1}\notag \\
&\frac{{\partial \ell ({\theta _l},\xi ,\mu )}}{{\partial {\mu _2}}} = 0 \Rightarrow {D_{l,t + 1}}({\theta _l} - {\theta _{l,0}}) = {\xi _2}\notag \\
&\frac{{\partial \ell ({\theta _l},\xi ,\mu )}}{{\partial {\mu _3}}} = 0 \Rightarrow {D_{l,t}}({\theta _l} - {\theta _{l,0}}) = {\xi _3}\notag
\end{align}

Based on (\ref{20.1}), we can obtain the recursive updating policy between $\theta _{l,t}$ and $\theta _{l,t+1}$ as follows:
\begin{align} \label{20}
{\theta _{l,t + 1}} &={\theta _{l,t}} - {\eta _{l,t + 1}}[(D_{l,t}^TD_{l,t}^{} + k \cdot D_{l,t + 1}^TD_{l,t + 1}^{}- k \cdot D_{l,t}^TD_{l,t}^{}){\theta _{l,t}} - D_{l,t}^T{Y_t}]  \notag\\
&+ k \cdot {\eta _{l,t + 1}}(D_{l,t + 1}^TD_{l,t + 1}^{} - D_{l,t}^TD_{l,t}^{}){\theta _{l,0}} 
\end{align}

Without loss of generality, given $\theta _{l,0} = 0$ or $\{D_{l,t + 1},D_{l,t }\}$ both drawn from i.i.d assumption, to allow the learner to start from no prior knowledge under strict conditions, we have:
\begin{align} \label{21}
{\theta _{l,t + 1}} &={\theta _{l,t}} - {\eta _{l,t + 1}}[(D_{l,t}^TD_{l,t}^{} + k D_{l,t + 1}^TD_{l,t + 1}^{}- k D_{l,t}^TD_{l,t}^{}){\theta _{l,t}} - D_{l,t}^T{Y_t}]  
\end{align}

This suggests that the relevant weight is not learned and remains 0 until after a new class arrives. Using this methodology, we can elegantly obtain the variable learning rate and weight updating policy. (\ref{21}) endows the edRVFL-$k$F with the innate ability of \textit{non-replay} and \textit{non-increase parameter}, because the values are only relevant to the chunks of current and next data batches. The model can deliver immediate and accurate decisions, profiting from recursive closed-form solutions. The algorithmic procedures of edRVFL-$k$F in OTCIL could be found in \textbf{Algorithm 1}.\\
Proof finished.\hfill$\square$

\end{document}